%% file: main.tex
\documentclass[10pt,twocolumn,letterpaper]{article}

\usepackage[pagenumbers]{cvpr} 

\input{preamble}

\definecolor{cvprblue}{rgb}{0.21,0.49,0.74}
\usepackage[pagebackref,breaklinks,colorlinks,citecolor=cvprblue]{hyperref}

\usepackage{graphicx}
\usepackage{amsmath}
\usepackage{amssymb}
\usepackage{mathtools}
\usepackage{amsthm}
\usepackage[algo2e]{algorithm2e}
\usepackage{multirow}
\usepackage{algorithm} 
\usepackage{algorithmic}
\usepackage{comment}
\usepackage{color, colortbl,soul}
\definecolor{Gray}{gray}{0.9}
\usepackage{booktabs}
\usepackage{tabularx, booktabs, makecell, caption}
\usepackage{siunitx}
\usepackage{multicol}
\usepackage{subcaption}
\usepackage{lipsum,calc}
\usepackage{xspace}

\theoremstyle{plain}
\newtheorem{theorem}{Theorem}

\newtheorem{lemma}{Lemma}

\theoremstyle{definition}

\theoremstyle{remark}

\def\paperID{14237} 
\def\confName{CVPR}
\def\confYear{2024}

\title{Communication-Efficient Federated Learning with Accelerated Client Gradient}

\author{
Geeho Kim\footnotemark[1]~~$^1$ \qquad  Jinkyu Kim\thanks{indicates equal contribution.}~~$^1$  \qquad Bohyung Han$^{1,2}$ \\
$^1$ECE \& $^{2}$IPAI, Seoul National University\\
 {\tt\small \{snow1234, jinkyu, bhhan\}@snu.ac.kr}
}

\begin{document}
\maketitle

\input{sections/abstract.tex}
\input{sections/introduction.tex}

\input{sections/related_work.tex}
\input{sections/method.tex}

\input{sections/experiments.tex}

\input{sections/conclusion.tex}
\input{sections/acknowledgments.tex}

{
    \small
    \bibliographystyle{ieeenat_fullname}
    \bibliography{main}
}

\input{supple/supple.tex}

\end{document}

%% file: preamble.tex
\usepackage[dvipsnames]{xcolor}


%% file: sections/abstract.tex

\begin{abstract}
Federated learning often suffers from slow and unstable convergence due to the heterogeneous characteristics of participating client datasets.
Such a tendency is aggravated when the client participation ratio is low since the information collected from the clients has large variations.
To address this challenge, we propose a simple but effective federated learning framework, which improves the consistency across clients and facilitates the convergence of the server model.
This is achieved by making the server broadcast a global model with a lookahead gradient.
This strategy enables the proposed approach to convey the projected global update information to participants effectively without additional client memory and extra communication costs.
We also regularize local updates by aligning each client with the overshot global model to reduce bias and improve the stability of our algorithm.
We provide the theoretical convergence rate of our algorithm and demonstrate remarkable performance gains in terms of accuracy and communication efficiency compared to the state-of-the-art methods, especially with low client participation rates.
The source code is available at our project page\footnote{\label{footnote} \url{https://github.com/geehokim/FedACG}}.

\end{abstract}

%% file: sections/introduction.tex

\section{Introduction} 
\label{sec:introduction}

Federated learning (FL)~\citep{mcmahan2017communication} is a large-scale machine learning framework that learns a shared model in a central server through collaboration with a large number of remote clients with separate datasets.
Such a decentralized learning framework achieves a basic level of data privacy because the data stored in local clients are unobservable by the server and other clients.
On the other hand, federated learning algorithms are particularly sensitive to communication and computational costs due to limited resources on many clients, such as mobile or IoT devices.

A baseline algorithm of federated learning, FedAvg~\citep{mcmahan2017communication}, updates a subset of client models using a gradient descent method based on their local data.
The resulting models are then uploaded to the server for estimating the global model parameters via model averaging.
As extensively discussed in the convergence of FedAvg~\citep{stich2018local,yu2019parallel,wang2021cooperative,stich2019error,basu2020qsparse}, multiple local updates conducted before server-side aggregation provide theoretical support and practical benefits of federated learning by reducing communication costs significantly.

Despite its initial success, federated learning faces two key challenges: high heterogeneity in training data distributed over clients and limited client participation rates.
%
Several studies~\citep{zhao2018federated,karimireddy2019scaffold} have shown that multiple local updates in the clients with non-\textit{i.i.d.}~(independent and identically distributed) data lead to client model drift, in other words, diverging updates in individual clients.
Such a phenomenon introduces a high variance issue in FedAvg steps for global model updates, which hampers convergence to the optimal average loss over all clients~\citep{li2020federated,wang2019adaptive,khaled2019first,li2019convergence,hsieh2020non,wang2020tackling}.
The challenge related to client model drift is exacerbated when the client participation rate per communication round is low.

To properly address the issue of client heterogeneity, we propose a novel federated learning algorithm, Federated averaging with Accelerated Client Gradient (FedACG), which conveys the momentum of the global gradient to clients and enables the momentum to be incorporated into the local updates in the individual clients.
Specifically, FedACG transmits the global model integrated with the global momentum in the form of a single message, which allows each client to perform its local gradient update step along the landscape of the global loss function.
This approach is effective in reducing the gap between global and local losses.
In addition, FedACG adds a regularization term in the objective function of clients to make the local gradients more consistent across clients.
We show that subtle differences in federated learning algorithms can have a significant impact on the final results and discuss the behavior of FedACG together with related methods.

The main contributions of this paper are summarized as follows.
%

 \begin{itemize}
 
	
	\item[$\bullet$] We propose a simple yet effective federated optimization algorithm, called FedACG, that proactively adjusts the initial local model using a lookahead gradient and aligns the gradients of individual clients with that of the server.


	 \item[$\bullet$] FedACG is free from additional communication costs, extra computation in the server, and memory overhead of clients; these properties are desirable for the real-world settings of federated learning. \vspace{1mm}




 \item[$\bullet$] FedACG demonstrates its outstanding performance in terms of communication efficiency and robustness to client heterogeneity, especially with the low client participation rates.

\item[$\bullet$] We introduce a federated learning benchmark$^{\text{\ref{footnote}}}$ to facilitate the evaluation of federated learning algorithms. The benchmark contains the implementations of various algorithms including FedACG.

\end{itemize}

The rest of the paper is organized as follows.
We first review the prior works in Section~\ref{sec:related}.
Section~\ref{sec:proporsed_method} discusses the technical details of the proposed approach. 
Section~\ref{sec:exp} validates its effectiveness.
We conclude our paper in Section~\ref{sec:conclusion}.

%% file: sections/related_work.tex

\section{Related Work} 
\label{sec:related}

Federated learning is first introduced by~\citet{mcmahan2017communication}.
They formulate the problem and propose FedAvg as a solution to address main challenges in federated learning, such as massively distributed clients and partial client participation.
Subsequent works attempt to address the challenge of non-\textit{i.i.d.}~client data in federated learning empirically~\citep{zhao2018federated} and derive convergence rates depending on the level of heterogeneity~\citep{li2020federated,wang2019adaptive,khaled2019first,li2019convergence,hsieh2020non,wang2020tackling}.

There exists a long line of research on client-side optimization aimed at reducing the divergence of clients from the global model.
FedProx~\citep{li2020federated} penalizes the difference between the server and client parameters, while FedDyn~\citep{acar2021federated} and FedPD~\citep{zhang2020fedpd} adopt cumulative gradients of each client for dynamic regularization of local updates.
FedDC~\citep{gao2022feddc} introduces the auxiliary drift variables of each client to reduce the impact of the local drift on the global objective.
Another line of work adopts variance reduction techniques in client updates to eliminate inconsistent updates across local models.
SCAFFOLD~\citep{karimireddy2019scaffold} and Mime~\citep{karimireddy2021breaking} use control variates for local updates, while FedDANE~\citep{li2019feddane} adds a gradient correction term based on the server gradient.
FedPA~\citep{al2020federated} reduces the bias in client updates by estimating the global posterior on the client side.
On the other hand, some approaches adopt a contrastive loss~\citep{li2021model,mu2021fedproc,seo2024relaxed}, knowledge distillation~\citep{kim2022multi,lee2022preservation}, logit calibration~\cite{zhang2022Federated}, feature decorrelation~\cite{shi2022towards}, or a generative model~\citep{zhu2021data} to ensure the similarity between the representations in the global and local models.
FedSAM~\citep{qu2022generalized} and FedASAM~\citep{caldarola2022improving} apply SAM~\citep{foret2020sharpness} as a client-side optimizer to reduce the gap between global and local losses.
However, most of these methods require full participation~\citep{zhang2020fedpd,khanduri2021stem,mu2021fedproc}, additional communication costs~\citep{xu2021fedcm,karimireddy2019scaffold,zhu2021data,karimireddy2021breaking,li2019feddane,das2020faster,gao2022feddc}, or extra client storage~\citep{acar2021federated,karimireddy2019scaffold,li2021model,gao2022feddc}, which are problematic in realistic federated learning scenarios.

Momentum-based optimization techniques have also been explored for the stability and speed-up of convergence.
These approaches incorporate a momentum SGD~\citep{hsu2019measuring,wang2019slowmo} or an adaptive gradient-descent method~\citep{reddi2021adaptive,caldarola2022improving} into server model updates while {FedCM~\citep{xu2021fedcm} and FedADC~\citep{ozfatura2021fedadc} employs global momentum to correct gradients in local updates.}
STEM~\citep{khanduri2021stem} and FedGLOMO~\citep{das2020faster} apply the STORM algorithm~\citep{cutkosky2019momentum} to both server- and client-level SGD procedures to reduce variance in server model updates.
Although these methods require additional communication overhead to transmit the global momentum for local updates, FedACG saves the costs by broadcasting the momentum-integrated model as a single message.


Meanwhile, there is another set of works that aims to reduce the communication costs per round by compressing the transmitted model.
FedPAQ~\citep{reisizadeh2020fedpaq}, FedCAMS~\citep{wang2022communication}, and FedCOMGATE~\citep{haddadpour2021federated} use low-bit precision to quantize the communicated messages, while
FedPara~\citep{nam2022fedpara} reparameterizes the model parameters using a low-rank Hadamard product.
These works are orthogonal to ours but can be integrated into our algorithm.

%% file: sections/method.tex
\section{Proposed Approach}
\label{sec:proporsed_method}


\subsection{Preliminaries}
\label{sec:prelim}

\paragraph{Problem setup}

Let $\mathcal{L}_i(\theta) := \mathbb{E}_{(\mathbf{x}, y) \sim \mathcal{D}_i}[\ell_i((\mathbf{x},y);\theta)]$ be the empirical loss function of the client $C_i \in \left\{C_1,\dots,C_N \right\}$ with a local dataset denoted by $\mathcal{D}_i$.
Then, our goal is to train a model that minimizes the average loss of all clients as follows:
\begin{equation}
\label{global_objective}
    \underset{\theta}{\operatorname{min}}\left\{\mathcal{L}(\theta) := \sum_{i=1}^N \omega_i \mathcal{L}_i(\theta)\right\},
\end{equation}
where $\theta$ is the parameter of the global model and $\omega_i$ is the normalized weight of the $i^\text{th}$ client, which is proportional to the size of the local dataset $|\mathcal{D}_i|$.
We focus on the non-\textit{i.i.d.} data setting, where local datasets have heterogeneous distributions.
Note that the communication of raw data between clients and the server is strictly prohibited in principle due to privacy concerns.

\begin{figure}[t]
\centering
\includegraphics[width=1\linewidth]{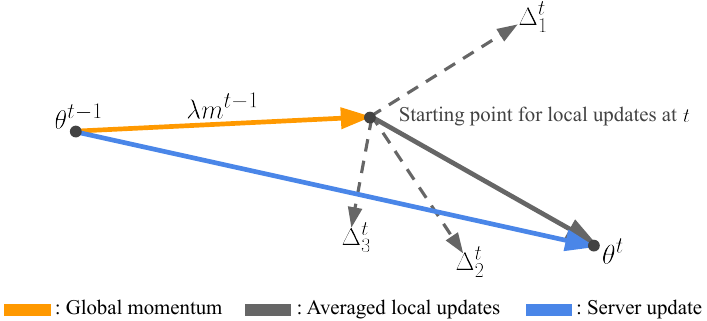}
\caption{\textbf{An illustration of the proposed accelerated client gradient method.}
We first partially update the global model in the direction of the global momentum (orange) and then aggregate local updates (gray), resulting in the server model in the next round (blue).
This anticipatory update aligns individual local updates with the global gradient, achieving speed-up of convergence.
}

\label{fig:gradient_acceleration}
\end{figure}

\paragraph{FedAvg algorithm}

FedAvg~\citep{mcmahan2017communication} is a standard solution of federated learning, where the server simply aggregates all the participating client models to obtain the global model.
Specifically, in the $t^\text{th}$ communication round, the server broadcasts the latest global model, represented as $\theta^{t-1}$ to the active clients in $\mathcal{S}_{t} \subseteq \{C_1,\dots, C_N\}$.
Each participating client optimizes its local model by using the global model as its initial point,~\ie, $\theta_{i,0}^t:=\theta^{t-1}$.
After $K$ iterations of the local training,  each client uploads its local updates $\Delta_{i}^t := \theta_{i,K}^t -\theta_{i,0}^t$ to the server, and then the server aggregates them as $\Delta^{t} := \sum_{i \in S_t} \omega_i \Delta_{i}^{t}$.
The server constructs the next server model $\theta^t := \theta^{t-1} + \Delta^{t}$ for broadcasting in the next round.
Due to non-\textit{i.i.d.}~data and limited client participation rate in each round of training, FedAvg suffers from client drift~\citep{karimireddy2019scaffold}.
Such a phenomenon results in the inconsistent updates of client models caused by overfitting to local data of individual clients, which consequently leads to the high variance of the global model.
This tendency is aggravated over multiple communication rounds in federated learning because each client initializes its parameters using the global model.

\subsection{Federated averaging with Accelerated Client Gradient (FedACG)}
\label{sec:method}

To reduce the inconsistency between the local models and the divergence of the resulting global model, we incorporate global momentum into the local models to guide local updates.

\paragraph{Accelerated client gradient}

The main idea of FedACG is to revise the initialization of client models using the global model integrated with a global gradient, allowing more effective and stable updates of local models.
Since the direct computation of the global gradient is impractical in FL, we utilize the global momentum $m^{t-1}$ as a viable approximation, which is updated as $m^{t-1} := \lambda m^{t-2} + \Delta^{t-1}$, at each round.
Specifically, in the $t^\text{th}$ communication round, the server augments the recent global model $\theta^{t-1}$ with the global momentum $m^{t-1}$.
As illustrated in Figure~\ref{fig:gradient_acceleration}, The server then broadcasts this accelerated global model, represented as $\theta^{t-1} + \lambda m^{t-1}$, as a single message to the active clients in $\mathcal{S}_{t} \subseteq \{1,\dots, N\}$.
Note that $\lambda \in [0, 1)$ controls the importance of the global momentum.
Each participating client optimizes its local model from the momentum-integrated initialization.
This proactive initialization allows each client to find its local optimal solution along the trajectory of the global gradient, which improves the consistency of local updates in FedACG.
Our approach has a similar motivation with meta-learning~\citep{finn2017model}, where a meta-learner identifies the optimal point to facilitate the optimization in all target tasks.
After $K$ iterations of local training, the server updates its momentum and constructs the next server model, denoted as $\theta^t := \theta^{t-1} + m^{t}$, in preparation for the next round.
Algorithm~\ref{alg:proposed_method} outlines the procedure of FedACG.

\paragraph{Regularization with momentum-integrated model}

In addition to the initial acceleration for local training, we augment the client's loss function with the quadratic term $\frac{\beta}{2}\|\theta_{i,k}^t- (\theta^{t-1} +${$ \lambda m^{t-1})\|^2$} which penalizes the difference between the local online model $\theta_{i,k}^t$ and the accelerated global model $\theta^{t-1} + \lambda m^{t-1}$.
Note that, $\beta$ controls the intensity of the penalty. 
The penalized term takes advantage of the global gradient information $\lambda m^t$ to reduce the variations of client-specific gradients, $\Delta_i^t$.
This regularization term further enforces the local model not to deviate from the accelerated point, preventing each client from falling into biased local minima.

\setlength{\textfloatsep}{14pt}
\begin{algorithm}[t]
\caption{FedACG}
\label{alg:proposed_method}
\SetAlgoLined
\KwIn{$\beta$, $\lambda$, initial server model $\theta^0$, number of clients $N$,  number of communication rounds $T$, number of local iterations $K$, local learning rate $\eta$}
Initialize global momentum $m^{0}=0$

 \For{$\text{each round}~t = 1,2, \dots ,T$}{
  Sample subset of clients $S_t \subseteq \{C_1,\dots, C_N\}$

Server sends $\theta^{t-1} +${$ \lambda m^{t-1}$} for all clients in $S_t$

\For{$\text{each client}~ C_i \in S_t,~\textbf{in parallel}$}{
Initialize local model $\theta_{i,0}^t \leftarrow \theta^{t-1} +${$ \lambda m^{t-1}$}

\For{$\text{each local iteration}~k = 1,2, \dots ,K$}{
Compute mini-batch loss 

$f_{i}(\theta^t_{i,k-1}) \leftarrow \ell_i(\theta_{i,k-1}^t)$ \\
 ${}$\hspace{4em}  $+ \frac{\beta}{2}\|\theta_{i,k-1}^t-  (\theta^{t-1} + \lambda m^{t-1})  \|^2$

$\theta_{i,k}^t \leftarrow \theta_{i,k-1}^t - \eta \nabla f_{i}(\theta^t_{i,k-1})$
}
$\Delta_{i}^t \leftarrow \theta_{i,K}^t -\theta_{i,0}^t$

Client sends  $\Delta_{i}^t$ back to the server
}
\textbf{In server:}

\quad $\Delta^{t}$ = $ \sum_{i \in S_t} \omega_i \Delta_{i}^{t}$

\quad $m^{t}=\lambda m^{t-1} + \Delta^{t}$

\quad $\theta^t = \theta^{t-1} + m^{t}$
}
\textbf{Return} ~$\theta^{t}$
\end{algorithm}

\subsection{Discussion}
While our formulation has something in common with the existing works that also address client heterogeneity using global gradient information for local updates, FedACG has major advantages.
%
First, unlike~\citep{karimireddy2019scaffold, xu2021fedcm, gao2022feddc}, the server and clients in FedACG communicate only model parameters without imposing additional network overhead for transmitting gradients and other information; the server broadcasts $(\theta^{t-1} + \lambda m^{t-1})$ as a single message and each client sends $\Delta_{i}^t$ to the server.
This is a critical benefit because the rise in communication cost challenges many practical federated learning applications involving clients with limited network bandwidths.
%
Second, FedACG is robust to the low participation rate of clients and allows new-arriving clients to join the training process without a warm-up period because, unlike~\citep{karimireddy2019scaffold, acar2021federated, li2021model, gao2022feddc}, clients are supposed to neither store their local states nor use them for model updates.

\paragraph{Comparison with FedAvgM}

Although FedACG apparently looks similar to FedAvgM in the sense that both methods employ the global momentum for optimization, they have critical differences.
To analyze the difference between the two algorithms, we decompose the process of updating the global model in FedACG into two steps:
1) updating the previous global model, $\theta^{t-1}$, with the global momentum term, $\lambda m^t$, and computing the interim model $\Phi^{t-1} \! =  \theta^{t-1} \!+\! \lambda m^{t-1}$, and
2) updating the interim model with the aggregated local gradients, $\Delta^{t}|_{\Phi^{t-1}}$, to derive the new global model, $\theta^t = \Phi^{t-1} + \Delta^{t}$.
While FedAvgM performs the local updates from the previous global model,~\ie, $\Delta^{t} \!:= \!\Delta^{t}|_{\theta^{t-1}}$, FedACG computes the local updates from the interim model parameter,~\ie, $\Delta^{t}\!:= \!\Delta^{t}|_{\Phi^{t-1}}$.
In other words, the two algorithms have different initializations, $\theta^{t-1}$ vs. $\Phi^{t-1}$, for training local models.
%
As will be discussed in Section~\ref{sec:ablation}, FedACG's lookahead initialization results in more robust models compared to FedAvgM.
Such a difference accumulates over multiple rounds communication rounds and eventually leads to significant performance gaps.
\subsection{Convergence analysis of FedACG}

We now discuss the convergence of the FedACG algorithm in general non-convex FL scenarios. 
To establish the convergence analysis for FedACG, we make three assumptions; (1) the local loss function $\mathcal{F}_i(\cdot)$ is $L$-smooth, (2) its stochastic gradient $\nabla {f}_i(\mathbf{x}) :=  \nabla \mathcal{F}_i(\mathbf{x}; \mathcal{D}_i)$ is unbiased and possesses a bounded variance,~\ie, $\mathbb{E}_{\mathcal{D}_i} \| \nabla f_i ({\mathbf{x}}) - \nabla \mathcal{F}_i(\mathbf{x}) \| < \sigma^{2}$, and (3) the average norm of local gradients is bounded by a function of the global gradient magnitude as $\frac{1}{N} \sum_{i=1}^N\left\|\nabla \mathcal{F}_i(\mathbf{x})\right\|^2 \leq \sigma _{g}^2+B^2\|\nabla \mathcal{F}(\mathbf{x})\|^2$, where $\sigma_g \geq 0$ and $B \geq 1$.
These assumptions are widely used for analyzing the non-convex loss functions in FL in the previous works~\citep{karimireddy2019scaffold,xu2021fedcm,acar2021federated,reddi2021adaptive,karimireddy2021breaking,qu2022generalized}.
Note that our convergence proof is free from the bounded gradient assumption of the global or local loss while it is commonly used for the proofs in momentum-based or adaptive optimization methods~\citep{reddi2021adaptive,xu2021fedcm,wu2023faster}.

We now state the convergence result of FedACG; the detailed proof is provided in Section~\ref{sec:convergence_proof} in the supplementary document.

\begin{theorem}
\label{thm:convergence}
\textbf{(Convergence for non-convex functions)} Suppose that local functions $  \lbrace \mathcal {F}_i \rbrace _{i=1}^{N}$ are non-convex and $L$-smooth.
By setting $\eta \leq \frac{(1-\lambda)^2}{64KL(B^2+1)}$, FedACG satisfies
\begin{align}
\label{eq:convergence}
&\min_{t=1, \ldots, T} \mathbb{E}\left\|\nabla \mathcal{F}\left(\theta^{t-1} + \lambda m^{t-1} \right)\right\|^2 \nonumber \\ 
 &  \hspace{0.0em} \leq \mathcal{O}\left(\frac{ M_1 \sqrt{LD}}{\sqrt{T K |S_t|}}
 \! + \! \frac{\left(LD(1-\lambda)^2\right)^{\frac{2}{3}}M_2^{\frac{1}{3}}}{(T+1)^{\frac{2}{3}}}
 \! + \! \frac{B^2 L D}{T}\right), \nonumber
\end{align}
where $M_1^2:=\sigma^2+K\left(1-\frac{|S_t|}{N}\right) \sigma_g^2$, $M_2:=\frac{\sigma^2}{K} + \sigma_g^2$, and $D:=\frac{\mathcal{F}\left(\theta^0\right)-\mathcal{F}(\theta^{*})}{1 - \lambda}$.
\end{theorem}
%
Theorem~\ref{thm:convergence} provides the convergence rate of FedACG, which matches the best convergence rate of existing FL methods~\citep{karimireddy2019scaffold,karimireddy2021breaking}.

%% file: sections/experiments.tex

\section{Experiments}
\label{sec:exp} 

We present empirical evaluation results of FedACG and competing federated learning methods.
Refer to the supplementary document for more details about our implementation including hyperparameter setting and results from ablation studies.

\begin{table*}[t!]
\caption{
Results on three benchmarks with two different federated learning settings. 
For (a) a moderate-scale experiment, the number of clients and the participation rate, are set to 100, and 5\%, respectively, while (b) a large-scale setting has 500 clients with a 2\% participation rate. The Dirichlet parameter is commonly set to 0.3.
Accuracies at the target round and the communication round to reach target test accuracy are based on exponential moving averages with parameter $0.9$. 
The arrows indicate whether higher ($\uparrow$) or lower ($\downarrow$) is better. 
FedCM$^\dagger$ and FedDC$^\ddagger$ require 50\% and 100\% additional communication costs at each communication round, respectively.}
\label{tab:dir03}
\begin{center}
\vspace{-4mm}
\begin{subtable}[t]{1\textwidth}
\centering
\caption{Moderate-scale: 100 clients, 5\% participation}
\scalebox{0.85}{
\setlength\tabcolsep{8.7pt}
\hspace{-0.1cm}
\begin{tabular}{lcccccccccccc} 
\toprule
\multirow{3}{*}{Method}    & \multicolumn{4}{c}{CIFAR-10}                                                                     & \multicolumn{4}{c}{CIFAR-100}                                                                & \multicolumn{4}{c}{Tiny-ImageNet}                                                                                                 \\ 
\cmidrule(lr){2-5} \cmidrule(lr){6-9} \cmidrule(lr){10-13}
& \multicolumn{2}{c}{Acc. (\%, $\uparrow$)} & \multicolumn{2}{c}{Rounds ($\downarrow$)}  & \multicolumn{2}{c}{Acc. (\%, $\uparrow$)} & \multicolumn{2}{c}{Rounds ($\downarrow$)} & \multicolumn{2}{c}{Acc. (\%, $\uparrow$)} & \multicolumn{2}{c}{Rounds ($\downarrow$)}  \\
& \multicolumn{1}{c}{500R} & \multicolumn{1}{c}{1000R} & 81\% & 85\%                            & 500R           & \multicolumn{1}{c}{1000R}    & 47\%         & 55\%                           & \multicolumn{1}{c}{500R} & \multicolumn{1}{c}{1000R}            & \multicolumn{1}{c}{35\%} & \multicolumn{1}{c}{38\%}                     \\ 
\midrule
FedAvg~\cite{mcmahan2017communication}  & 74.36 & 82.53 & 840 & 1000+ & 41.88  & 47.83  & 924 & 1000+ & 33.94 & 35.37 & 645 & 1000+ \\
FedProx~\cite{li2020federated}        & 73.70 & 82.68 & 826 & 1000+ & 42.43  & 48.32  & 881 & 1000+ & 34.14 & 35.53 & 613 & 1000+ \\
FedAvgM~\cite{hsu2019measuring}      & 80.56 & 85.48 & 519 & 828 & 46.98  & 53.29  & 515 & 1000+ & 36.32 & 38.51 & 416 & 829 \\
FedADAM~\cite{reddi2021adaptive}        & 72.33 & 81.73 & 908 & 1000+ & 44.80  & 52.48  & 691 & 1000+ & 33.22 & 38.91 & 658 & 945 \\
FedDyn~\cite{acar2021federated}       &84.82 & 88.10& {392} & 646  & 48.38 &55.79 &424 &883 & 37.35& 41.18& 344& 573  \\
MOON~\cite{li2021model}            & 83.32 & 86.30 & {371} & 686 & 53.15 & 58.37 & 284 & 640 & 36.62 & 40.33 & 410 & 627\\ 
\rowcolor{Gray}
FedCM$^\dagger$~\cite{xu2021fedcm}            & 78.92 & 83.71 & 624 & 1000+ & 52.44  & 58.06  & 293 & 747 & 31.61 & 37.87 & 694 & 1000+\\
FedMLB~\cite{kim2022multi} & 74.98 & 84.04 & 714 & 1000+ & 47.39 & 54.58 & 490 & 1000+ & 37.20 & 40.16 & 415 & 539 \\
FedLC~\cite{zhang2022Federated} & 78.37 & 84.79 & 680 & 1000+ & 42.74 & 47.23 & 980 & 1000+ & 35.03  & 35.95 & 500 & 1000+ \\
FedNTD~\cite{lee2022preservation} & 76.05 & 83.78 & 685 & 1000+ & 43.22 & 49.29 & 721 & 1000+ & 33.91 & 37.33 & 547 & 1000+ \\
\rowcolor{Gray}
FedDC$^{\ddagger}$~\cite{gao2022feddc}          & {\bf86.52} & 87.47 & {323} & 519 & 54.25 & 59.01 & 333 & 553 & 40.32 & 45.51 & 340 &  403\\ 
FedDecorr~\cite{shi2022towards} & 76.62 & 83.40 & 728 & 1000+ & 43.52 & 49.17 & 767 & 1000+ & 33.40 & 34.86 & 1000+ & 1000+ \\
\textbf{FedACG (ours)}                       & 85.13 & {\bf89.10} & {{\bf319}} & {\bf450} & {\bf55.79}  & {\bf62.51}  & {\bf260} & {\bf409} & {\bf42.26}& \bf{46.31} & {\bf226} & {\bf331}\\
\bottomrule
\label{tab:dir03_moderate}
\end{tabular}}
\end{subtable}
\end{center}
\vspace{-8mm}
%
%
\begin{center}
\begin{subtable}[t]{1\textwidth}
\centering
\caption{Large-scale: 500 clients, 2\% participation}
\scalebox{0.85}{
\setlength\tabcolsep{8.6pt}
\hspace{-0.3cm}
\begin{tabular}{lcccccccccccc} 
\toprule
\multirow{3}{*}{Method}  & \multicolumn{4}{c}{CIFAR-10} & \multicolumn{4}{c}{CIFAR-100} & \multicolumn{4}{c}{Tiny-ImageNet} \\ 
\cmidrule(lr){2-5} \cmidrule(lr){6-9} \cmidrule(lr){10-13}
 & \multicolumn{2}{c}{Acc. (\%, $\uparrow$)} & \multicolumn{2}{c}{Rounds ($\downarrow$)} & \multicolumn{2}{c}{Acc. (\%, $\uparrow$)} & \multicolumn{2}{c}{Rounds ($\downarrow$)} & \multicolumn{2}{c}{Acc. (\%, $\uparrow$)} & \multicolumn{2}{c}{Rounds ($\downarrow$)}  \\
                                          & 500R           & \multicolumn{1}{c}{1000R}     & 73\%         & 77\%                            & 500R           & \multicolumn{1}{c}{1000R}    & 36\%         & 40\%                           & \multicolumn{1}{c}{500R} & \multicolumn{1}{c}{1000R}            & 24\%         & 30\%                     \\ 
\midrule
FedAvg~\cite{mcmahan2017communication} & 58.74 & 71.45 & 1000+ & 1000+ & 30.16& 38.11& 842& 1000+ & 23.63 &29.48 & 523  & 1000+ \\
FedProx~\cite{li2020federated} & 57.88 & 70.75 & 1000+ & 1000+ & 29.28 & 36.16 & 966 & 1000+ &25.45  &31.71 &445 &799 \\
FedAvgM~\cite{hsu2019measuring}  & 65.85 & 77.49 & 753  & 959 & 31.80 & 40.54 & 724 & 955 & 26.75 & 33.26 & 386 & 687 \\
FedADAM~\cite{reddi2021adaptive}   & 61.53 & 69.94 & 1000+ & 1000+  & 24.56 & 34.36 & 1000+ & 1000+ &21.88 &28.08  & 648& 1000+ \\
FedDyn~\cite{acar2021federated}   & 65.49 & 77.92 & 732 & 936  & 31.58 & 41.01 & 691  & 927 & 24.35 & 29.54 & 483 & 1000+ \\
MOON~\cite{li2021model}           & 69.15 & 78.06 & 617 & 872 & 33.51 & 42.41 & 601 & 828 & 26.69 & 31.81 & 382 & 741 \\ 
\rowcolor{Gray}
FedCM$^\dagger$~\cite{xu2021fedcm}  & 69.27  & 76.57 & 742  & 1000+ & 27.23 & 38.79 & 872 & 1000+ &19.41 &24.09 & 975 & 1000+ \\ 
FedMLB~\cite{kim2022multi} & 58.68 & 71.38 & 1000+  & 1000+ & 32.30 & 42.61 & 643 & 803 & 28.39 & 33.67 & 382 & 579 \\
FedLC~\cite{zhang2022Federated} & 60.16 & 70.10 & 1000+ & 1000+ & 29.58 & 36.78 & 936 & 1000+ & 22.14 & 26.83 & 676 & 1000+ \\
FedNTD~\cite{lee2022preservation} & 60.65 & 73.20 & 991 &  1000+ & 28.95 & 36.31 & 995 & 1000+ & 24.67 & 32.16 & 475 & 800 \\
\rowcolor{Gray}
FedDC$^\ddagger$~\cite{gao2022feddc}           & 71.86 & {\bf83.49} & 518 & 686 & 34.64 & 45.93 & 569 & 741 & 25.72 & 28.92 & 420 & 1000+ \\ 
FedDecorr~\cite{shi2022towards} & 60.01 & 72.83 & 1000+ & 1000+ & 30.56 & 38.20 & 850 & 1000+ & 24.34 & 30.28 & 499 & 959 \\
\textbf{FedACG (ours)}       & \textbf{73.61} & 82.80 & \textbf{484} & \textbf{605} & \textbf{35.68} & \textbf{48.40}  & \textbf{505} & \textbf{616} &\textbf{31.47} &\textbf{38.48} & \textbf{246} & \textbf{447}\\
\bottomrule
\label{tab:dir03_large}
\end{tabular}}
\end{subtable}
\end{center}
\vspace{-1.0cm}
\end{table*}

\subsection{Experimental setup}
\label{sec:setup}

\paragraph{Datasets}
We conduct a set of experiments on three datasets, CIFAR-10~\citep{krizhevsky2009learning}, CIFAR-100~\citep{krizhevsky2009learning}, and Tiny-ImageNet~\citep{le2015tiny}, with different levels of data heterogeneity and participation rates.
We generate \textit{i.i.d.}~data splits by randomly assigning training examples to individual clients without replacement.
For non-\textit{i.i.d.}~datasets, we simulate the data heterogeneity by sampling the label ratios from a Dirichlet distribution with a symmetric parameter, 0.3 or 0.6, following the strategies in~\citet{hsu2019measuring}.
In both \textit{i.i.d.}~and non-\textit{i.i.d.}~cases, each client holds the same number of examples as in other works~\cite{xu2021fedcm,kim2022multi}.
We extend our experiments to the widely adopted FL benchmark, LEAF~\citep{caldas2018leaf}, known for its realistic settings.
LEAF introduces heterogeneity in class distribution, data quantity, and feature alignment.
For FEMNIST, CelebA, and ShakeSpeare, we use the non-\textit{i.i.d.} data splits provided by LEAF.

\paragraph{Baselines}
We compare our method, FedACG, with state-of-the-art federated learning techniques, which include FedAvg~\citep{mcmahan2017communication}, FedProx~\citep{li2020federated}, FedAvgM~\citep{hsu2019measuring}, FedADAM~\citep{reddi2021adaptive}, FedDyn~\citep{acar2021federated}, FedCM~\citep{xu2021fedcm}, MOON~\citep{li2021model},
FedMLB~\cite{kim2022multi},
FedNTD~\cite{lee2022preservation},
FedLC~\cite{zhang2022Federated},
FedDC~\citep{gao2022feddc}, and FedDecorr~\cite{shi2022towards}.
The standard ResNet-18~\citep{he2016deep} is employed as our backbone network for all experiments after replacing the batch normalization with the group normalization, following~\citet{hsieh2020non}.

\paragraph{Evaluation metrics}
To evaluate the generalization performance of the methods, we use the entire test set of CIFAR-10, CIFAR-100, and Tiny-ImageNet.
Since both the training speed as well as the final accuracy are important factors in federated learning, we measure: (i) the performance achieved at a specified number of rounds and (ii) the number of rounds required for an algorithm to attain the desired level of target accuracy, following Al-Shedivat~\etal~\cite{al2020federated}.
For the methods that fail to achieve the target accuracies within the maximum communication round, we append a $+$ sign to the communication round number.

\begin{table*}[t]
\caption{Results from reduced participation rates (2\% for 100 clients, 1\% for 500 clients) on CIFAR-10 and CIFAR-100 with the Dirichlet parameter 0.3.
FedCM$^\dagger$ and FedDC$^\ddagger$ require 50\% and 100\% additional communication costs for each communication round, respectively.
}
\vspace{-4mm}
\begin{center}
\label{tab:abl_500cl_1partrate}
\scalebox{0.85}{
\setlength\tabcolsep{5.5pt}
\hspace{-0.27cm}
\begin{tabular}{lcccccccccccc} 
\toprule
   \multirow{4}{*}{Method}     & \multicolumn{6}{c}{CIFAR-10}               & \multicolumn{6}{c}{CIFAR-100}                                                                                                                                                                \\ 
\cmidrule(lr){2-7} \cmidrule(lr){8-13}
& \multicolumn{3}{c}{100 clients } &  \multicolumn{3}{c}{500 clients} & \multicolumn{3}{c}{100 clients} &  \multicolumn{3}{c}{500 clients} \\
 & \multicolumn{2}{c}{Acc. (\%, $\uparrow$)} & \multicolumn{1}{c}{Rounds ($\downarrow$)}  & \multicolumn{2}{c}{Acc. (\%, $\uparrow$)} & \multicolumn{1}{c}{Rounds ($\downarrow$)} 
 & \multicolumn{2}{c}{Acc. (\%, $\uparrow$)} & \multicolumn{1}{c}{Rounds ($\downarrow$)} & \multicolumn{2}{c}{Acc. (\%, $\uparrow$)} & \multicolumn{1}{c}{Rounds ($\downarrow$)} \\
 & \multicolumn{1}{c}{500R} & \multicolumn{1}{c}{1000R} &  78\%  & 500R           & \multicolumn{1}{c}{1000R}   & 68\%  & \multicolumn{1}{c}{500R} & \multicolumn{1}{c}{1000R} & 44\% & 500R           & \multicolumn{1}{c}{1000R}    & 35\%                 \\ 
\midrule
FedAvg~\cite{mcmahan2017communication} & 65.92 & 78.13 & 977 & 54.71 & 68.96 &  949 & 38.19 & 44.62 & 924 &   26.94 & 35.69   & 950  \\
FedProx~\cite{li2020federated}         & 65.78 & 75.82 & 1000+ & 55.18  & 69.80 &  919 & 36.69 & 45.16 & 921 &  26.92 & 35.41  &  963 \\
FedAvgM~\cite{hsu2019measuring}        &68.09 & 79.91 & 748 & 57.82 & 71.12 & 812  &39.24 & 53.47 & 504 & 29.29 & 39.36  & 755 \\
FedADAM~\cite{reddi2021adaptive}       & 68.09 & 78.61 & 978 & 48.26 & 54.60 &  1000+ & 40.95 & 51.14 & 592 & 18.21 & 23.70   & 1000+ \\
FedDyn~\cite{acar2021federated}        & 74.27 & 80.20 & 660 & 54.86 & 70.78 &  858  & 38.94 & 48.88 & 716 & 27.86 & 36.31 &   896 \\
MOON~\cite{li2021model}                & 71.52 & 75.42 & 1000+ & {\bf64.55} & 73.89  & {645} & 39.91 & 46.51 & 730 & 28.29 & 36.37  & 886 \\ 
\rowcolor{Gray}
FedCM$^\dagger$~\cite{xu2021fedcm}     & 52.45 & 64.50 & 1000+ & 49.21& 60.38 & 1000+ & 14.52 & 23.06 & 1000+ & 16.32 & 22.59  & 1000+  \\ 
FedMLB~\cite{kim2022multi}             & 65.85  & 79.45 & 899 & 52.81 &  66.86& 1000+ & 40.09 & 53.34 & 565 & 29.78 &39.64  &724 \\
FedLC~\citep{zhang2022Federated}       & 72.90 & 80.90 & 736 & 54.89 & 68.31 & 967 & 39.70 & 42.10 & 1000+ & 27.73 & 35.24 & 918 \\
FedNTD~\cite{lee2022preservation}      & 69.11 & 80.43 & 797 & 54.53 & 68.69 & 961  & 38.13 & 48.03 & 708 & 27.56 & 35.86 & 932 \\
\rowcolor{Gray}
FedDC$^\ddagger$~\cite{gao2022feddc}     & {\bf77.76} & 82.86 & {\bf473}  & 60.56 & 75.06 &  {681} & 41.50 & 51.37 & 670 & 29.14 & 38.84  & 789 \\ 
FedDecorr~\citep{shi2022towards}       & 71.29 & 78.99 & 817 & 56.62 & 70.24 & 845 & 39.42 & 48.45 & 718 & 31.03 & 38.70 &  705 \\
\textbf{FedACG (ours)}                         & 76.36 & {\bf84.73} & 543 & 63.70 & {\bf76.45}  & {\bf618} & {\bf49.56} & {\bf56.89} & {\bf358} & {\bf 31.74} & {\bf45.18} & {\bf581} \\
\bottomrule
\end{tabular}}
\end{center}
\vspace{-4mm}
\end{table*}

\begin{table}[t!]
\centering
\captionof{table}{
Results on the Dirichlet (0.3) split of CIFAR-100 with dynamic client updates during training: we maintain 250 clients but each client is replaced with a probability of 0.5 at every 100 rounds.
The experiment runs for 10 stages and the client participation ratio is 4\%.
}
\vspace{-0.1cm}
\centering
\scalebox{0.85}{
\setlength\tabcolsep{11.5pt}
\hspace{-0.2cm}
\begin{tabular}{lcccc}
\toprule
\multirow{2}{*}{Method} & \multicolumn{2}{c}{Acc. ($\%$, $\uparrow$)} & \multicolumn{2}{c}{Rounds ( $\downarrow$)}\\
& {500R } & {1000R } & {30$\%$} & {38$\%$}\\
\midrule
FedAvg~\cite{mcmahan2017communication} & 28.61 & 35.87 & 577 & 1000+ \\
FedProx~\cite{li2020federated} & 28.17 & 35.89 & 602 & 1000+ \\
FedDyn~\cite{acar2021federated}        & 29.45 & 38.47 & 517 & 941 \\
MOON~\cite{li2021model}               & 30.88 & 39.57 & 430 & 852 \\
FedNTD~\cite{lee2022preservation}     & 28.45 & 36.26 & 578 & 1000+ \\
\rowcolor{Gray}
FedDC$^\ddagger$~\cite{gao2022feddc}    & 31.35 & 36.82 & 469 & 1000+ \\

\textbf{FedACG (ours)}                           & {\bf32.70} & {\bf41.51} & {\bf376} & {\bf769} \\
\bottomrule
\end{tabular}
}
\label{tab:incre}
\end{table}
%


\subsection{Main results}


We first present the performance of the proposed approach, FedACG, on CIFAR-10, CIFAR-100, and Tiny-ImageNet by varying the number of clients, data heterogeneity, and participation rate.
Our experiments have been performed in two different settings; one is a moderate scale, which involves 100 clients with a 5\% participation rate per round, and the other is with a large number of clients, 500 with a participation rate of 2\%.
Because the number of clients in the large-scale setting is five times higher than that in the moderate-scale experiment, the number of examples per client is reduced by 80\%.

Table~\ref{tab:dir03_moderate} demonstrates that FedACG improves accuracy and convergence speed significantly and consistently compared with other federated learning methods in most cases.
This is partly because FedACG allows each client to look ahead in the direction of the potential global update and aligns the local model updates with the global gradient trajectory. 
Note that FedCM and FedDC respectively require $1.5 \times$ and $2 \times$ network costs for each communication round since they communicate the current model and the associated gradient information per round, while the rest of the algorithms only need to transmit model parameters.

For the large-scale setting, Table~\ref{tab:dir03_large} illustrates the outstanding performance in the three benchmarks, except for the accuracy at 1K rounds on CIFAR-10.
A noticeable observation is that the overall performance is lower than the case with a moderate number of clients. 
This is because the number of training data for each client decreases and each client suffers more from heterogeneous data distributions.
Nevertheless, we observe that FedACG outperforms other methods consistently in most cases; the accuracy gap between FedACG and its strongest competitor becomes larger in these more challenging scenarios.
The results from the large-scale experiments exhibit the robustness of FedACG to the data heterogeneity and low client participation rates.
We present more comprehensive results for the convergence of FedACG in the supplementary document.

\begin{figure*}[!t]
\centering
\begin{subfigure}[b]{\linewidth}
\centering
\includegraphics[width=0.86\linewidth]{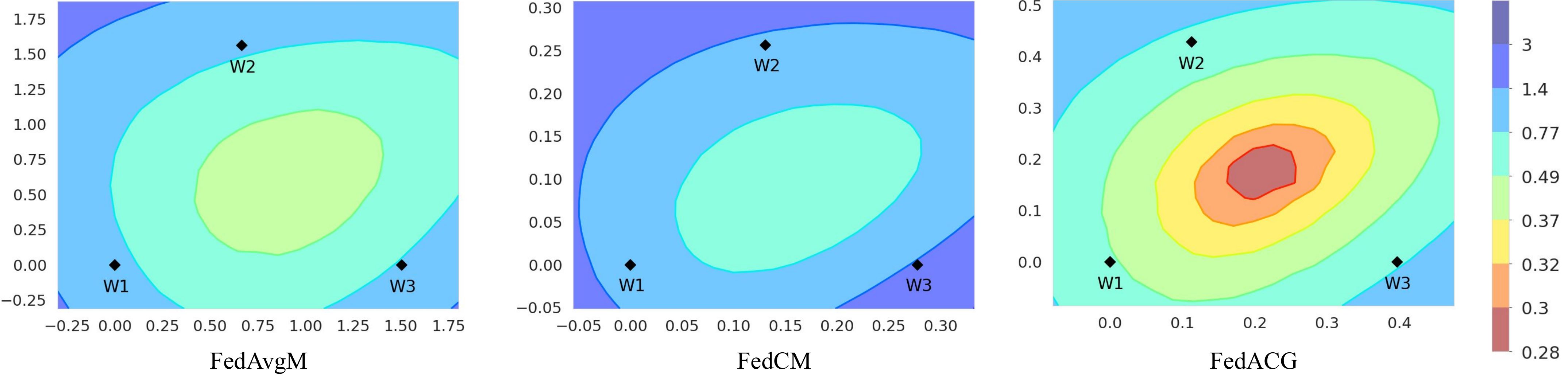}
\caption{Visualization of global training loss surfaces}
\label{fig:loss_surface}
\end{subfigure}
\centering
\begin{subfigure}[b]{0.41\linewidth}
\centering
\includegraphics[width=1\linewidth]{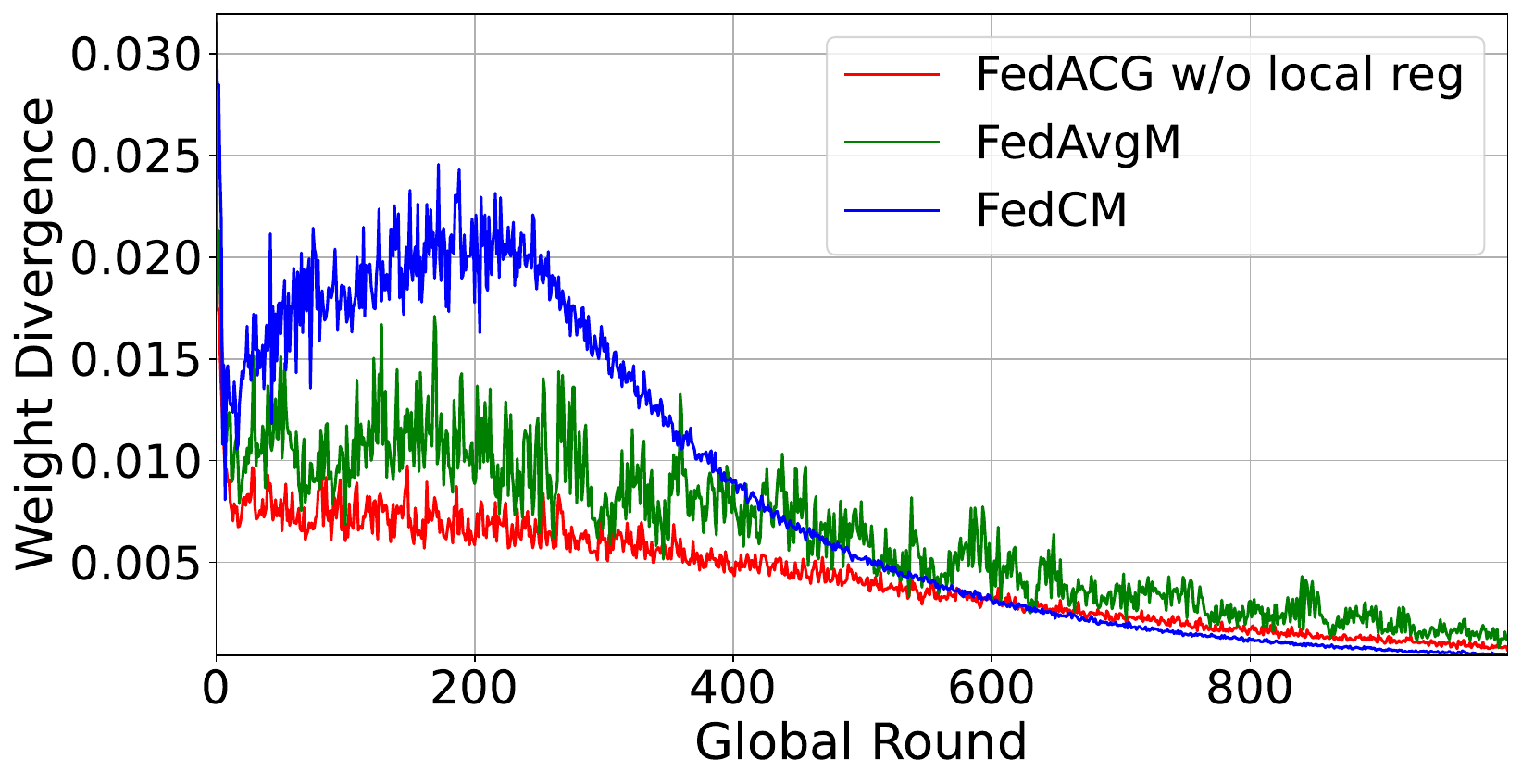}
\caption{Weight divergence}
\label{fig:weight_divergence}
\end{subfigure}
\hspace{0.5cm}
\begin{subfigure}[b]{0.41\linewidth}
\centering
\includegraphics[width=1\linewidth]{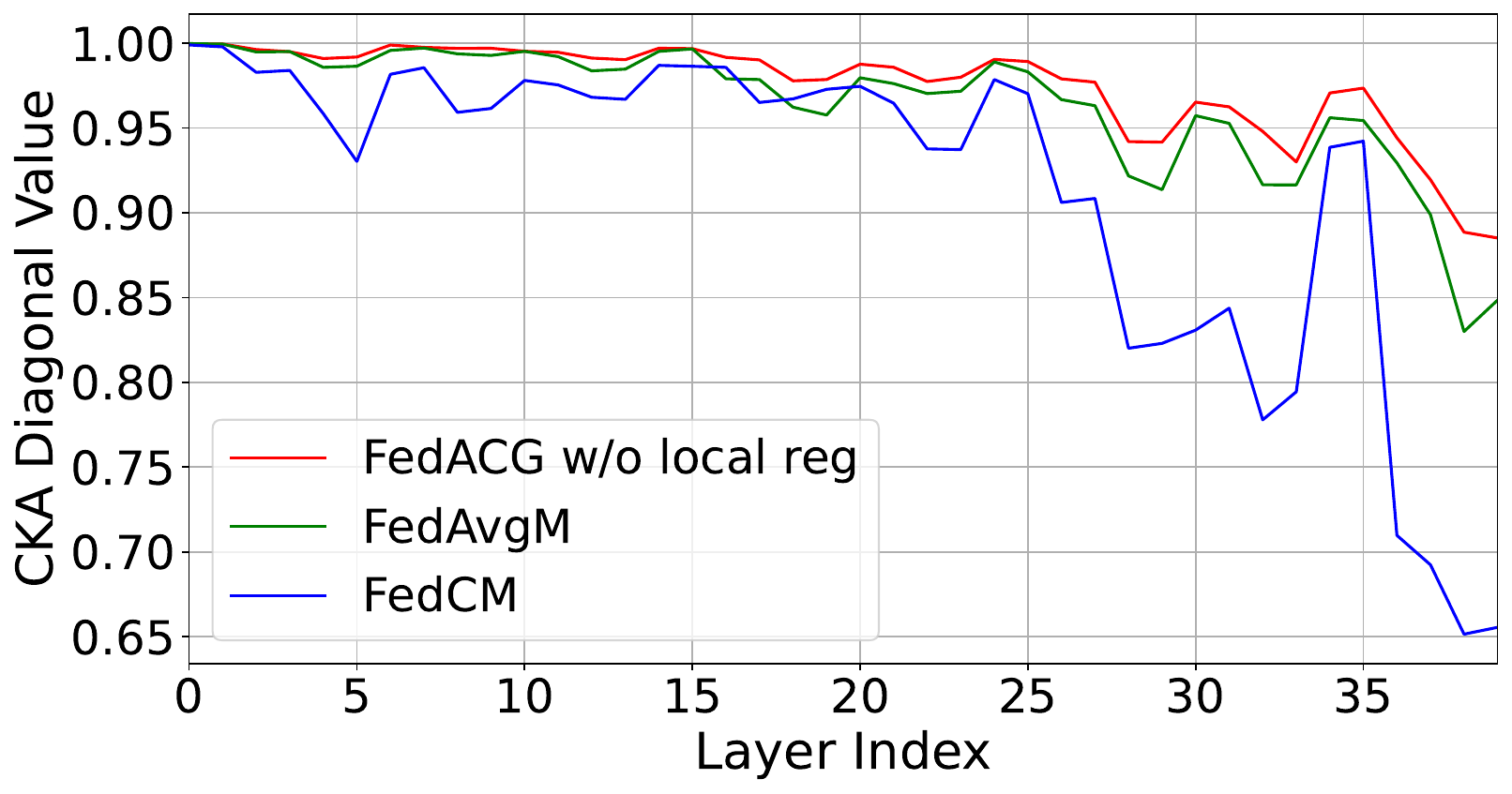}
\caption{Layer-wise CKA values}
\label{fig:cka}
\end{subfigure}
\caption{
\textbf{Benefit of accelerated client gradient.} 
For FedAvgM, FedCM, and FedACG (without local regularization for fair comparisons) on CIFAR10, we visualize (a) global training loss surfaces with three local models as black circles in the parameter space, (b) weight divergence, and (c) layer-wise CKA values. In (c), the $x$-axis denotes the layer index of ResNet-18 while the $y$-axis corresponds to CKA values measured on the global validation set.
}
\vspace{-0.2cm}
\label{fig:acg_analysis}
\end{figure*}

\begin{table}[t!]
\centering
\caption{Contribution of individual components in FedACG. The results are measured after $1K$ rounds on CIFAR-10 and CIFAR-100 with 2\% participation rate over 500 clients.}
\centering
    \scalebox{0.85}{
\setlength\tabcolsep{6pt}
 \hspace{-0.2cm}
    \begin{tabular}{ccccc}
  \toprule
	   \makecell{Server update \\w/ momentum} & \makecell{Accelerated\\gradient} & \makecell{Local\\reg.} & CIFAR-10 & CIFAR-100 \\
  \midrule
         & &  & 71.45 & 38.11  \\ 
         & & \checkmark &  70.75 & 36.16\\ 
        \checkmark &  &  & 77.49 & 40.54  \\ 
        \checkmark & & \checkmark & 76.16 & 44.64  \\ 
         \checkmark & \checkmark &  & 82.20  & 46.80\\ 
         \checkmark & \checkmark & \checkmark & \textbf{82.80}  & \textbf{48.40}  \\                     
        \bottomrule
	\end{tabular}
    }
\label{tab:accel_ablation}
\end{table}

\subsection{Analysis}
\label{sec:ablation}

\paragraph{Effect of low participation rate}

One of the critical challenges in federated learning is the partial participation of clients, which can slow down the convergence of the global model. 
To verify the robustness of FedACG to low client participation rates, we conduct experiments with 500 clients and a participation rate as low as 1\%. 
Since the numbers of local epochs and iterations are set to 5 and 50, respectively, each client has little training opportunity with few training examples and client heterogeneity increases significantly. 
As shown in Table~\ref{tab:abl_500cl_1partrate}, FedACG outperforms the other methods in most cases, with the performance gap between FedACG and the second best method, FedDC, being even more significant than when the participation rate is 2\% with 500 clients, from -0.69\%p to 1.39\%p on CIFAR-10 and from 2.47\%p to 6.34\%p on CIFAR-100 at round 1000. 
This is partly because the local states in FedDC become stale quickly in this scenario, requiring extra iterations for convergence, whereas FedACG is not affected by this issue.

\paragraph{Evaluation on dynamic client set}

Since FedACG does not require storing local model history for local updates, it is conceptually better suited for scenarios with newly participating clients. 
To validate this property, we conduct an experiment, where we maintain 250 clients in each round but replace half of the clients on average every 100 rounds by setting the replacement probability of each client to 0.5. 
The experiment has been performed on CIFAR-100 with Dirichlet (0.3) splits, assuming a participation ratio of 4\% for each communication round. 
Table~\ref{tab:incre} shows that FedACG outperforms FedAvg and FedDyn. 
Note that FedDyn performs worse than FedAvg since the client models suffer from heterogeneity and divergence when new clients have no informative local states.

\paragraph{Ablation study}
Table~\ref{tab:accel_ablation} presents the contributions of individual components in the experiment on CIFAR-10 for the large-scale setting. 
We observe that the accelerated client gradient for local training makes a more critical impact on accuracy after 1,000 rounds. 
It is worth noting that the proposed regularization term in the local loss function shows a larger performance gain when used with the accelerated client gradient while employing the regularization term alone does not necessarily lead to a beneficial outcome on CIFAR-10 and CIFAR-100.

\paragraph{Comparison with FedAvgM and FedCM}

To better understand the effectiveness of the accelerated client gradient, we compare two momentum-based algorithms, FedAvgM and FedCM, by visualizing global loss surfaces, weight divergence, and layer-wise CKA values during training.
Figure~\ref{fig:loss_surface} highlights a better generalization of FedACG's local models to global loss compared to other methods.
Figures~\ref{fig:weight_divergence} and \ref{fig:cka} reveal that the local models of FedACG exhibit less divergence in the parameter space and more consistent feature representations, respectively.
These findings demonstrate that the accelerated client gradient in FedACG effectively mitigates client drift stemming from data heterogeneity.

\paragraph{Hyperparameters}
We test the accuracy of our algorithm for the Dirichlet (0.3) and \textit{i.i.d.}~splits by varying the values of $\lambda$ and $\beta$, which control the momentum integration of the server model and the weight of the proximal term, respectively.
As shown in Table~\ref{tab:lamb_ablation}, the performance of FedACG remains stable in a range of $\lambda$ values from 0.75 to 0.9.
Despite minor fluctuations, the accuracy remains high, peaking at $\lambda = 0.85$.
Table~\ref{tab:lamb_ablation} also shows that the accuracy is stable with respect to $\beta$.

\paragraph{Integration into quantization approaches}
Our framework is orthogonal to quantization-based FL algorithms, which enables seamless integration with such approaches.
Table~\ref{tab:fedpaq} presents consistent and remarkable improvements when combining FedACG with quantization-based algorithms, such as FedPAQ~\cite{reisizadeh2020fedpaq}.

\begin{table}[t]
\centering
\caption{Ablation study for \iffalse for the two hyperparameters, \fi $\lambda$ (a) and $\beta$ (b), w.r.t the accuracy at $1K^\text{th}$ round on CIFAR-10 with 2\% participation and 500 clients.}
\begin{subtable}[t]{1\linewidth}
\centering
\caption{Sensitivity of FedACG with respect to $\lambda$}
\scalebox{0.85}{
\setlength\tabcolsep{8.5pt} \hspace{-0.2cm}
\begin{tabular}{ccccc}
\toprule
$\lambda$          & 0.75 & 0.8 & 0.85 & 0.9  \\
\midrule
Dir(0.3)  & 81.32 & 82.52 & 82.80& 82.64 \\
\textit{i.i.d.}             & 85.52 & 86.82 & 86.83 & 86.16 \\
\bottomrule
\label{tab:lamb_ablation}
\end{tabular}}
\end{subtable}
\begin{subtable}[t]{1\linewidth}
\centering
\caption{Sensitivity of FedACG with respect to $\beta$}
\scalebox{0.85}{
\setlength\tabcolsep{8.5pt} \hspace{-0.2cm}
\begin{tabular}{ccccc}
\toprule
$\beta$       & 0.001 & 0.01 & 0.1 & 1\\
\midrule
Dir(0.3) & 82.10 & 82.80 & 82.32 & 82.44\\
\textit{i.i.d.}             & 86.54 & 86.83 & 86.72 & 85.92\\
\bottomrule
\label{tab:beta_ablation}
\end{tabular}}
\end{subtable}
\vspace{-2mm}
\end{table}
\begin{table}[t]
    \centering
    \caption{Integration of FedACG into quantization-based federated learning approach under non-\textit{i.i.d.}~settings on CIFAR-100; 5\% participation rate for 100 clients and 2\% for 500 clients.
    The Dirichlet parameter is commonly set to 0.3.
    }
    \scalebox{0.85}{
    \setlength\tabcolsep{10pt}
    \hspace{-0.25cm}
    \begin{tabular}{lcccc}
        \toprule
        \multirow{2}{*}{Method} & \multicolumn{2}{c}{100 clients} & \multicolumn{2}{c}{500 clients} \\
        & 500R & 1000R & 500R & 1000R \\
        \midrule
        FedAvg~\cite{mcmahan2017communication} & 41.88 & 47.83 & 30.16 & 38.11 \\
        FedPAQ~\cite{reisizadeh2020fedpaq} &36.57 & 41.99 & 23.80& 30.21 \\
        FedACG + FedPAQ &\textbf{43.04} & \textbf{50.53} & \textbf{31.94}& \textbf{39.55} \\
        \bottomrule
    \end{tabular}}
    \label{tab:fedpaq}
\end{table}

\subsection{Experiments on realistic datasets}

We conduct experiments on additional realistic datasets, FEMNIST and CelebA in LEAF~\citep{caldas2018leaf}, which include other kinds of non-\textit{i.i.d.}~scenarios such as feature skewness and data imbalance between clients.
For these experiments, we set the number of clients to 2,000, with data splits following~\cite{caldas2018leaf}, and randomly select 5 clients to participate in training during each communication round.
We employ a two-layer CNN for FEMNIST and a four-layer CNN for CelebA.
Table~\ref{tab:leaf} illustrates that FedACG also outperforms other baselines on both datasets for most cases, highlighting its robustness to heterogeneity with data quantity and feature alignment.
Note that, while FedACG requires 20 more communication rounds than FedDC to reach the target accuracy on FEMNIST, it sends 56.8\% fewer parameters than FedDC.

We also evaluate FedACG in a different domain, next word prediction task, on the ShakeSpeare in LEAF, which also involves a significant data imbalance between clients.
We adopt an LSTM as the backbone network, and the client participation rate per round is set to 5\%.
Table~\ref{tab:leaf} presents that FedACG is also effective for the language domain, while FedCM exhibits poor performance even with extensive hyperparameter tuning.

\begin{table}[t]
\caption{
Results on the realistic datasets involving feature skewness and data imbalance between clients.
FedCM$^\dagger$ and FedDC$^\ddagger$ require 50\% and 100\% additional communication costs per communication round, respectively.
}
\vspace{-2mm}
\label{tab:leaf}
\begin{center}
\scalebox{0.85}{
\setlength\tabcolsep{3pt}
\hspace{-0.3cm}
\begin{tabular}{lcccc}

\toprule
\multirow{3}{*}{Method}                                   & \multicolumn{2}{c}{FEMNIST}                                                                     & \multicolumn{2}{c}{CelebA}                                                                
\\ 
\cmidrule(lr){2-3} \cmidrule(lr){4-5}
& \multicolumn{1}{c}{Acc. (\%, $\uparrow$)} & \multicolumn{1}{c}{Rounds ($\downarrow$)} & \multicolumn{1}{c}{Acc. (\%, $\uparrow$)} & \multicolumn{1}{c}{Rounds ($\downarrow$)} 
\\
                                                   & \multicolumn{1}{c}{500R}     & 78\%                                             & \multicolumn{1}{c}{500R}            & 88\%                                         \\ 
\midrule
FedAvg~\cite{mcmahan2017communication} & 78.38 & 328 & 89.92 & 134  \\
FedProx~\cite{li2020federated}       & 78.34 & 328 & 89.90 & 132 \\
FedAvgM~\cite{hsu2019measuring}     & 78.37 & 256 & 89.85 & 113 \\
FedADAM~\cite{reddi2021adaptive}     & 75.96 & 500+ & 87.00 & 500+ \\
FedDyn~\cite{acar2021federated}      &79.80 & 227 & 89.74 & 126   \\
MOON~\cite{li2021model}           & 78.33 & 336  & 87.95 & 500+ \\ 
\rowcolor{Gray}
FedCM$^\dagger$~\cite{xu2021fedcm}           & 72.79 & 500+ & 88.89 & 222\\ 
FedNTD~\cite{lee2022preservation}     & 78.42 & 330 & 89.31 & 122 \\
\rowcolor{Gray}
FedDC$^\ddagger$~\cite{gao2022feddc}          & 80.11 & {\bf149} &  88.97 &  126 \\ 
\textbf{FedACG (ours)}                     & {\bf80.61} & 169 &   {\bf90.09} & {\bf108} \\
\bottomrule
\label{tab:leaf}
\end{tabular}}
\end{center}
\vspace{-0.8cm}
\end{table}
%
%
\begin{table}[t]
\caption{
Results in the language domain on the next word prediction task under non-\textit{i.i.d.} setting using the ShakeSpeare dataset.
}
\vspace{-2mm}
\label{tab:shakespeare}
\begin{center}
\scalebox{0.85}{
\setlength\tabcolsep{11.5pt}
\hspace{-0.3cm}
\begin{tabular}{lcccc} 
\toprule

\multirow{2}{*}{Method}& \multicolumn{2}{c}{Acc. (\%, $\uparrow$)}   & \multicolumn{2}{c}{Rounds ($\downarrow$)} 
\\
 & \multicolumn{1}{c}{500R}     & 1000R                                             & \multicolumn{1}{c}{42\%}            & 45\%                                         \\ 
\midrule
FedAvg~\cite{mcmahan2017communication}        & 45.01 & 46.55 & 94 & 500  \\
FedProx~\cite{li2020federated}       & 45.09 & 46.29 & 99 & 477 \\
FedAvgM~\cite{hsu2019measuring}       & 44.63 & 45.91 & 63 & 690 \\
FedADAM~\cite{reddi2021adaptive}       & 44.89 & 44.30 & 68 & 1000+ \\
FedDyn~\cite{acar2021federated}        & 39.23 & 44.10 & 749 & 1000+   \\
MOON~\cite{li2021model}          & 42.02 & 42.65  & 499 & 1000+ \\ 
\rowcolor{Gray}
FedCM$^\dagger$~\cite{xu2021fedcm}           & -- & -- & -- & --\\ 
FedNTD~\cite{lee2022preservation}     & 45.01 & 46.5 & 94 & 513 \\
\rowcolor{Gray}
FedDC$^\ddagger$~\cite{gao2022feddc}          & 30.62 & 44.27 &  926 &  1000+ \\ 
\textbf{FedACG (ours)}                     & {\bf46.36} & {\bf48.23} &   {\bf57} & {\bf290} \\
\bottomrule
\end{tabular}}
\end{center}
\vspace{-0.4cm}
\end{table}

%% file: sections/conclusion.tex

\section{Conclusion} 
\label{sec:conclusion}

This paper addresses a realistic federated learning scenario, where a large number of clients with heterogeneous data and limited participation constraints hinder the convergence and performance of trained models. 
To tackle these issues, we proposed a novel federated learning framework that aggregates past global gradient information for guiding client updates and regularizes the local update directions aligned with the global information.
The proposed algorithm provides global gradient information to individual clients without incurring additional communication or memory overhead.
We made a formal proof of the convergence rate of the proposed approach.
FedACG demonstrates the effectiveness in terms of robustness and communication efficiency in the presence of client heterogeneity through extensive evaluation on multiple benchmarks.

%% file: sections/acknowledgments.tex

\vspace{-0.2cm}
\paragraph{Acknowledgments}
\label{Acknowledgments}


This work was partly supported by Samsung Advanced Institute of Technology (SAIT), and by the National Research Foundation of Korea grant [No.2022R1A2C3012210] and the Institute for Information \& communications Technology Planning \& Evaluation (IITP) grants [2022-0-00959; 2021-0-02068; 2021-0-01343] funded by the Korea government (MSIT).

%% file: supple/supple.tex




%
\newpage
\onecolumn
\definecolor{cvprblue}{rgb}{0.21,0.49,0.74}

\renewcommand\thesection{\Alph{section}}
\renewcommand\thetable{\Alph{table}}
\renewcommand\thefigure{\Alph{figure}}
\newcommand\numberthis{\addtocounter{equation}{1}\tag{\theequation}}
\appendix
\onecolumn
\maketitle

\setcounter{table}{0}
\setcounter{figure}{0}
\setcounter{theorem}{0}
\section{Experimental Setup}
\label{sec:exp_setup}

\paragraph{Implementation details}
We adopt PyTorch~\citep{paszke2019pytorch} to implement FedACG and the other baselines.
We follow the evaluation protocol of~\citep{acar2021federated} and~\citep{xu2021fedcm}.
For local updates, we use the SGD optimizer with a learning rate of $0.1$ for all approaches on the three benchmarks.
We apply no momentum to the local SGD, but incorporate the weight decay of $0.001$ to prevent overfitting.
We also employ gradient clipping to increase the stability of the algorithms.

For the experiments on CIFAR-10 and CIFAR-100, we choose 5 as the number of local training epochs (50 iterations).
We set the batch size of the local update to 50 and 10 for the 100 and 500 client participation, respectively.
The learning rate decay parameter of each algorithm is selected from \{0.995, 0.998, 1\} to achieve the best performance.
The global learning rate is set to 1, except for FedAdam, which is set to 0.01. 

For the experiments on Tiny-ImageNet, we match the total local iterations of local updates with other benchmarks by setting the batch size of local updates as 100 and 20 for the 100 and 500 client participation, respectively.

\paragraph{Hyperparameter selection}

To reproduce other compared algorithms, we primarily follow the configurations outlined in the original papers, adjusting the parameters only when it leads to improved performance.
Specifically, $\alpha$ is chosen from \{0.1, 0.3, 0.5\} in FedCM, \{0.001, 0.01, 0.1\} in FedDyn, and is set to 0.01 in FedDC. 
$\tau$ in FedADAM is fixed at 0.001, while $\mu$ in MOON is set to 1. For $\beta$, in FedAvgM, choices are from \{0.4, 0.6, 0.8\}; in FedProx and FedACG, from \{0.1, 0.01, 0.001\}.
In FedProx, FedNTD, and FedDecorr, $\beta$ is set to 0.001, 0.3, and 0.01, respectively. Finally, $\lambda$ in FedACG is selected from \{0.8, 0.85, 0.9\}.

\section{Additional Experiments}

\subsection{Additional analysis for the effect of accelerated client gradient}
\label{sec:additional_ablation}

FedACG uses a lookahead model, $\theta ^{t-1} + \lambda m^{t-1}$, to start local training. 
This helps clients match their local solutions with the global loss, ensuring consistent updates.
We observe more empirical evidence that supports our claim.

\begin{figure}[h!]
\centering
\includegraphics[width=0.45\linewidth]{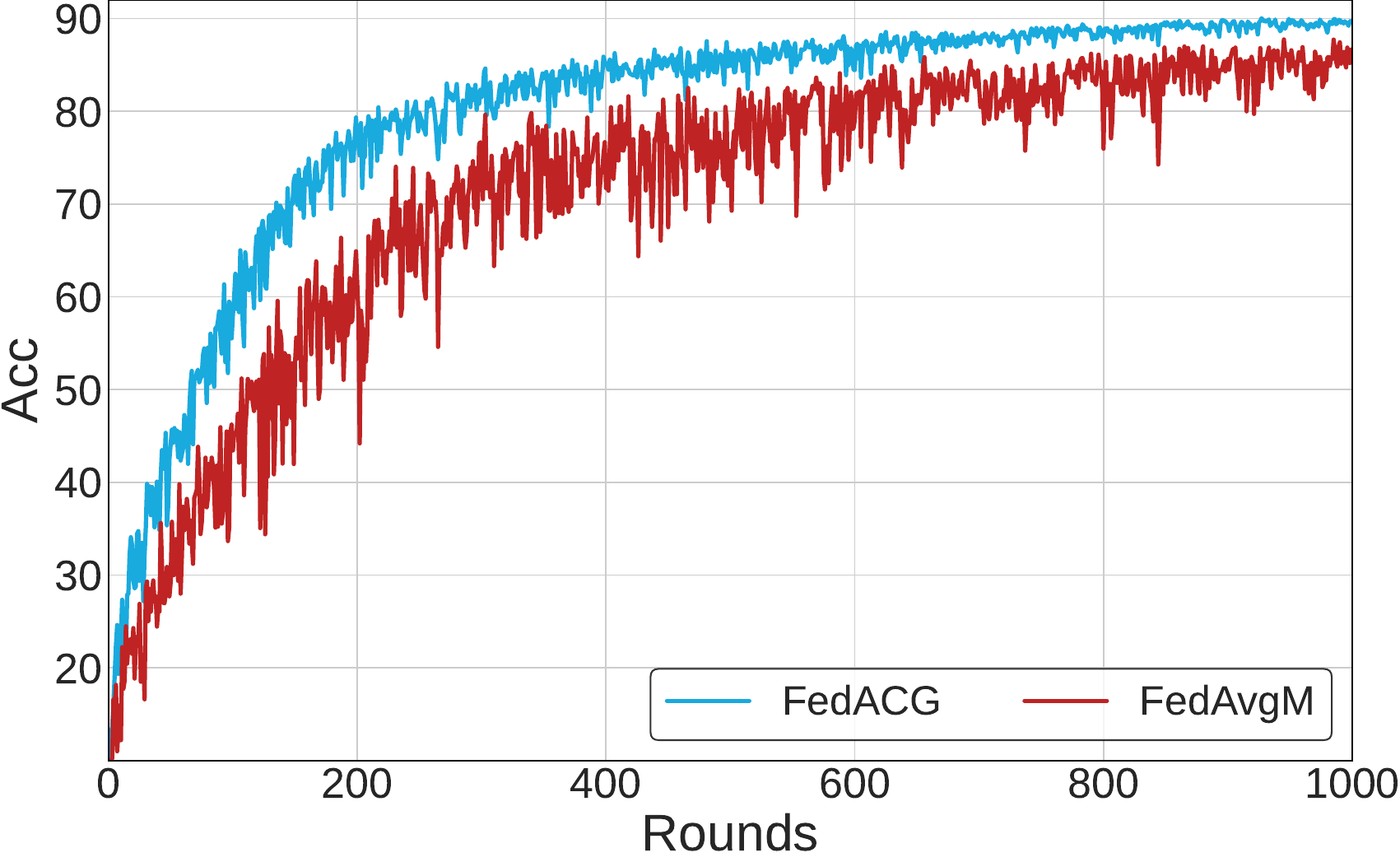}
\caption{Training curves of FedACG and FedAvgM on CIFAR-10 in a moderate-scale setting without smoothing.}
\label{fig:acc_acg_avgM}
\end{figure}

Figure~\ref{fig:acc_acg_avgM} shows the convergence curves of FedACG and FedAvgM on CIFAR-10 in the moderate-scale setting without smoothing. 
For the experiments, we set the momentum coefficient to 0.85 for both algorithms. 
We observe that FedACG consistently outperforms FedAvgM and has a smaller accuracy variation throughout the training procedure.
Specifically, when we compute the average squared difference between the accuracy at time step $t$ without smoothing ($\text{Acc}^t$) and the accuracy given by the simple moving average ($\text{Acc}^t_{\text{SMA}}$) over 1,000 rounds of communication,~\ie, $\frac{1}{T} \sum_{t=0}^{T-1} (\text{Acc}^t-\text{Acc}^t_{\text{SMA}})^2$, the differences are 2.26 and 10.30 for FedACG and FedAvgM, respectively. 
We believe that this is partly because the proposed accelerated gradient allows each client's update to compensate for the potential noise in momentum, which is possible because the local updates start from the anticipated point, $\theta^{t-1} + \lambda m^{t-1}$.

\subsection{FedACG with other local objectives}
In Table~\ref{tab:fedmlb}, we incorporate accelerated client gradient into a client-side optimization technique, FedMLB~\citep{kim2022multi}, FedLC~\citep{zhang2022Federated}, and FedDecorr~\citep{shi2022towards} to test its benefits.
"+ACG" means adopting the proposed accelerated client gradient.
It shows that the momentum-integrated initialization helps client-side optimization approaches achieve significant improvements without any additional communication costs.

\begin{table*}[h!]
\caption{Results of incorporating accelerated client gradient (ACG) into client-side optimization techniques on CIFAR-100 and Tiny-ImageNet under non-\textit{i.i.d.} settings.
}
\label{tab:fedmlb}
\begin{subtable}[t]{1\textwidth}
\begin{center}
\caption{100 clients, 5\% participation, Dirichlet (0.3)}
\scalebox{0.85}{
\begin{tabular}{lcccccccc} 
\toprule
\multirow{3}{*}{Method}    & \multicolumn{4}{c}{CIFAR-100}                                                                     & \multicolumn{4}{c}{Tiny-ImageNet}  \\ 
\cmidrule(lr){2-5} \cmidrule(lr){6-9}
 & \multicolumn{2}{c}{Acc. (\%, $\uparrow$)} & \multicolumn{2}{c}{Rounds ($\downarrow$)} & \multicolumn{2}{c}{Acc. (\%, $\uparrow$)} & \multicolumn{2}{c}{Rounds ($\downarrow$)}  \\
                                          & \multicolumn{1}{c}{500R}           & \multicolumn{1}{c}{1000R}     & 47\%         & 55\%                            & 500R           & \multicolumn{1}{c}{1000R}    & 35\%         & 38\%                                     \\ 
\midrule \midrule
FedMLB~\citep{kim2022multi}                       & {47.39} &  {54.58} & 488 & 1000+ &{37.20} &{40.16} & {414}&  {539}\\
FedMLB + ACG                 & \bf{61.32} &  \bf{65.67} & \bf{216} & \bf{316} &\bf{46.11} &\bf{50.54} & \bf{205}&  \bf{260}\\
\midrule \midrule
FedLC~\cite{zhang2022Federated}                       & 42.74 &  47.23 & 980 & 1000+ & 35.03 & 35.95 &  500 & 1000+ \\
FedLC + ACG               & \bf{57.18} & \bf{62.09}  & \bf{239}   &  \bf{420} & \bf{43.43} & \bf{44.57}  & \bf{187}  &  \bf{268} \\
\midrule \midrule
FedDecorr~\cite{shi2022towards}                        & 43.52 &  49.17 & 767 & 1000+ & 33.40 & 34.86 & 1000+ & 1000+\\
FedDecorr + ACG               & \bf{57.95} & \bf{ 63.02} &\bf{218}   &\bf{380} & \bf{43.09}& \bf{44.52}   &\bf{241}  &    \bf{304}   \\
\bottomrule
\end{tabular}}
\end{center}
\end{subtable}

\begin{center}
\begin{subtable}[t]{1\textwidth}
\centering
\caption{500 clients, 2\% participation, Dirichlet (0.3)}
\scalebox{0.85}{
\begin{tabular}{lcccccccc} 
\toprule
\multirow{3}{*}{Method}    & \multicolumn{4}{c}{CIFAR-100}                                                                     & \multicolumn{4}{c}{Tiny-ImageNet}  \\ 
\cmidrule(lr){2-5} \cmidrule(lr){6-9}
 & \multicolumn{2}{c}{Acc. (\%, $\uparrow$)} & \multicolumn{2}{c}{Rounds ($\downarrow$)} & \multicolumn{2}{c}{Acc. (\%, $\uparrow$)} & \multicolumn{2}{c}{Rounds ($\downarrow$)}  \\
                                          & \multicolumn{1}{c}{500R}           & \multicolumn{1}{c}{1000R}     & 36\%         & 40\%                            & 500R           & \multicolumn{1}{c}{1000R}    & 24\%         & 30\%                                     \\ 
\midrule \midrule
FedMLB~\citep{kim2022multi}                       & {32.30} &  {42.61} & 642 & 800 &{28.39} &{33.67} & {384}&  {489}\\
FedMLB + ACG                    & \bf{41.10} &  \bf{55.27} & \bf{402} & \bf{479} &\bf{35.92} &\bf{43.85} & \bf{209}&  \bf{313}\\
\midrule \midrule
FedLC~\cite{zhang2022Federated}                          & 29.58 &  {36.78} & 936 & 1000+ & {22.14} &{26.83} & {676}&  1000+\\
FedLC + ACG                      & \bf 35.87 &\bf 43.51 &\bf 503 &\bf 675 & \bf 29.13 & \bf 33.17 & \bf 263 & \bf 557 \\
\midrule \midrule
FedDecorr~\cite{shi2022towards}                     & {30.56} &  {38.20} & 850 & 1000+ & {24.34} &{30.28} & {499}&  {959}\\
FedDecorr + ACG                & \bf 41.18 & \bf 49.93 & \bf 367 & \bf 473 & \bf 29.24 & \bf 34.71 & \bf 290 & \bf 540 \\

\bottomrule
\end{tabular}}
\end{subtable}
\end{center}
\end{table*}

\subsection{Evaluation on Various Data Heterogeneity}
Tables~\ref{tab:dir06} and~\ref{tab:iid} show that FedACG also matches or outperforms the performance of competitive methods when data heterogeneity is not severe (Dirichlet 0.6) or very low (\textit{i.i.d.}) on CIFAR-10 and CIFAR-100 in most cases. 

\begin{table*}[h!]
\caption{Results with Dirichlet (0.6) data on CIFAR-10 and CIFAR-100 for two different settings. 
}
\vspace{-2mm}
\label{tab:dir06}
\begin{subtable}[t]{1\textwidth}
\begin{center}
\caption{Dirichlet (0.6), 100 clients, 5\% participation}
\scalebox{0.85}{
\begin{tabular}{lcccccccc} 
\toprule
\multirow{3}{*}{Method}                                   & \multicolumn{4}{c}{CIFAR-10}                                                                     & \multicolumn{4}{c}{CIFAR-100}                                                                 \\ 
\cmidrule(lr){2-5} \cmidrule(lr){6-9}
 & \multicolumn{2}{c}{Acc. (\%, $\uparrow$)} & \multicolumn{2}{c}{Rounds ($\downarrow$)} & \multicolumn{2}{c}{Acc. (\%, $\uparrow$)} & \multicolumn{2}{c}{Rounds ($\downarrow$)}  \\
                                          & \multicolumn{1}{c}{500R}           & \multicolumn{1}{c}{1000R}     & 81\%         & 87\%                            & 500R           & \multicolumn{1}{c}{1000R}    & 50\%         & 56\%                                     \\ 
\midrule
FedAvg~\cite{mcmahan2017communication}   &80.56 & 85.97& 520 & 1000+ & 43.91 & 49.18 & 1000+ & 1000+ \\
FedProx~\cite{li2020federated}        & 80.39 & 85.53 & 524 & 1000+ & 43.15 &48.45 &1000+ &  1000+ \\
FedAvgM~\cite{hsu2019measuring}      & 84.65 & 87.96 & 355 & 811 &46.66  & 52.49 & 735 & 1000+  \\
FedADAM~\cite{reddi2021adaptive}        & 80.25 & 83.52 &526& 1000+ &45.95 & 51.63 & 778& 1000+  \\
FedDyn~\cite{acar2021federated}      & 87.23 & 89.49 & 310 & 487 & 50.51 &56.78 & 488& 886    \\
MOON~\cite{li2021model}           & 84.95 & 87.99 & 272 & 728 & 55.76 & 61.42 & 338 & 527 \\ 
\rowcolor{Gray}
FedCM$^\dagger$~\cite{xu2021fedcm}            & 82.84 & 86.64 & 385  & 1000+ &53.75 &60.48 &331 & 468    \\
FedMLB~\citep{kim2022multi}      & 79.85 & 85.98 & 574 & 1000+ & 49.31 & 56.70 & 526  & 925\\
FedLC~\cite{zhang2022Federated}      & 80.40 & 85.48 & 559 & 1000+ & 43.99 & 48.92 & 1000+ & 1000+ \\
FedNTD~\cite{lee2022preservation}      & 81.2 & 86.44 & 498 & 1000+ & 44.26 & 50.34 & 916 & 1000+ \\
\rowcolor{Gray}
FedDC$^{\ddagger}$~\cite{gao2022feddc}           & \bf88.05 & 89.58 & 270 & {\bf437} & 56.00 & 60.58 & 347 & 491 \\ 
FedDecorr~\cite{shi2022towards}      & 81.01 & 85.19 & 500 & 1000+ & 43.64 & 49.03 & 1000+ & 1000+ \\
FedACG (ours)                       & {87.57} &  \bf{90.56} & \bf{218} & 453 &\bf{58.82} &\bf{63.88} & \bf{243}&  \bf{396}\\
\bottomrule
\end{tabular}}
\end{center}
\end{subtable}
\begin{center}
\begin{subtable}[t]{1\textwidth}
\centering
\caption{Dirichlet (0.6), 500 clients, 2\% participation}
\scalebox{0.85}{
\begin{tabular}{lcccccccc} 
\toprule
\multirow{3}{*}{Method}                                   & \multicolumn{4}{c}{CIFAR-10}                                                                     & \multicolumn{4}{c}{CIFAR-100}                                                                 \\ 
\cmidrule(lr){2-5} \cmidrule(lr){6-9}
 & \multicolumn{2}{c}{Acc. (\%, $\uparrow$)} & \multicolumn{2}{c}{Rounds ($\downarrow$)} & \multicolumn{2}{c}{Acc. (\%, $\uparrow$)} & \multicolumn{2}{c}{Rounds ($\downarrow$)}  \\
                                          & \multicolumn{1}{c}{500R}           & \multicolumn{1}{c}{1000R}     & 69\%         & 80\%                            & 500R           & \multicolumn{1}{c}{1000R}    & 32\%         & 41\%                                     \\ 
\midrule
FedAvg~\cite{mcmahan2017communication}  & 62.79 & 75.17 & 671 &1000+  & 29.41 & 36.62 &648&1000+ \\
FedProx~\cite{li2020federated}          & 62.48 & 75.10 &688&1000+& 29.62 & 36.70 &647&1000+ \\
FedAvgM~\cite{hsu2019measuring}         & 69.10 & 80.26 &498&981& 32.78 & 41.93 &468  &942 \\
FedADAM~\cite{reddi2021adaptive}         & 68.48 & 78.92 &535&1000+& 37.57 & 48.29 &341&624 \\
FedDyn~\cite{acar2021federated}          & 68.53 & 80.33 &513&983& 32.06 & 43.28 &498& 917 \\
MOON~\cite{li2021model}                 & 74.29 & 80.66 & 368 & 921 & 31.64 & 41.61 & 515 & 931 \\ 
\rowcolor{Gray}
FedCM$^\dagger$~\cite{xu2021fedcm}                 & 71.42 & 78.94 &429&1000+& 26.82 & 39.78 &714&1000+\\
FedMLB~\citep{kim2022multi}      & 62.60 & 74.36 & 729 & 1000+ & 33.79 & 43.52 & 432 &  831\\
FedLC~\cite{zhang2022Federated}      & 62.77 & 73.56 & 694 & 1000+ & 30.07 & 36.97 & 620 & 1000+ \\
FedNTD~\cite{lee2022preservation}      & 61.9 & 74.38 & 717 & 1000+ & 28.85 & 35.88 & 691 & 1000+ \\
\rowcolor{Gray}
FedDC$^{\ddagger}$~\cite{gao2022feddc}               & 77.74 & {\bf86.32} & 324 & 596 & 34.24 & 44.69 & 444 & 825\\ 
FedDecorr~\cite{shi2022towards}      & 63.63 & 74.89 & 658 & 1000+ & 29.99 & 37.72 & 615 & 1000+ \\
FedACG (ours)                            & {\bf78.49} & 85.28 & {\bf289}& {\bf565} & {\bf39.61} & {\bf49.70} & {\bf304}& {\bf540}\\
\bottomrule
\end{tabular}}
\end{subtable}
\end{center}
\vspace{-0.4cm}
\end{table*}

\begin{table*}[h!]
\caption{Results with \textit{i.i.d.} data on CIFAR-10 and CIFAR-100 for two different settings. 
}
\vspace{-2mm}
\label{tab:iid}
\begin{subtable}[t]{1\textwidth}
\begin{center}
\caption{\textit{i.i.d.}, 100 clients, 5\% participation}
\scalebox{0.85}{
\begin{tabular}{lcccccccc} 
\toprule
\multirow{3}{*}{Method}                                   & \multicolumn{4}{c}{CIFAR-10}                                                                     & \multicolumn{4}{c}{CIFAR-100}                                                                 \\ 
\cmidrule(lr){2-5} \cmidrule(lr){6-9}
 & \multicolumn{2}{c}{Acc. (\%, $\uparrow$)} & \multicolumn{2}{c}{Rounds ($\downarrow$)} & \multicolumn{2}{c}{Acc. (\%, $\uparrow$)} & \multicolumn{2}{c}{Rounds ($\downarrow$)}  \\
                                          & \multicolumn{1}{c}{500R}           & \multicolumn{1}{c}{1000R}     & 82\%         & 89\%                            & 500R           & \multicolumn{1}{c}{1000R}    & 52\%         & 58\%                                     \\ 
\midrule
FedAvg~\cite{mcmahan2017communication}   & 85.28 &88.69 & 372 & 1000+ & 43.96 & 48.20 & 1000+ & 1000+ \\
FedProx~\cite{li2020federated}        & 84.79 & 87.99 & 384 & 1000+ & 43.57 & 47.75 & 1000+ & 1000+ \\
FedAvgM~\cite{hsu2019measuring}      & 87.67& 89.96 & 258 & 375 & 47.43 & 52.83 & 880 & 1000+  \\
FedADAM~\cite{reddi2021adaptive}        & 85.29 & 87.97 & 286 & 1000+ & 52.23 & 57.73 & 496 & 1000+   \\
FedDyn~\cite{acar2021federated}      & 89.19 & 90.70 & 269 & 492 & 50.37 & 56.88 & 592 &898  \\
MOON~\cite{li2021model}          & 88.24 & 89.96 & 207 & 628 & 58.50 & \bf64.73 & 311 & 484 \\ 
\rowcolor{Gray}
FedCM$^\dagger$~\cite{xu2021fedcm}           & 87.38 & 89.65 & 182 &  782 & 57.10 & {62.48} & 266 & 466 \\
FedMLB~\citep{kim2022multi}      & 86.32 & 89.65 & 359 & 784 & 50.12 & 56.40 & 586 & 1000+ \\
FedLC~\cite{zhang2022Federated}      & 84.48 & 88.26 & 393 & 1000+ & 43.84 & 46.70 & 1000+ & 1000+ \\
FedNTD~\cite{lee2022preservation}     & 85.68 & 89.43 & 367 & 870 & 44.93 & 50.51 & 1000+ & 1000+ \\
\rowcolor{Gray}
FedDC$^{\ddagger}$~\cite{gao2022feddc}             & 90.07 & 90.80 & 194 & 425 & 55.17 &
61.00 & 400 & 633 \\ 
FedDecorr~\cite{shi2022towards}      & 85.21 & 88.17 & 364 & 1000+ & 45.16 & 49.16 & 1000+ & 1000+ \\
FedACG (ours)                       & \bf{90.57} &  \bf{92.29} & \bf{157} & \bf{354} & \bf{59.82} & 64.08 &  \bf{244} & \bf{342} \\
\bottomrule
\end{tabular}}
\end{center}
\end{subtable}

\begin{center}
\begin{subtable}[t]{1\textwidth}
\centering
\caption{\textit{i.i.d.}, 500 clients, 2\% participation}
\scalebox{0.85}{
\begin{tabular}{lcccccccc} 
\toprule
\multirow{3}{*}{Method}                                   & \multicolumn{4}{c}{CIFAR-10}                                                                     & \multicolumn{4}{c}{CIFAR-100}                                                                 \\ 
\cmidrule(lr){2-5} \cmidrule(lr){6-9}
 & \multicolumn{2}{c}{Acc. (\%, $\uparrow$)} & \multicolumn{2}{c}{Rounds ($\downarrow$)} & \multicolumn{2}{c}{Acc. (\%, $\uparrow$)} & \multicolumn{2}{c}{Rounds ($\downarrow$)}  \\
                                          & \multicolumn{1}{c}{500R}           & \multicolumn{1}{c}{1000R}     & 75\%         & 83\%                            & 500R           & \multicolumn{1}{c}{1000R}    & 33\%         & 42\%                                     \\ 
\midrule
FedAvg~\cite{mcmahan2017communication}   & 68.70 & 78.21 & 652 & 1000+ & 30.71 & 37.85 & 664 & 1000+ \\
FedProx~\cite{li2020federated}       & 68.74 & 77.96 & 643 & 1000+ & 30.11 & 37.13 & 685 & 1000+ \\
FedAvgM~\cite{hsu2019measuring}      & 74.34 & 83.64 & 523 & 943 & 33.54 & 42.55 & 479 & 971\\
FedADAM~\cite{reddi2021adaptive}      & 75.32 & 84.01 & 491 & 915 & 38.74 & 48.94 & 328 & 636 \\
FedDyn~\cite{acar2021federated}      & 74.81 & 84.71 & 398 & 823 & 33.20 & 42.91 &  492 & 936  \\
MOON~\cite{li2021model}              & 69.86 & 81.89 & 586 & 1000+ & 28.82 & 41.26 & 649 & 1000+\\
\rowcolor{Gray}
FedCM$^\dagger$~\cite{xu2021fedcm}             & 77.84 & 83.26 & 491 & 959 & 29.59 & 42.04 & 653 & 991\\
FedMLB~\citep{kim2022multi}      & 62.60 & 80.16 & 729 & 1000+ & 34.56  & 44.95 & 440 & 817 \\
FedLC~\cite{zhang2022Federated}      & 68.92 & 79.09 & 727 & 1000+ & 29.91 & 37.18 & 677 & 1000+ \\
FedNTD~\cite{lee2022preservation}      & 68.61 & 80.65 & 706 & 1000+ & 30.04 & 36.63 & 706 & 1000+ \\ 
\rowcolor{Gray}
FedDC$^{\ddagger}$~\cite{gao2022feddc}              & {\bf80.87} & {\bf87.53} & 358 & {\bf574} & 33.93 & 45.80 & 476 & 817 \\ 
FedDecorr~\cite{shi2022towards}      & 68.12 & 77.39 & 802 & 1000+ & 30.41 & 37.53 & 585 & 1000+ \\
FedACG (ours)                         & 80.15 & 87.47 & {\bf316} & 578 & {\bf41.16} & {\bf49.10} & {\bf299} & {\bf525}\\
\bottomrule
\end{tabular}}
\end{subtable}
\end{center}
\end{table*}

\clearpage

\section{Convergence Plot}

\subsection{Evaluation on various federated learning scenarios}
\label{sec:plot_main}
\begin{figure}[h!]
\centering
\begin{subfigure}[b]{0.4\linewidth}
\centering
\includegraphics[width=1\linewidth]{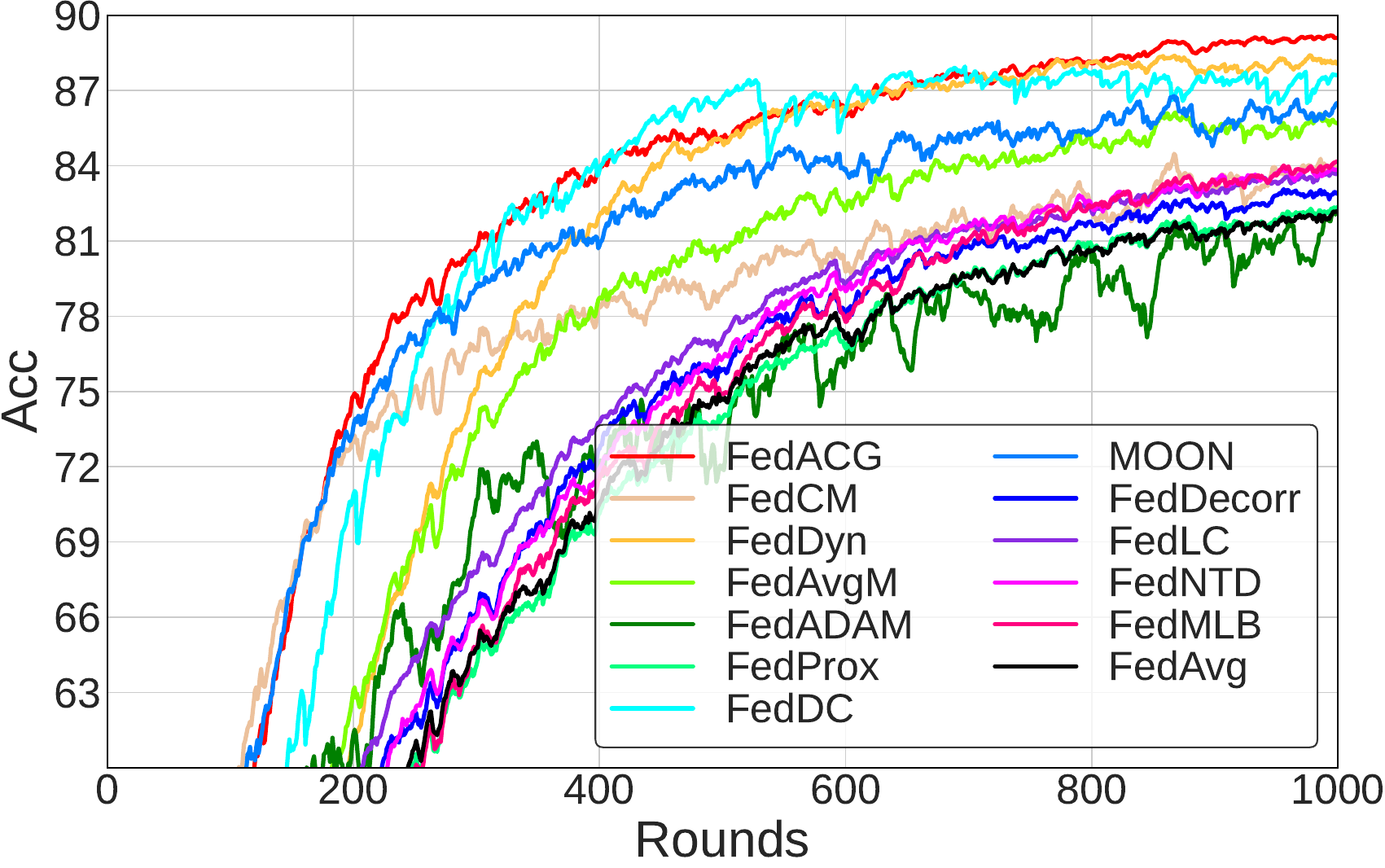}
\caption{Dirichlet (0.3), 100 clients, 5\% participation}
\end{subfigure}
\hspace{0.4cm}
\begin{subfigure}[b]{0.4\linewidth}
\centering
\includegraphics[width=1\linewidth]{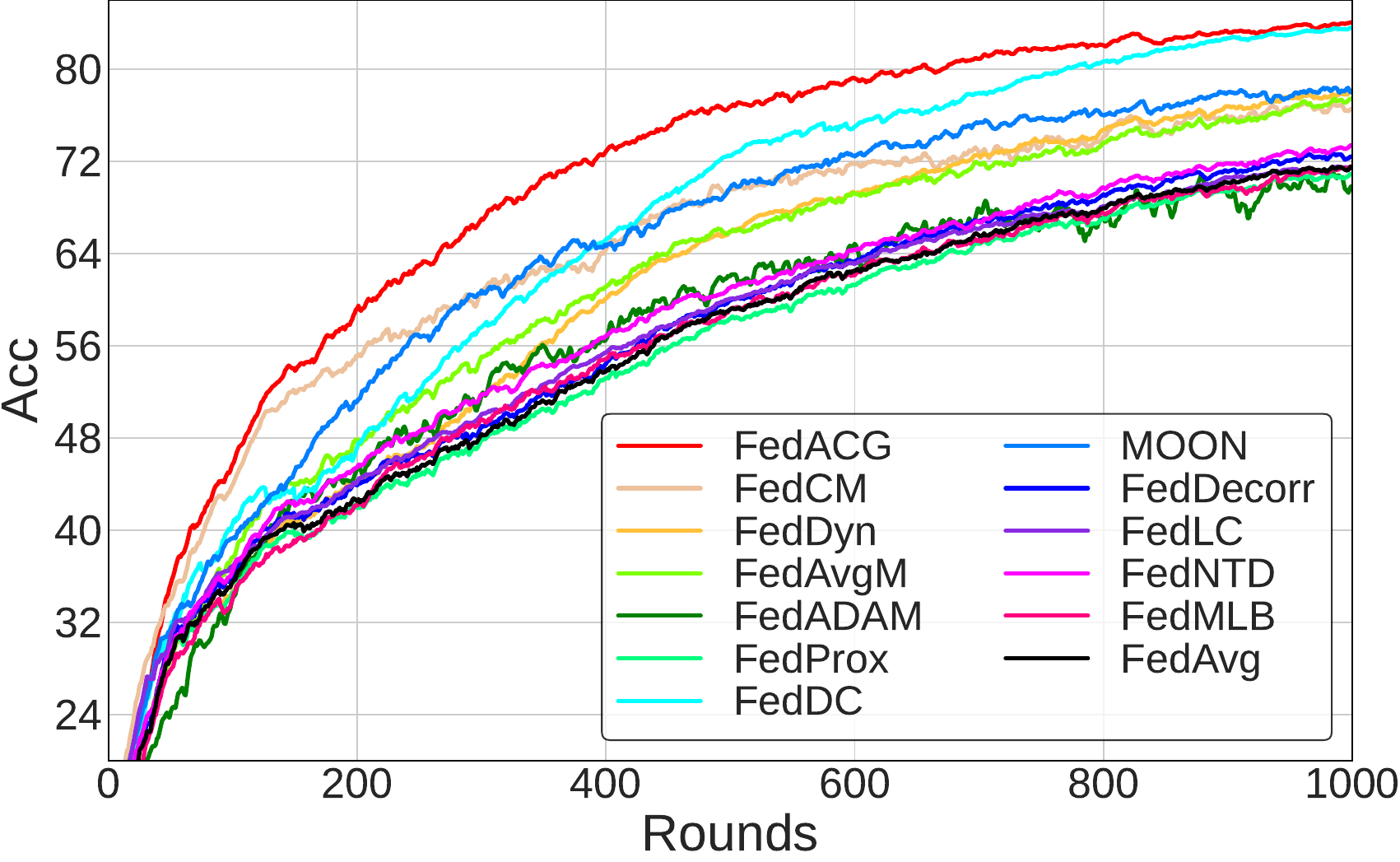}
\caption{Dirichlet (0.3), 500 clients, 2\% participation}
\end{subfigure}

\vspace{0.8cm}

\begin{subfigure}[b]{0.4\linewidth}
\centering
\includegraphics[width=1\linewidth]{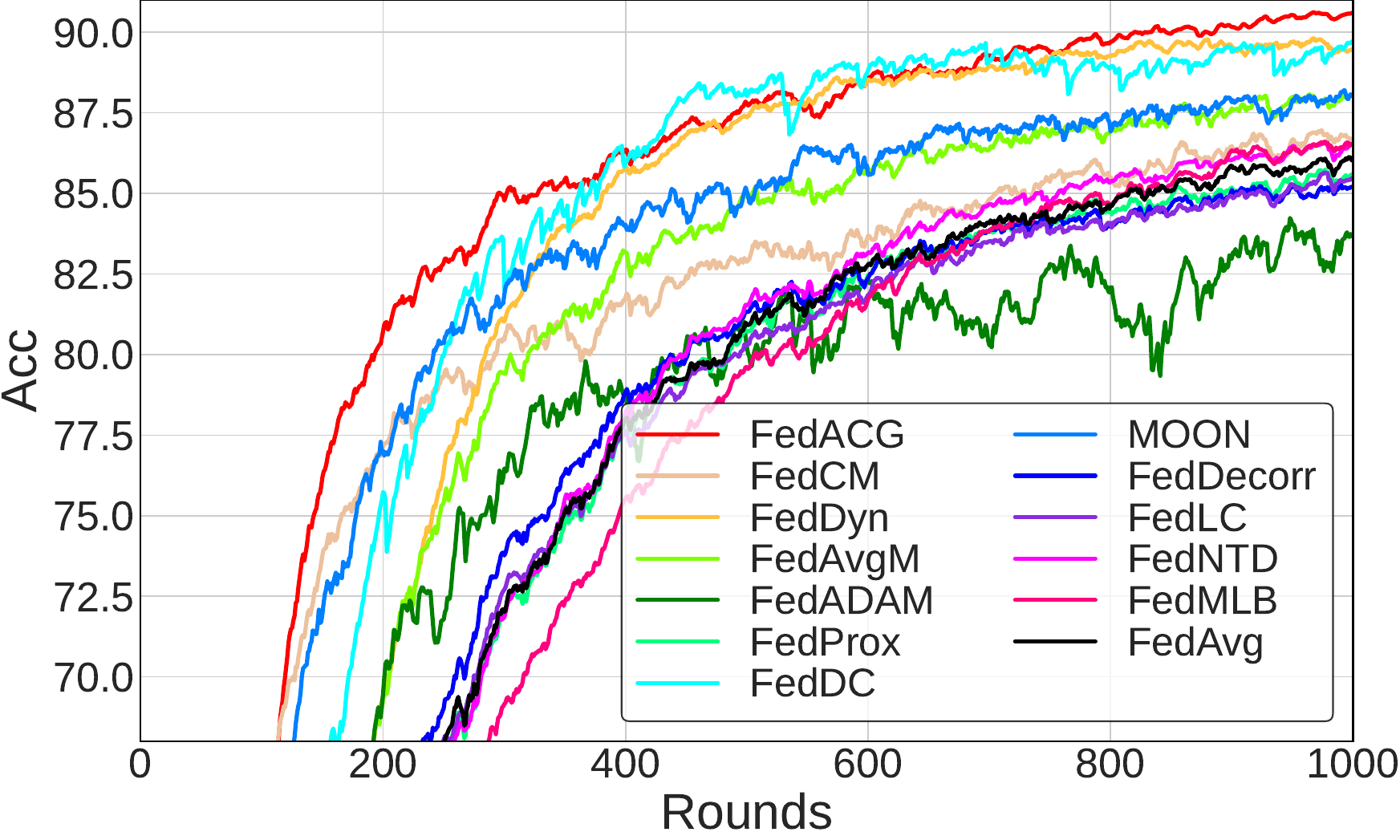}
\caption{Dirichlet (0.6), 100 clients, 5\% participation}
\end{subfigure}
\hspace{0.4cm}
\begin{subfigure}[b]{0.4\linewidth}
\centering
\includegraphics[width=1\linewidth]{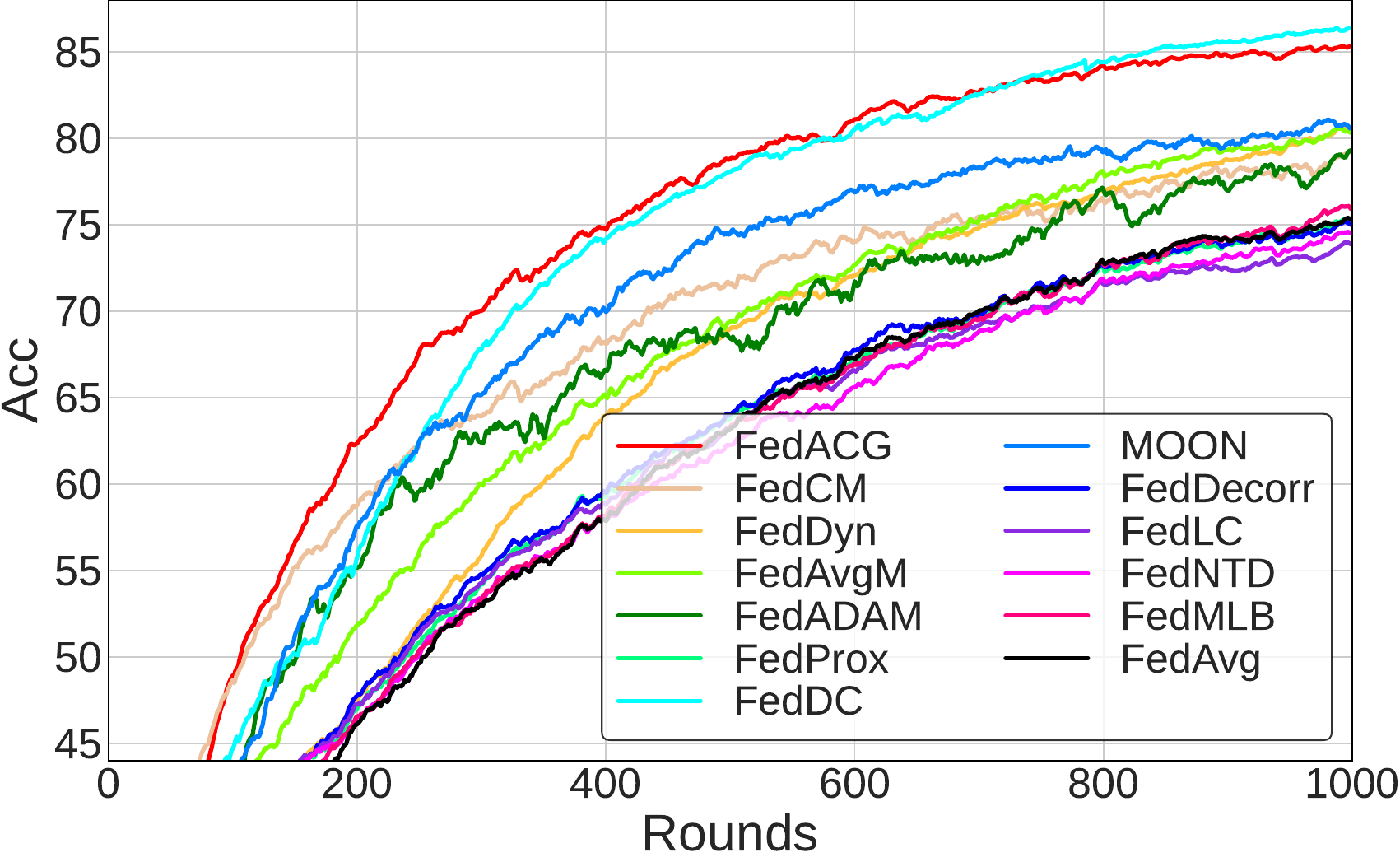}
\caption{Dirichlet (0.6), 500 clients, 2\% participation}
\end{subfigure}

\vspace{0.8cm}

\begin{subfigure}[b]{0.4\linewidth}
\centering
\includegraphics[width=1\linewidth]{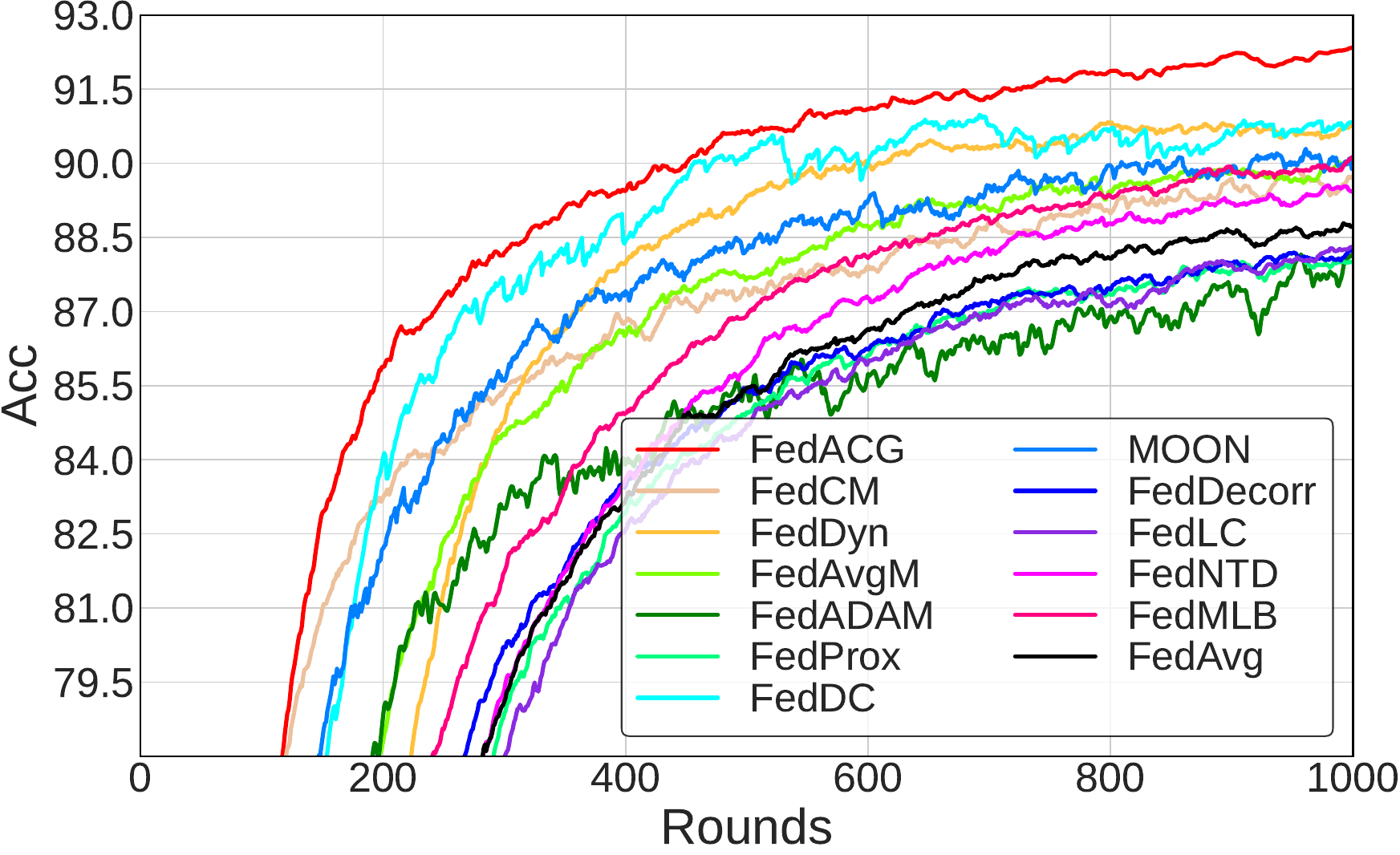}
\caption{\textit{i.i.d.}, 100 clients, 5\% participation}
\end{subfigure}
\hspace{0.4cm}
\begin{subfigure}[b]{0.4\linewidth}
\centering
\includegraphics[width=1\linewidth]{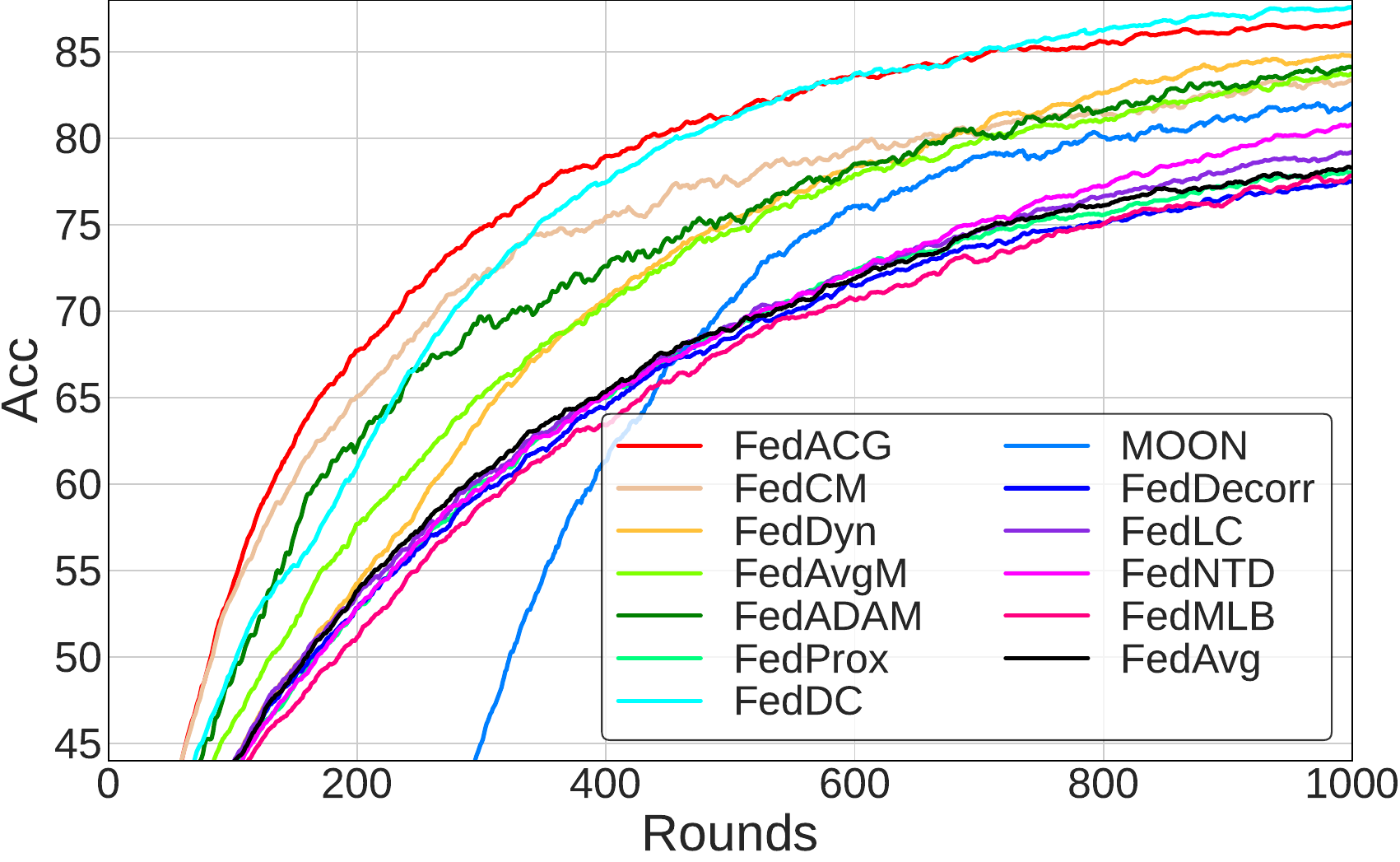}
\caption{\textit{i.i.d.}, 500 clients, 2\% participation}
\end{subfigure}
\caption{
The convergence plots of FedACG and the baselines on CIFAR-10 with different federated learning scenarios.
}
\label{fig:convergence_cifar10}
\end{figure}

\begin{figure}[h!]
\centering
\begin{subfigure}[b]{0.4\linewidth}
\centering
\includegraphics[width=1\linewidth]{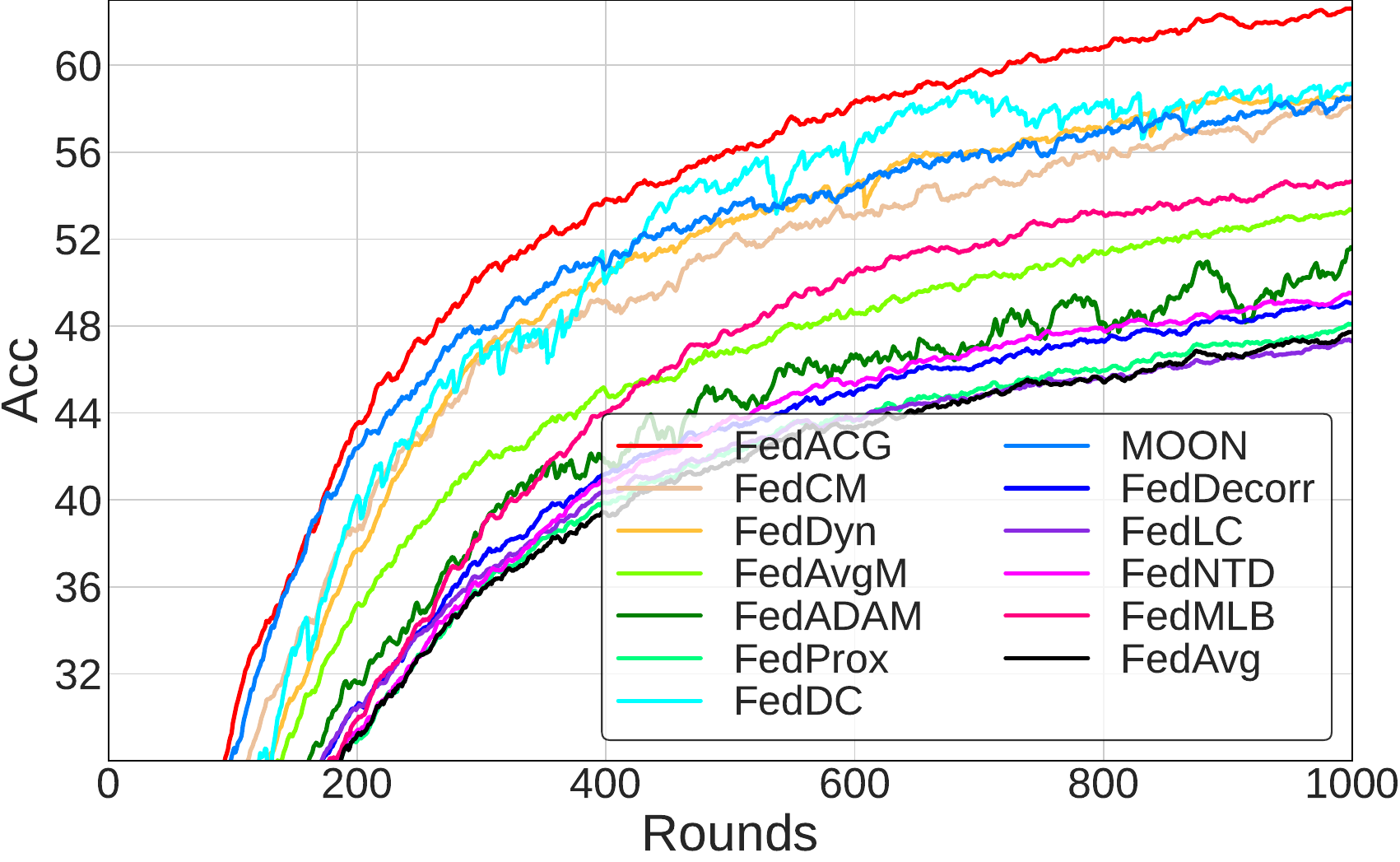}
\caption{Dirichlet (0.3), 100 clients, 5\% participation}
\end{subfigure}
\hspace{0.4cm}
\begin{subfigure}[b]{0.4\linewidth}
\centering
\includegraphics[width=1\linewidth]{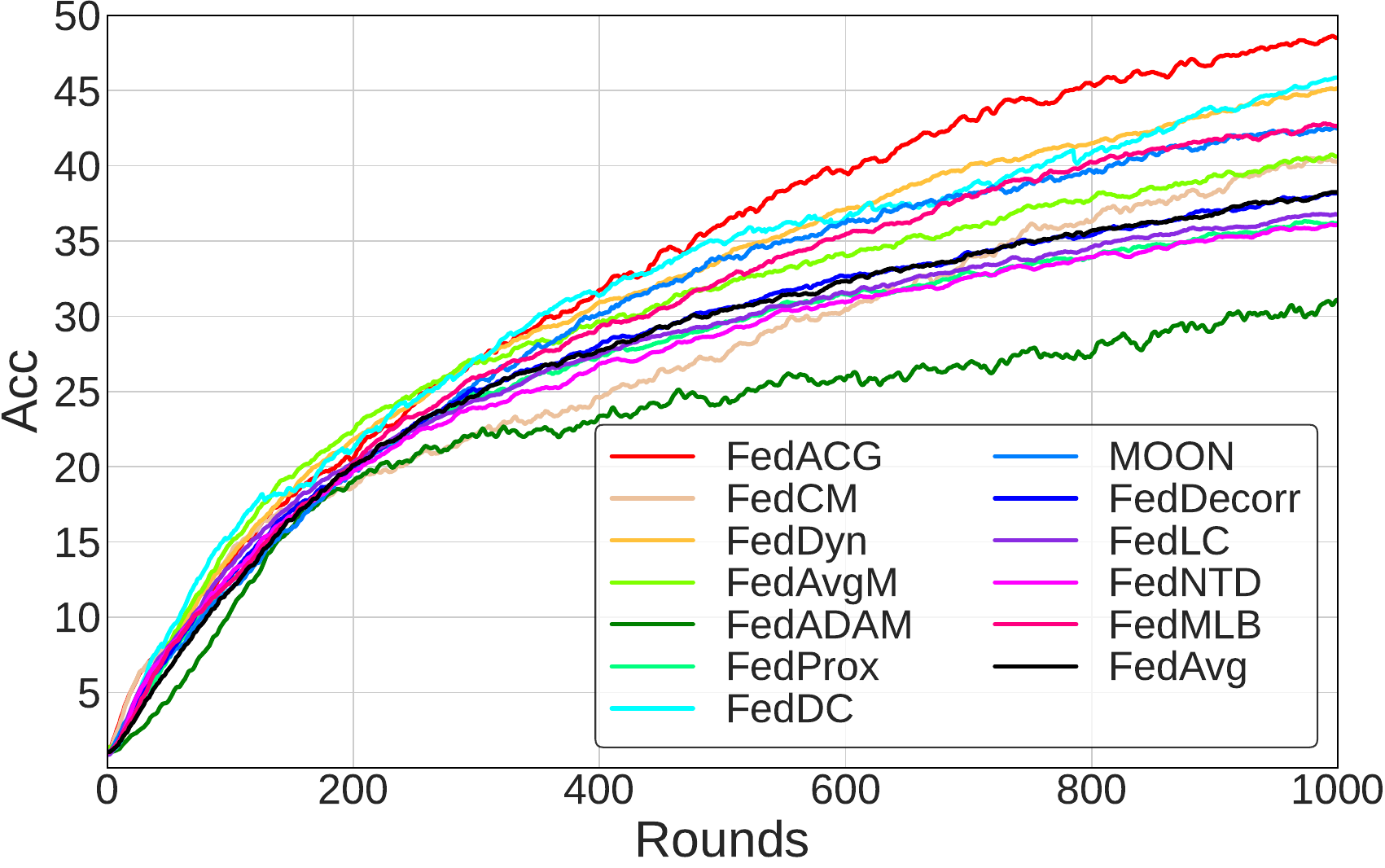}
\caption{Dirichlet (0.3), 500 clients, 2\% participation}
\end{subfigure}

\vspace{0.3cm}
\begin{subfigure}[b]{0.4\linewidth}
\centering
\includegraphics[width=1\linewidth]{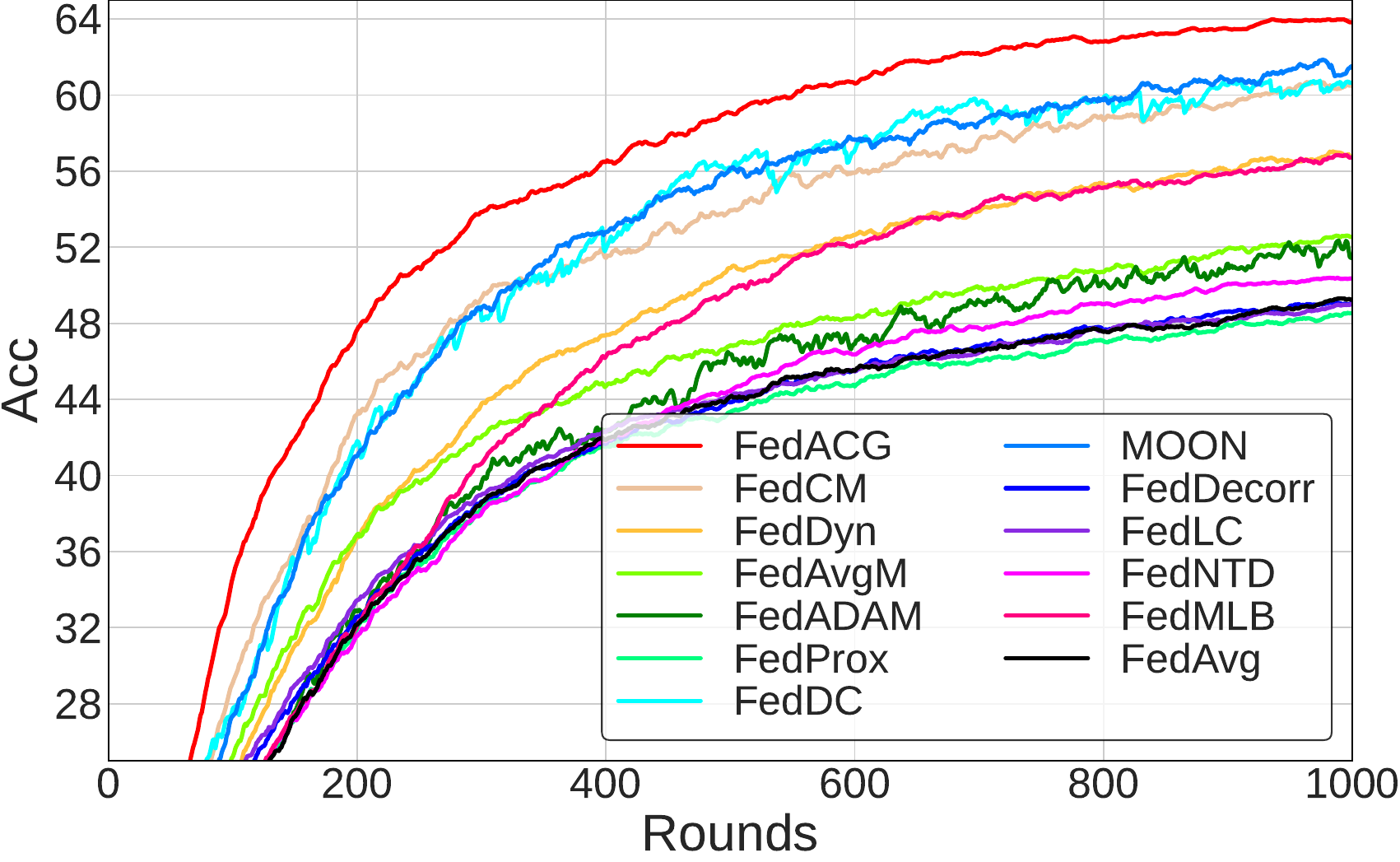}
\caption{Dirichlet (0.6), 100 clients, 5\% participation}
\end{subfigure}
\hspace{0.4cm}
\begin{subfigure}[b]{0.4\linewidth}
\centering
\includegraphics[width=1\linewidth]{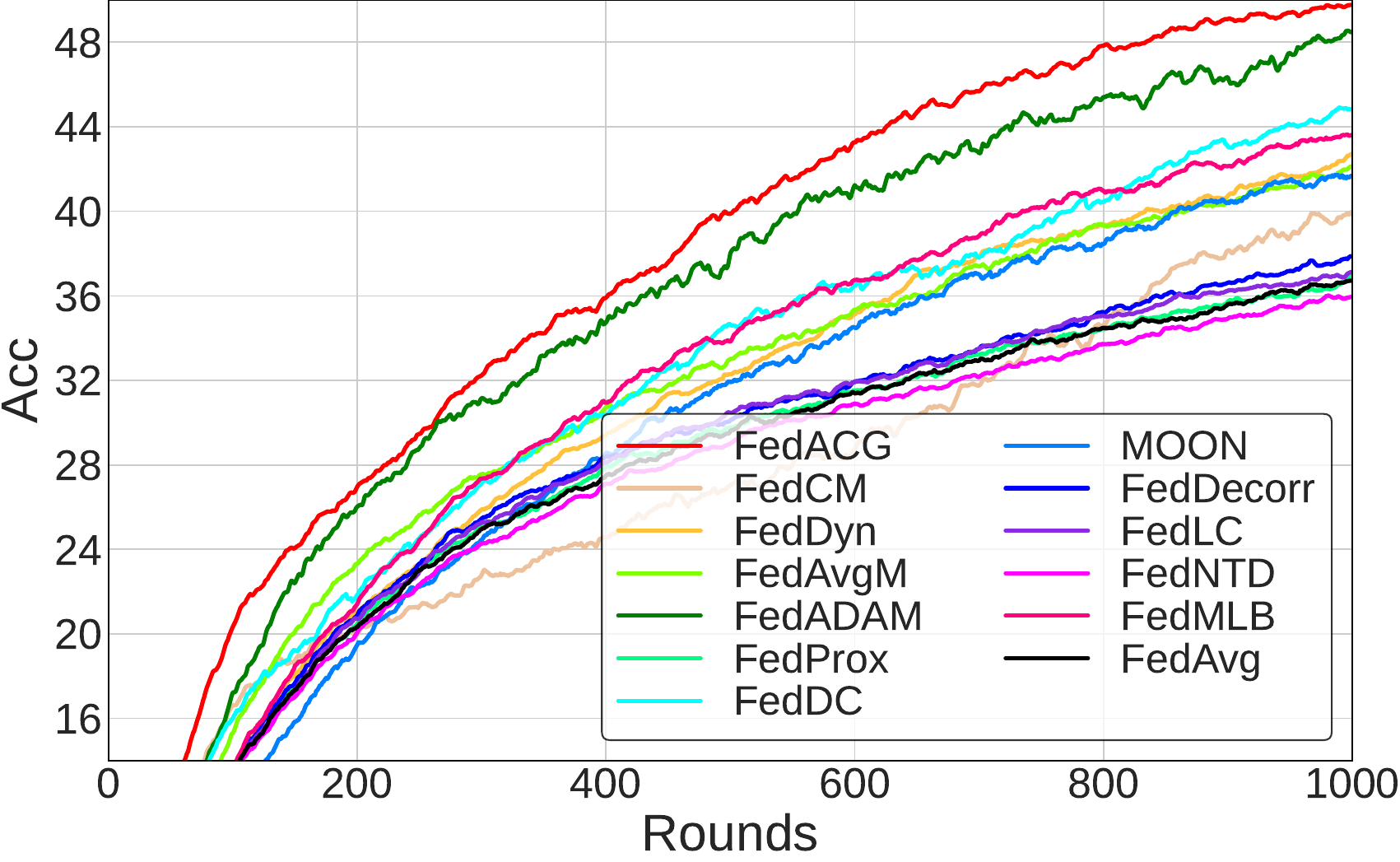}
\caption{Dirichlet (0.6), 500 clients, 2\% participation}
\end{subfigure}

\vspace{0.3cm}

\begin{subfigure}[b]{0.4\linewidth}
\centering
\includegraphics[width=1\linewidth]{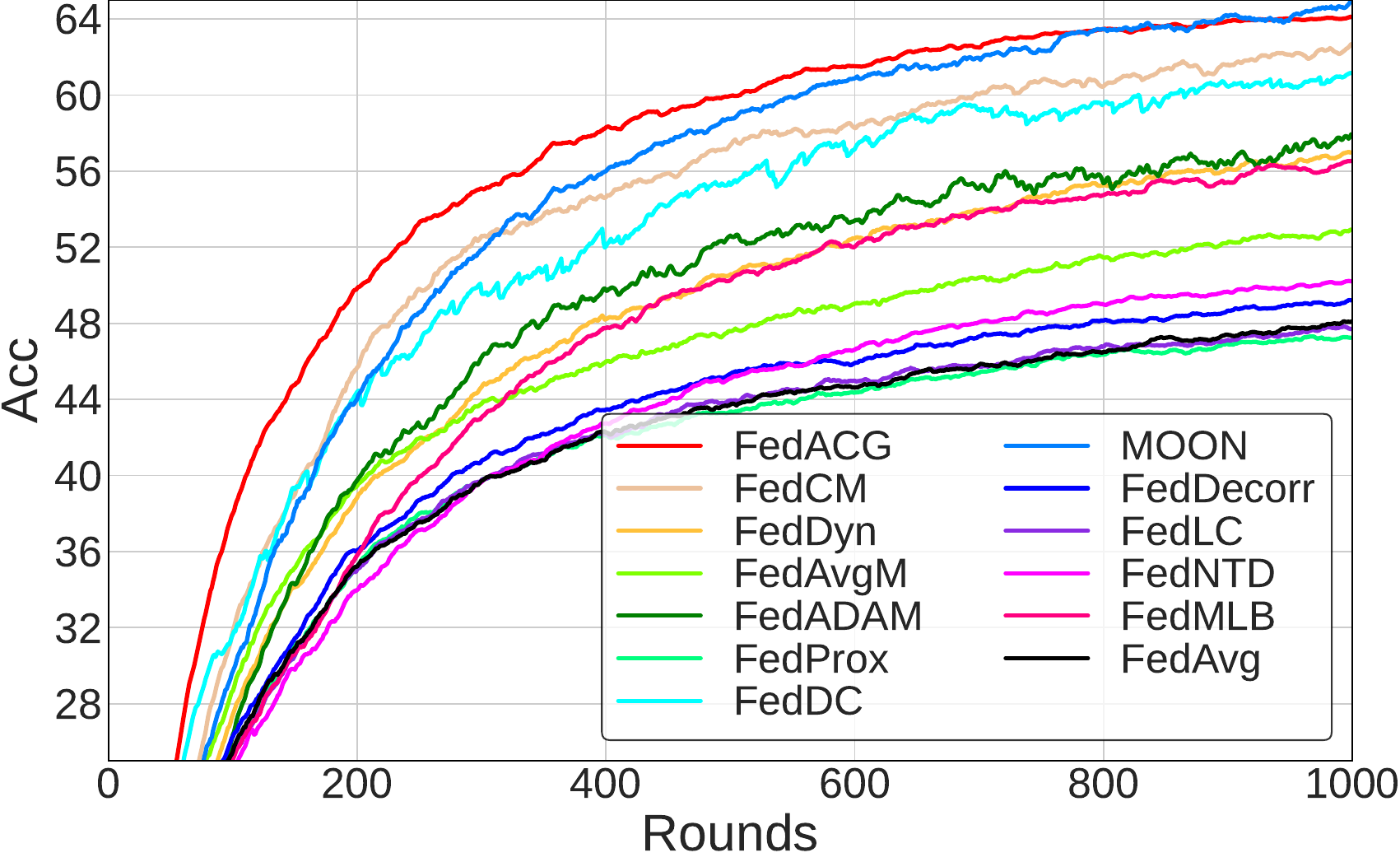}
\caption{\textit{i.i.d.}, 100 clients, 5\% participation}
\end{subfigure}
\hspace{0.4cm}
\begin{subfigure}[b]{0.4\linewidth}
\centering
\includegraphics[width=1\linewidth]{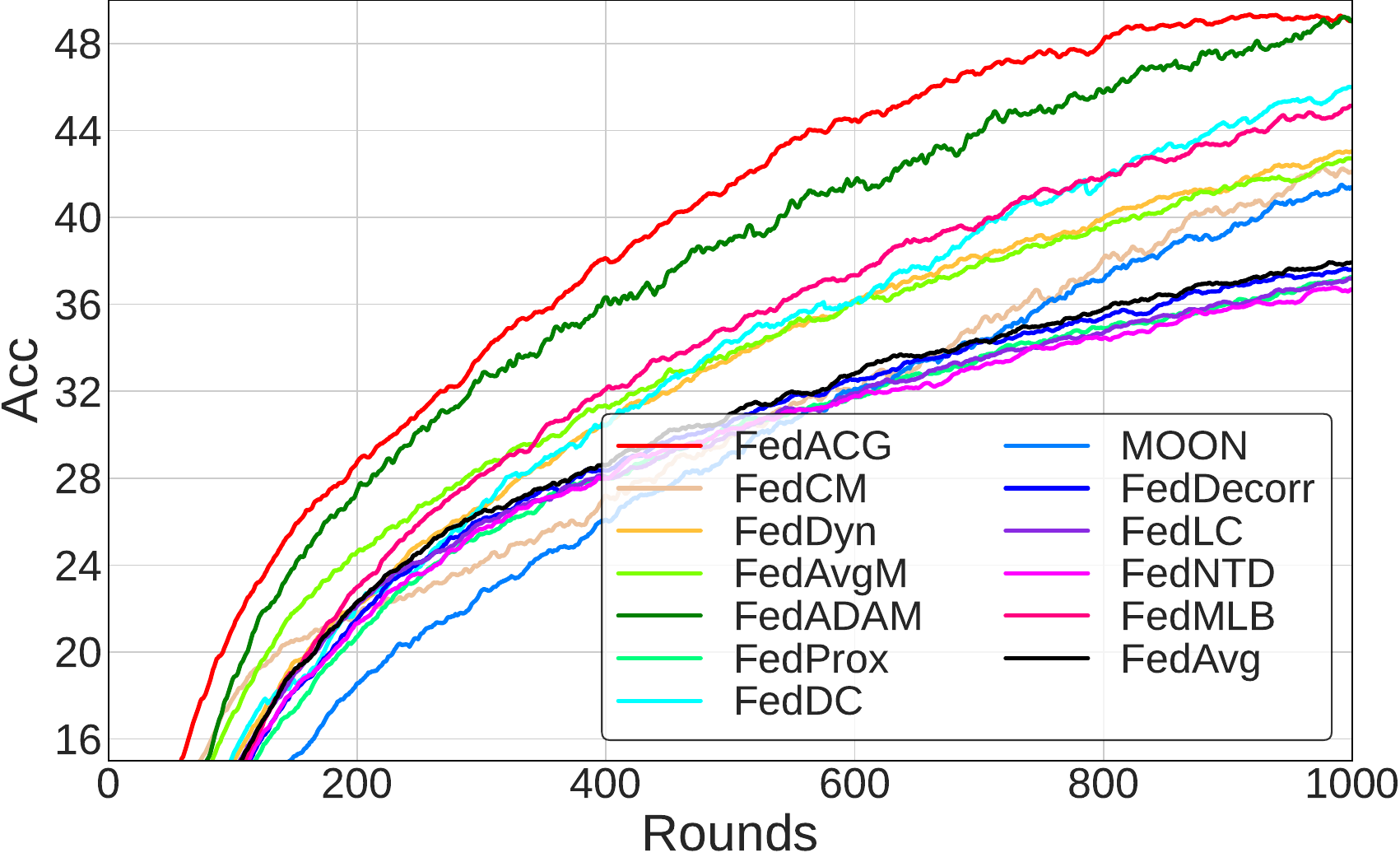}
\caption{\textit{i.i.d.}, 500 clients, 2\% participation}
\end{subfigure}
\caption{
The convergence plots of FedACG and the baselines on CIFAR-100 with different federated learning scenarios.
}
\label{fig:convergence_cifar100}
\end{figure}

\begin{figure}[h!]
\centering
\begin{subfigure}[b]{0.4\linewidth}
\centering
\includegraphics[width=1\linewidth]{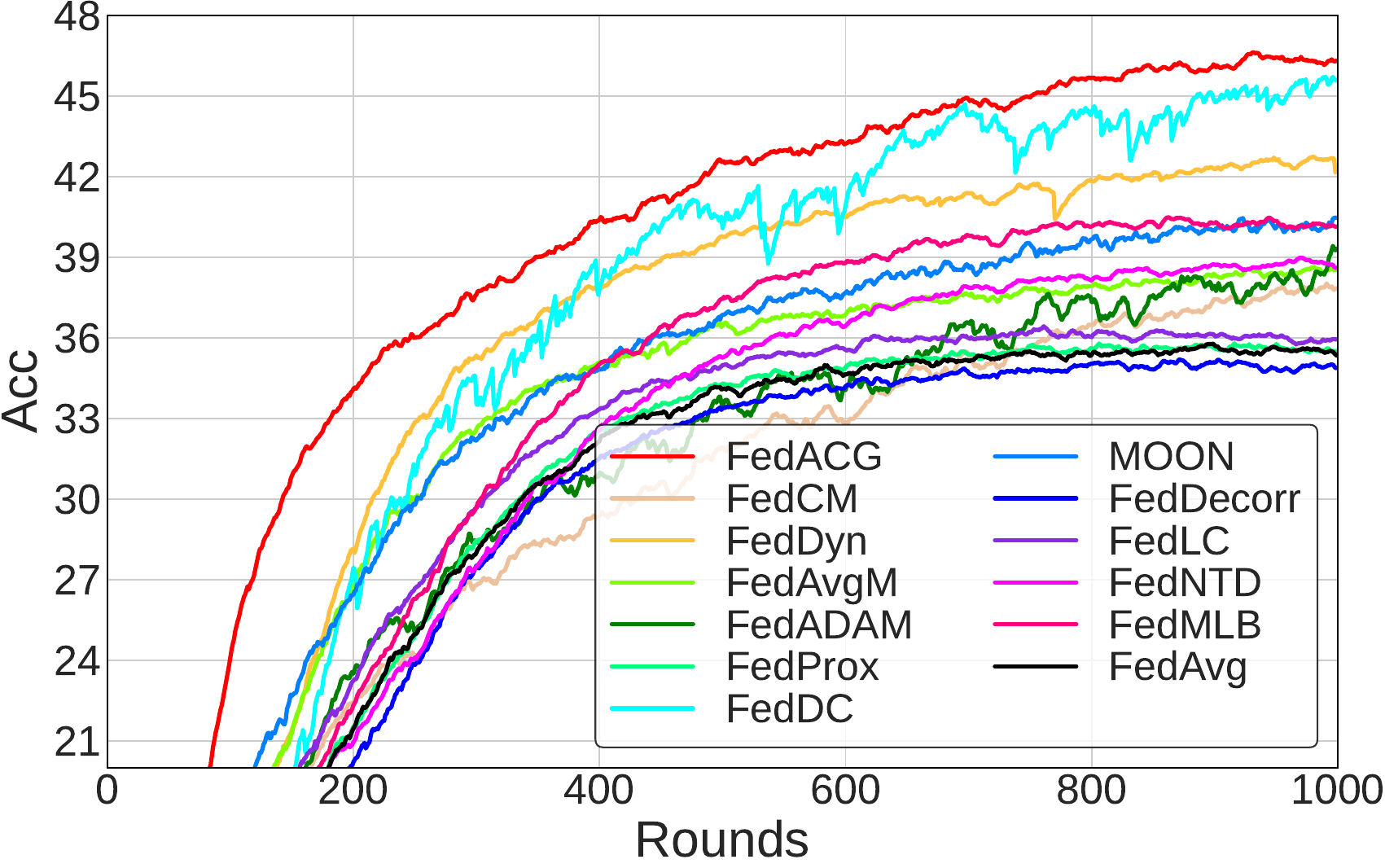}
\caption{Dirichlet (0.3), 100 clients, 5\% participation}
\end{subfigure}
\hspace{0.4cm}
\begin{subfigure}[b]{0.4\linewidth}
\centering
\includegraphics[width=1\linewidth]{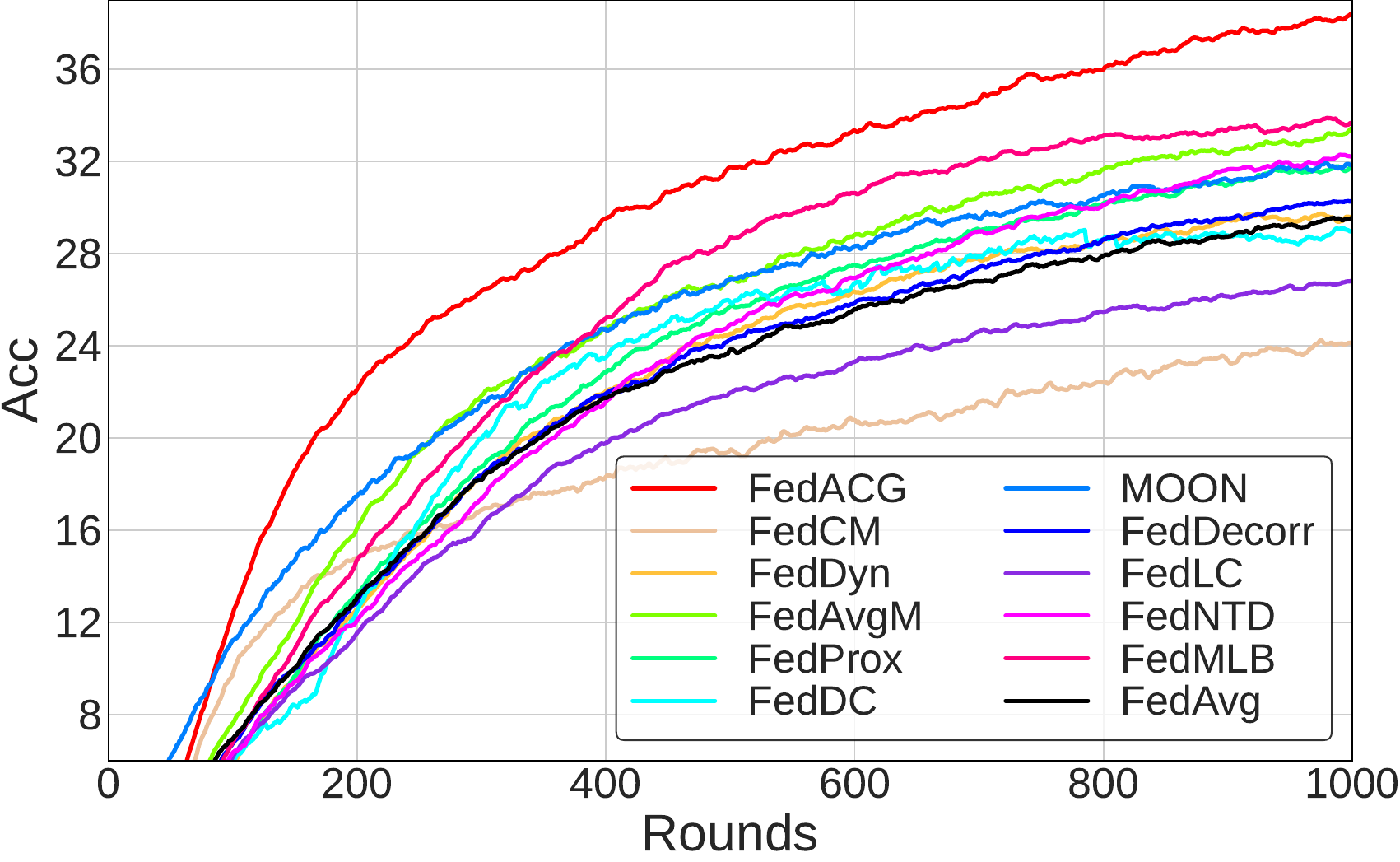}
\caption{Dirichlet (0.3), 500 clients, 2\% participation}
\end{subfigure}

\caption{
The convergence plots of FedACG and the baselines on Tiny-ImageNet with different federated learning scenarios.
}
\label{fig:convergence_tiny}
\end{figure}

{Figure~\ref{fig:convergence_cifar10} to Figure~\ref{fig:convergence_tiny} show the convergence of FedACG and the compared algorithms on CIFAR-10, CIFAR-100, and Tiny-ImageNet for various federated learning settings: varying the number of total clients, participation rates, data heterogeneity.}
{FedACG continuously matches or exceeds the performance of the most powerful of our competitors in most learning sections.}

\begin{figure}[t]
\centering
\begin{subfigure}[b]{0.4\linewidth}
\centering
\includegraphics[width=1\linewidth]{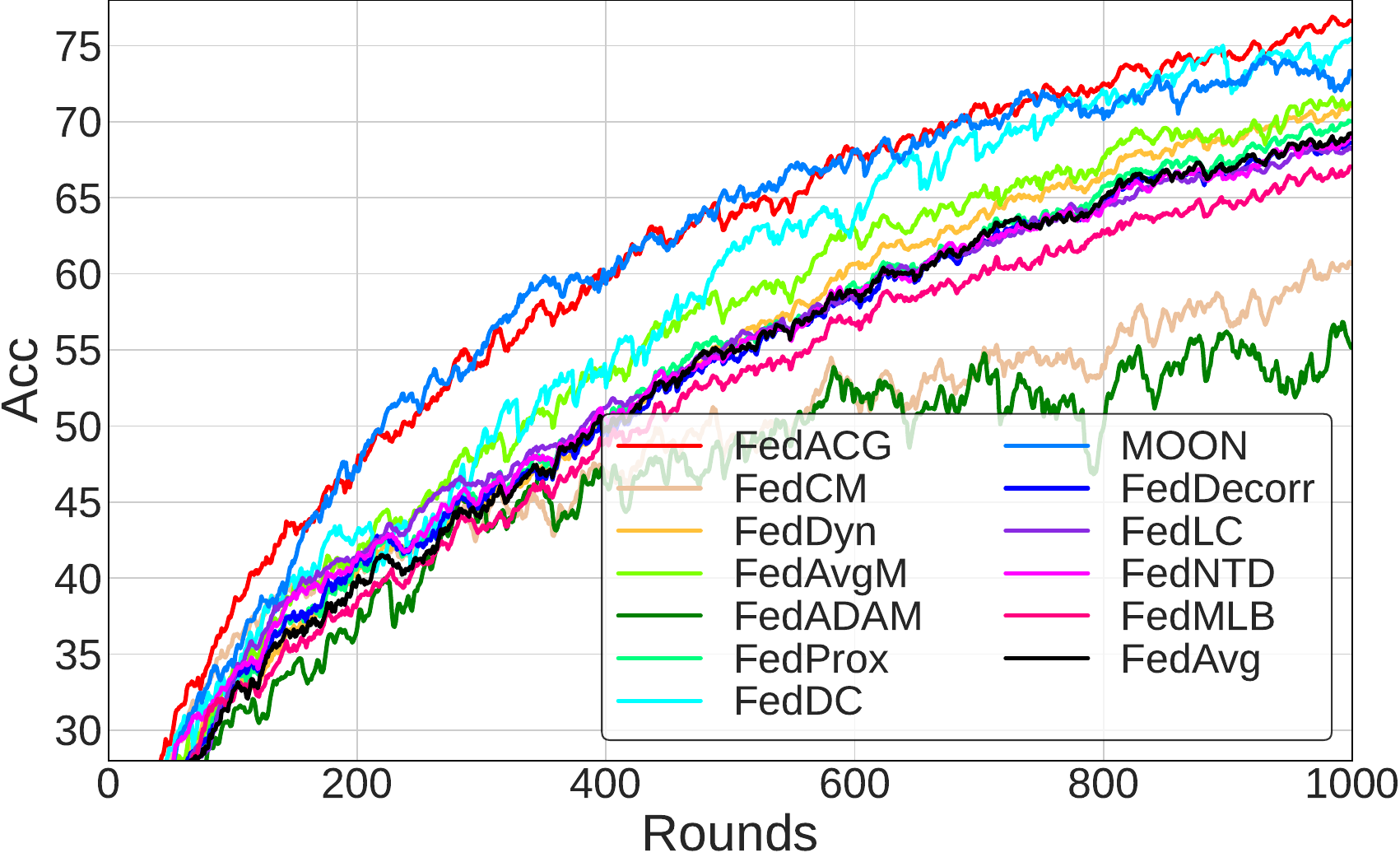}
\caption{CIFAR-10}
\end{subfigure}
\hspace{0.4cm}
\begin{subfigure}[b]{0.4\linewidth}
\centering
\includegraphics[width=1\linewidth]{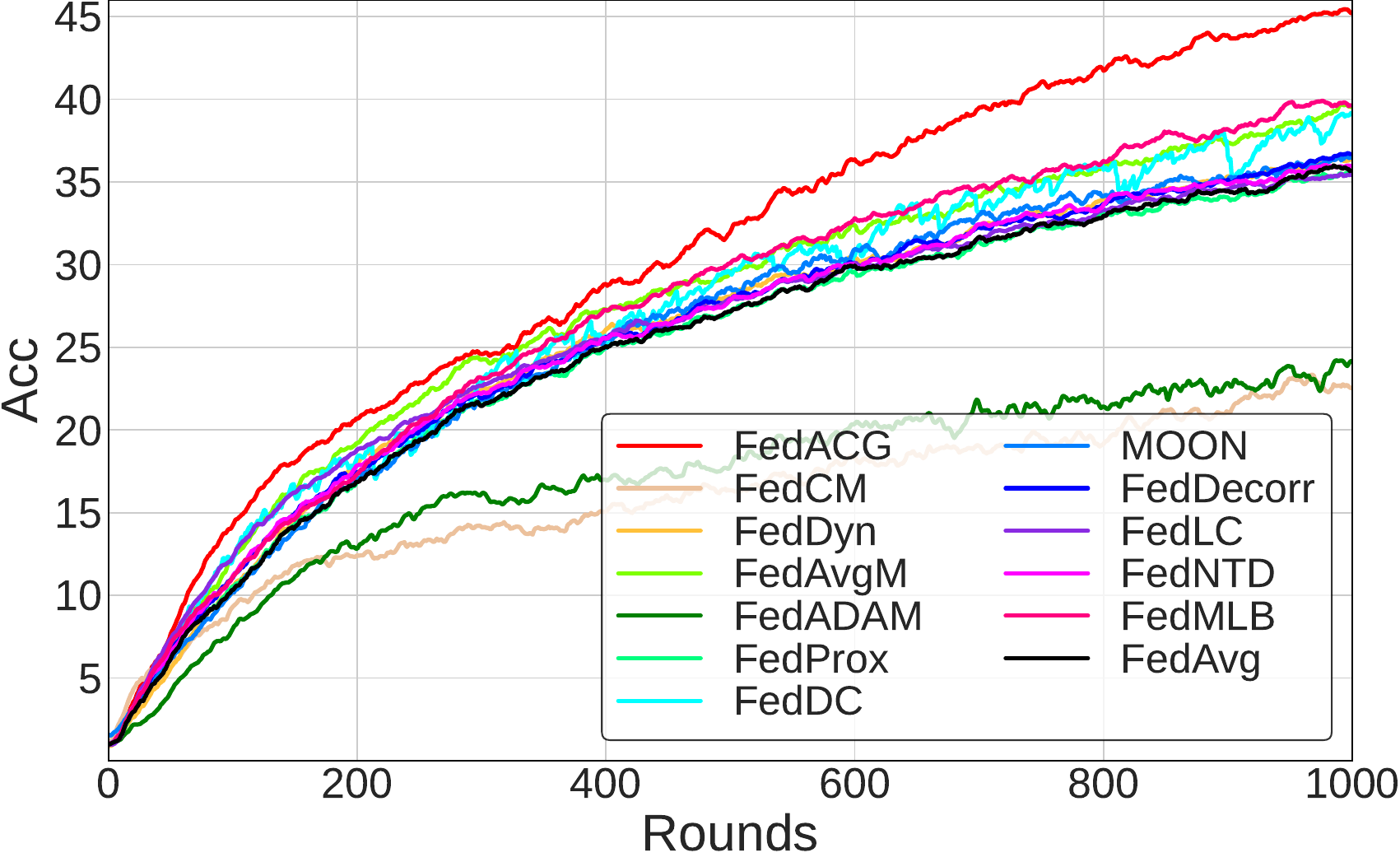}
\caption{CIFAR-100}
\end{subfigure}
\caption{
The convergence plots of FedACG and the baselines when the participation rate is low (1\%) for 500 clients on CIFAR-10 and CIFAR-100. The Dirichlet parameter is commonly set to $0.3$ for the experiments.
}
\label{fig:cifar_d_03}
\end{figure}

Figure~\ref{fig:cifar_d_03} shows the convergence plots under massive clients with lower participation rates.
The result shows that FedACG takes the lead in most learning sections, which also demonstrates the effectiveness of FedACG.

\clearpage

\subsection{Evaluation on dynamic client set}
\begin{figure*}[h!]
\centering
\begin{subfigure}[b]{0.45\linewidth}
\centering
\includegraphics[width=1\linewidth]{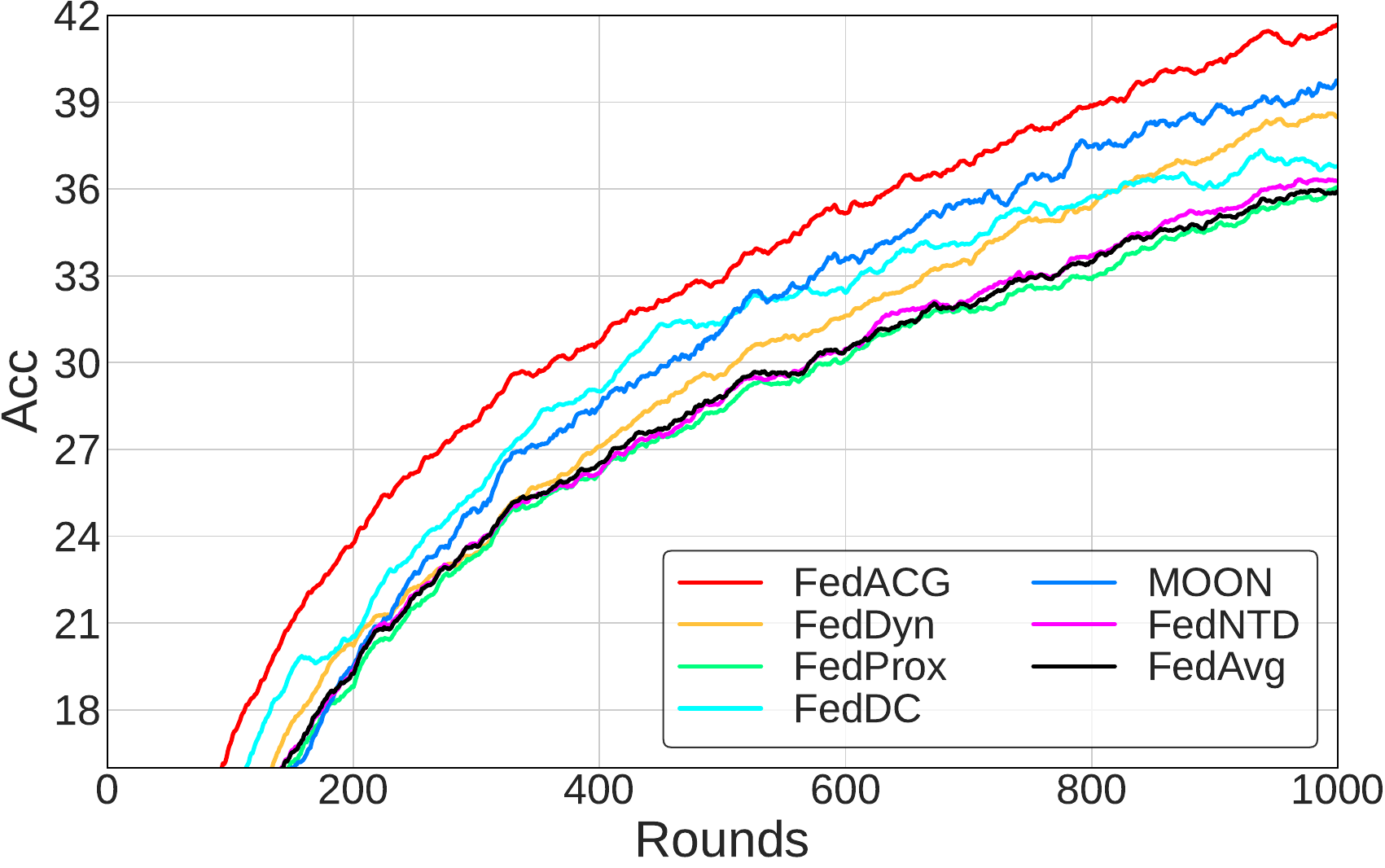}
\end{subfigure}
\caption{
The convergence plots of FedACG and other compared methods on CIFAR-100 when the client set changes over dynamically: we sample 250 clients out of 500 clients as a candidate clients set at every 100 rounds over 10 stages on Dirichlet (0.3) split.
10 clients out of the sampled client set participate for the local training for each communication round. 
}
\label{fig:incre_curve}
\end{figure*}

Figure~\ref{fig:incre_curve} shows a convergence plot when the entire client's pool changes during training. 
The result shows that FedACG outperforms the baselines in most learning sections.
Note that FedDyn shows worse performance than FedACG in the overall section of learning.
This is partly because it needs to store local states for local training in each client, which requires a kind of warm-up period for newly participating clients to contain useful information.
In contrast, FedACG, which is free from these restrictions, shows strength in a realistic federated learning scenario where the pool of entire clients changes during training.

\section{Convergence of FedACG}
\label{sec:convergence_proof}

We now present the theoretical convergence result of FedACG.
We first state a few assumptions for the local loss functions $\mathcal{F}_i(\cdot)$, which are commonly used in several previous works on federated
optimization~\citep{karimireddy2019scaffold, xu2021fedcm, reddi2021adaptive}.
First, the local function $\mathcal{F}_i(\cdot)$ is assumed to be $L$-smooth for all $C_i \in \{C_1,\dots,C_N\}$, \ie,
\begin{align}
\|\nabla \mathcal{F}_i ({x}) -\nabla \mathcal{F}_i ({ y})\|\leq
L \|{ x} - { y}\|\ \ \forall { x}, { y}. 
\end{align}
This also implies 
\begin{align}
\mathcal{F}_i(y) \leq \mathcal{F}_i(x) + \langle \nabla \mathcal{F}_i(x), y - x \rangle + \frac{L}{2} \| y-x\|^2.
\label{as:smooth_convex_grad}
\end{align}
%
Second, we assume the stochastic gradient of the local loss function $\nabla {f}_i(x) :=  \nabla {\mathcal{F}}_i(x; \mathcal{D}_i)$ is unbiased and possesses a bounded variance,~\ie $\mathbb{E}_{\mathcal{D}_i} [ \| \nabla f_i ({ x}) - \nabla \mathcal{F}_i(x) \|^2] \leq \sigma^{2}$. Third, we assume the average norm of local gradients is bounded by a function of the global gradient magnitude as $\frac{1}{N} \sum_{i=1}^N\left\|\nabla \mathcal{F}_i(x)\right\|^2 \leq \sigma _{g}^2+B^2\|\nabla \mathcal{F}(x)\|^2$, where $\sigma_g \geq 0$ and $B \geq 1$.
Based on the above assumptions, we derive the following asymptotic convergence bound of FedACG.

\subsection{Preliminary Lemmas}

We present several technical lemmas that are useful for subsequent proofs.

\begin{lemma}[relaxed triangle inequality]\label{lem:norm-sum}
    Let $\{v_1,\dots, v_\tau\}$ be $\tau$ vectors in $\mathbb{R}^d$. Then the following are true:
(1) $\|v_i + v_j\|^2 \leq (1 + a)\|v_i\|^2 + (1 + \tfrac{1}{a})\|v_j\|^2$ for any $a >0$, and
(2) $\|\sum_{i=1}^\tau v_i\|^2 \leq \tau \sum_{i=1}^\tau\|v_i\|^2$.
\end{lemma}

\begin{lemma}[sub-linear convergence rate]\label{lemma:general}
For every non-negative sequence $\{d_{r-1}\}_{r \geq 1}$ and any
parameters $\eta_{\max}  \geq 0$, $c \geq 0$,
$R \geq 0$, there exists a constant step-size
$\eta \leq \eta_{\max}$ and weights $w_r  = 1$ such that,
\begin{align*}
    \Psi_R := \frac{1}{R+1}\sum_{r=1}^{R+1} \left( \frac{d_{r-1}}{\eta}  - \frac{d_r}{\eta} + c_1 \eta +c_2 \eta^2\right) \leq  \frac{d_0}{\eta_{\max}(R+1)} + \frac{2\sqrt{c_1 d_0}}{\sqrt{R +1}} + 2\left(\frac{d_0}{R+1}\right)^{\frac{2}{3}} c_2^{\frac{1}{3}}\,.
\end{align*}
\end{lemma}
\begin{proof} Unrolling the sum, we can simplify
\[
    \Psi_R \leq \frac{d_0}{\eta (R+1)} + c_1 \eta + c_2 \eta^2\,.
\]
The lemma can be established through the adjustment of $\eta$. 
We consider the following two cases based on the magnitudes of R and $\eta_{\max}$:
\begin{itemize}
    \item When $R+1 \leq \frac{d_0}{c_1 \eta_{\max}^2}$ and $R+1 \leq \frac{d_0}{c_2 \eta_{\max}^3}$, selecting $\eta = \eta_{\max}$ satisfies
    \[
        \Psi_R \leq \frac{d_0}{\eta_{\max} (R+1)} + c_1 \eta_{\max} + c_2 \eta_{\max}^2 \leq \frac{d_0}{\eta_{\max} (R+1)} + \frac{\sqrt{c_1 d_0}}{\sqrt{R +1}}+ \left(\frac{d_0}{R+1}\right)^{\frac{2}{3}} c_2^{\frac{1}{3}} \,.
    \]
    \item In the other case, we have $\eta_{\max}^2 \geq \frac{d_0}{c_1(R+1)}$ or $\eta_{\max}^3 \geq \frac{d_0}{c_2(R+1)}$. Choosing $\eta = \min\left\{\sqrt{\frac{d_0}{c_1(R+1)}}, \sqrt[3]{\frac{d_0}{c_2(R+1)}}\right\}$ satisfies
    \[
        \Psi_R \leq \frac{d_0}{\eta (R+1)} + c \eta = \frac{2\sqrt{c_1 d_0}}{\sqrt{R +1}} + 2\sqrt[3]{\frac{d_0^2 c_2}{(R+1)^2}} \,.\vspace{-5mm}
    \]
\end{itemize}
\end{proof}

\begin{lemma}[separating mean and variance]\label{lem:independent}
Given a set of $\tau$ random variables $\{\mathbf{x}_1,\dots,\mathbf{x}_{\tau}\}$ in $\mathbb{R}^d$, where $\mathbb{E}[\mathbf{x}_i | \mathbf{x}_{i-1}, \dots \mathbf{x}_{1}] = \xi_i$ and $\mathbb{E}[\|\mathbf{x}_i - \xi_i\|^2]\leq \sigma^2$ represent their conditional mean and variance, respectively, the variables $\{\mathbf{x}_i - \xi_i\}$ form a martingale difference sequence.
Based on this setup, the following holds
\[
    \mathbb{E}[\|\sum_{i=1}^\tau \mathbf{x}_i\|^2] \leq 2\|\sum_{i=1}^\tau \xi_i\|^2+ 2\tau \sigma^2\,.
\]
\end{lemma}
\noindent\textit{Proof.} 
\begin{align}
     \mathbb{E}[\|\sum_{i=1}^\tau \mathbf{x}_i\|^2] 
     & \leq 2\|\sum_{i=1}^\tau \xi_i\|^2 +  2\mathbb{E}[\|\sum_{i=1}^\tau \mathbf{x}_i - \xi_i\|^2]  \nonumber \\
     & = 2\|\sum_{i=1}^\tau \xi_i\|^2 + 2 \sum_{i} \mathbb{E}[\|\mathbf{x}_i - \xi_i\|^2] \nonumber \\
     & \leq 2\|\sum_{i=1}^\tau \xi_i\|^2 + 2 \tau \sigma^2\,.
\end{align}

\noindent The first inequality comes from the relaxed triangle inequality and the following equality holds because $\{\mathbf{x}_i - \xi_i\}$ forms a martingale difference sequence. $\square$

\subsection{Convergence of FedACG for non-convex functions}

\begin{theorem}
\textbf{(Convergence for non-convex functions)} Suppose that local functions $  \lbrace \mathcal {F}_i \rbrace _{i=1}^{N}$ are non-convex and $L$-smooth.
By setting $\eta \leq \frac{(1-\lambda)^2}{64KL(B^2+1)}$, FedACG satisfies
\begin{align}
&\min_{t=1, \ldots, T} \mathbb{E}\left\|\nabla \mathcal{F}\left(\theta^{t-1} + \lambda m^{t-1} \right)\right\|^2 \nonumber \\ 
 &  \hspace{0.0em} \leq \mathcal{O}\left(\frac{ M_1 \sqrt{LD}}{\sqrt{T K |S_t|}}
 +\frac{\left(LD(1-\lambda)^2\right)^{\frac{2}{3}}M_2^{\frac{1}{3}}}{(T+1)^{\frac{2}{3}}}
 +\frac{B^2 L D}{T}\right), \nonumber
 \end{align}
where $M_1^2:=\sigma^2+K\left(1-\frac{|S_t|}{N}\right) \sigma_g^2$, $M_2:=\frac{\sigma^2}{K} + \sigma_g^2$, and $D:=\frac{\mathcal{F}\left(\theta^0\right)-\mathcal{F}(\theta^{\ast})}{1-\lambda}$.
\end{theorem}

\noindent \textit{Proof.} Let $z^t = \theta^{t} + \frac{\lambda}{1-\lambda} m^{t}$ and $\Phi^t = \theta^t + \lambda m^t$. Note that $z^0 = \theta^0$ and $z^t - z^{t-1} = \frac{1}{1-\lambda}\Delta^t$.
By the smoothness of the function $\mathcal{F}(\mathbf{x})$, we have
\begin{align*}
\mathcal{F}(z^{t})
&\leq \mathcal{F}(z^{t-1}) + \langle\nabla\mathcal{F}(z^{t-1}), z^{t} - z^{t-1} \rangle +\frac{L}{2}\|z^{t} - z^{t-1}\|^2.
\end{align*}
%
%
By taking the expectation on both sides, we have
\begin{align*}
\mathbb{E}[\mathcal{F}(z^{t})] 
& \leq \mathbb{E}[\mathcal{F}(z^{t-1})] + \frac{1}{1-\lambda} \mathbb{E}[\langle\nabla\mathcal{F}(z^{t-1}), \Delta^{t} \rangle] +\frac{L}{2}  \mathbb{E}[\|z^{t} - z^{t-1}\|^2] \\
& = \mathbb{E}[\mathcal{F}(z^{t-1})] + \frac{1}{1-\lambda} \mathbb{E}[ \langle\nabla\mathcal{F}(z^{t-1}) - \nabla\mathcal{F}(\Phi^{t-1}), \Delta^{t} \rangle] + \frac{1}{1-\lambda} \mathbb{E}[ \langle\nabla\mathcal{F}(\Phi^{t-1}), \Delta^{t} \rangle] +\frac{L}{2(1-\lambda)^2}\mathbb{E}[\|\Delta^t\|^2]. \numberthis \label{eq:savepoint}
\end{align*}
We note that
\begin{align*}
\frac{1}{1-\lambda} \mathbb{E}[ \langle\nabla\mathcal{F}(z^{t-1}) - \nabla\mathcal{F}(\Phi^{t-1}), \Delta^{t} \rangle] &
\leq \frac{1-\lambda}{2\lambda^{3} L} \mathbb{E}[\|\nabla\mathcal{F}(z^{t-1}) - \nabla\mathcal{F}(\Phi^{t-1})\|^2] + \frac{\lambda^{3} L}{2(1-\lambda)^3}\mathbb{E}[\|\Delta^{t}\|^2] \\
& \leq \frac{(1-\lambda)L}{2\lambda^{3}}\mathbb{E}[\|z^{t-1} - \Phi^{t-1}\|^2] + \frac{\lambda^{3} L}{2(1-\lambda)^3}\mathbb{E}[\|\Delta^{t}\|^2] \\
& \leq \frac{L}{2(1 -\lambda)}\mathbb{E}[\|m^{t-1}\|^2] + \frac{L}{2(1-\lambda)^3} \mathbb{E}[\|\Delta^{t}\|^2], \numberthis \label{eq:angle_diffmodel}
\end{align*}
where the first inequality holds because $\langle a, b \rangle \leq \frac{1}{2} (\|a\|^{2} + \|b\|^{2})$, while the second inequality follows from the $L$-smoothness.
The third inequality follows because $z^t - \Phi^t = \frac{\lambda^{2}}{1-\lambda}m^t$ and $0 \leq \lambda < 1$. \\
%
%

\noindent We also note that
\begin{align*}
\frac{1}{1-\lambda} \mathbb{E}[ \langle\nabla\mathcal{F}(\Phi^{t-1}), \Delta^{t} \rangle ]
& = \frac{1}{1-\lambda} \mathbb{E}[\langle\nabla\mathcal{F}(\Phi^{t-1}), \frac{-\eta K}{KN}\sum_{k,C_i} \nabla \mathcal{F}_i(\theta^t_{i,k-1}) \rangle] \\
& \leq \frac{\eta K }{2(1-\lambda)} \bigg(\mathbb{E}[\| \nabla\mathcal{F}(\Phi^{t-1}) - \frac{1}{K N} \sum_{k,C_i} \nabla \mathcal{F}_i(\theta^t_{i,k-1})\|^2] -\mathbb{E}[\|\nabla\mathcal{F}(\Phi^{t-1})\|^2] \bigg) \\
& \leq \frac{\eta K }{2(1-\lambda)}  \bigg(\frac{L^2}{KN} \sum_{k,C_i}\mathbb{E}[\| \theta^t_{i,k-1} - \theta^t_{i,0} \|^2] -\mathbb{E}[\|\nabla\mathcal{F}(\Phi^{t-1})\|^2] \bigg),
 \numberthis \label{eq:angle_globallocal}
\end{align*}
where the first inequality holds because $\langle a, b \rangle \leq \frac{1}{2} \|a+b\|^{2} - \frac{1}{2} \|a\|^2$. \\

\noindent Substituting Eq.~\eqref{eq:angle_diffmodel} and Eq.~\eqref{eq:angle_globallocal} into Eq.~\eqref{eq:savepoint} yields
\begin{align*}
\mathbb{E}[\mathcal{F}(z^{t})] 
& \leq \mathbb{E}[\mathcal{F}(z^{t-1})]  +  \frac{\eta K }{2(1-\lambda)} \bigg(\frac{L^2}{KN} \sum_{k,C_i}\mathbb{E}[\| \theta^t_{i,k-1} - \theta^t_{i,0} \|^2] -\mathbb{E}[\|\nabla\mathcal{F}(\Phi^{t-1})\|^2] \bigg) \\
& \quad + \frac{L}{2(1 -\lambda)}\mathbb{E}[\|m^{t-1}\|^2]+(\frac{L}{2(1-\lambda)^3} + \frac{L}{2(1-\lambda)^2}) \mathbb{E}[\|\Delta^{t}\|^2].
\end{align*}
By rearranging the inequality above, we have
\begin{align*}
\frac{\eta K }{2(1-\lambda)}\mathbb{E}[\|\nabla\mathcal{F}(\Phi^{t-1})\|^2]  
& \leq (\mathbb{E}[\mathcal{F}(z^{t-1})] - \mathbb{E}[\mathcal{F}(z^{t})]) +  \frac{\eta K L^2}{2(1-\lambda)}\frac{1}{KN} \sum_{k,C_i}\mathbb{E}[\| \theta^t_{i,k-1} - \theta^t_{i,0} \|^2]  \\
& \quad+ \frac{L}{2(1 -\lambda)}\mathbb{E}[\|m^{t-1}\|^2] +(\frac{ L}{2(1-\lambda)^3} + \frac{L}{2(1-\lambda)^2}) \mathbb{E}[\|\Delta^{t}\|^2]. 
\end{align*}
Summing the above inequality for $t \in \{ 1,\dots, T\}$ yields
\begin{align*}
\frac{\eta K }{2(1-\lambda)}\sum\limits_{t=1}^{T}\mathbb{E}[\|\nabla\mathcal{F}(\Phi^{t-1})\|^2]  
& \leq (\mathbb{E}[\mathcal{F}(z^{0})] - \mathbb{E}[\mathcal{F}(z^{T})])  +  \frac{\eta K L^2}{2(1-\lambda)} \sum\limits_{t=1}^{T} \frac{1}{KN}\sum_{k,C_i}\mathbb{E}[\| \theta^t_{i,k-1} - \theta^t_{i,0} \|^2] \\
& \quad+ \frac{ L}{2(1 -\lambda)}\sum\limits_{t=1}^{T}\mathbb{E}[\|m^{t-1}\|^2]+(\frac{ L}{2(1-\lambda)^3} + \frac{L}{2(1-\lambda)^2})\sum\limits_{t=1}^{T} \mathbb{E}[\|\Delta^{t}\|^2]. 
\end{align*}
By applying Lemma \ref{lem:m^t},  Lemma \ref{lem:IV_nonconvex}, and Lemma \ref{lem:I^t}, we have
\begin{align*}
\frac{\eta K }{2(1-\lambda)}\sum\limits_{t=1}^{T}\mathbb{E}[\|\nabla\mathcal{F}(\Phi^{t-1})\|^2]  
& \leq (\mathbb{E}[\mathcal{F}(z^{0})] - \mathbb{E}[\mathcal{F}(z^{T})])  +  \frac{\eta K L^2}{2(1-\lambda)} \sum\limits_{t=1}^{T} \frac{1}{KN}\sum_{k,C_i}\mathbb{E}[\| \theta^t_{i,k-1} - \theta^t_{i,0} \|^2] \\
& \quad+(\frac{2 L}{2(1-\lambda)^3} + \frac{L}{2(1-\lambda)^2})\sum\limits_{t=1}^{T} \mathbb{E}[\|\Delta^{t}\|^2] \\
& \leq (\mathbb{E}[\mathcal{F}(z^{0})] - \mathbb{E}[\mathcal{F}(z^{T})])  \\
& \quad +  \frac{\eta K }{2(1-\lambda)}L^2(\frac{8\eta K L}{(1-\lambda)^2} + \frac{4\eta K L}{1-\lambda} + 1) \sum\limits_{t=1}^{T} \frac{1}{KN}\sum_{k,C_i}\mathbb{E}[\| \theta^t_{i,k-1} - \theta^t_{i,0} \|^2] \\
& \quad + \frac{\eta K}{2(1-\lambda)} (\frac{4\eta K L}{(1-\lambda)^2} + \frac{2\eta K L}{1-\lambda})\sum\limits_{t=1}^{T}\bigg(4(B^2 + 1) \|\nabla \mathcal{F}(\Phi^{t-1})\|^2 + \frac{4(1 - \frac{|S_t|}{N})}{ |S_t|}\sigma_g^2+ \frac{\sigma^2}{K |S_t|}\bigg)\\
 & \leq (\mathbb{E}[\mathcal{F}(z^{0})] - \mathbb{E}[\mathcal{F}(z^{T})])  \\
& \quad +  \frac{\eta K }{2(1-\lambda)}L^2(\frac{8\eta K L}{(1-\lambda)^2} + \frac{4\eta K L}{1-\lambda} + 1) \sum\limits_{t=1}^{T} \bigg(6 \eta^{2} K^{2}\left(\delta_{g}^{2}+B^{2}\mathbb{E}[\|\nabla \mathcal{F}\left(\Phi^{t-1}\right)\|^{2}]\right)+3 \eta^{2} K^{} \sigma^2\bigg) \\
& \quad + \frac{\eta K}{2(1-\lambda)} (\frac{4\eta K L}{(1-\lambda)^2} + \frac{2\eta K L}{1-\lambda})\sum\limits_{t=1}^{T}\bigg(4(B^2 + 1) \|\nabla \mathcal{F}(\Phi^{t-1})\|^2 + \frac{4(1 - \frac{|S_t|}{N})}{ |S_t|}\sigma_g^2+ \frac{\sigma^2}{K |S_t|}\bigg).
\end{align*}
If $\eta \leq \frac{(1-\lambda)^2}{64KL(B^2+1)}$, we can rewrite the above inequality as follows
\begin{align*}
\frac{\eta K }{4(1-\lambda)}\sum\limits_{t=1}^{T}\mathbb{E}[\|\nabla\mathcal{F}(\Phi^{t-1})\|^2]  & \leq (\mathbb{E}[\mathcal{F}(z^{0})] - \mathbb{E}[\mathcal{F}(z^{T})])   +  \frac{\eta K }{2(1-\lambda)}L^2(\frac{8\eta K L}{(1-\lambda)^2} + \frac{4\eta K L}{1-\lambda} + 1) \sum\limits_{t=1}^{T} \bigg(6 \eta^{2} K^{2}\delta_{g}^{2} +3 \eta^{2} K^{} \sigma^2\bigg) \\
& \quad + \frac{\eta K}{2(1-\lambda)} (\frac{4 \eta K L}{(1-\lambda)^2} + \frac{2\eta K L}{1-\lambda})\sum\limits_{t=1}^{T}\bigg(\frac{4(1 - \frac{|S_t|}{N})}{ |S_t|}\sigma_g^2+ \frac{\sigma^2}{K |S_t|}\bigg) \\
& \leq (\mathbb{E}[\mathcal{F}(z^{0})] - \mathbb{E}[\mathcal{F}(z^{T})])  +  35L^2 (1-\lambda)^5 \bigg(\frac{\eta K}{4(1-\lambda)^2}\bigg)^3 \sum\limits_{t=1}^{T} \bigg(6 \delta_{g}^{2} +\frac{3 }{K} \sigma^2\bigg) \\
& \quad + 8L(1-\lambda)\bigg(\frac{\eta K}{4(1-\lambda)^2}\bigg)^2 \sum\limits_{t=1}^{T}\bigg(\frac{4(1 - \frac{|S_t|}{N})}{ |S_t|}\sigma_g^2+ \frac{\sigma^2}{K |S_t|}\bigg).
\end{align*}

\noindent Let $\tilde{\eta} = \frac{\eta K }{4(1-\lambda)^2}$. By dividing both sides by $1-\lambda$, we have
\begin{align*}
\Tilde{\eta}\sum\limits_{t=1}^{T}\mathbb{E}[\|\nabla\mathcal{F}(\Phi^{t-1})\|^2] & \leq \frac{(\mathbb{E}[\mathcal{F}(z^{0})] - \mathbb{E}[\mathcal{F}(z^{T})])}{1-\lambda}  +  35L^2 (1-\lambda)^4 \Tilde{\eta}^3 T \bigg(6 \delta_{g}^{2} +\frac{3 }{K} \sigma^2\bigg) \\
& \quad + 8L\Tilde{\eta}^2 T\bigg(\frac{4(1 - \frac{|S_t|}{N})}{ |S_t|}\sigma_g^2+ \frac{\sigma^2}{K |S_t|}\bigg).
\end{align*}
Dividing both side by $\Tilde{\eta}T$ yields
\begin{align*}
\frac{1}{T}\sum\limits_{t=1}^{T}\mathbb{E}[\|\nabla\mathcal{F}(\Phi^{t-1})\|^2] &\leq \frac{(\mathbb{E}[\mathcal{F}(\theta^{0})] - \mathbb{E}[\mathcal{F}(\theta^{*})])}{\Tilde{\eta}T(1-\lambda)}  +  35L^2 (1-\lambda)^4 \Tilde{\eta}^2   \bigg(6 \delta_{g}^{2} +\frac{3 }{K} \sigma^2\bigg) \\
& \quad + 8L\Tilde{\eta} \bigg(\frac{4(1 - \frac{|S_t|}{N})}{ |S_t|}\sigma_g^2+ \frac{\sigma^2}{K |S_t|}\bigg).
\end{align*}
Now we get the desired rate by applying Lemma \ref{lemma:general}, which finishes the proof. $\square$

\begin{lemma}\label{lem:m^t}
Algorithm~\ref{alg:proposed_method} satisfies
\begin{align*}
\sum\limits_{t=1}^{T} \mathbb{E}[\left\|m^{t}\right\|^{2}] \leq \frac{1}{(1-\lambda)^2} \sum_{t=1}^{T} \mathbb{E}[ \| \Delta^t\|^2 ]
\end{align*}
\end{lemma}

\noindent \textit{Proof.} Unrolling the recursion of the momentum $m^t$,~\ie, $m^t = \sum_{r=1}^{t} \lambda^{t-r}\Delta^r$
\begin{align*}
\mathbb{E}[\| m^{t}\|^{2}]&
=\mathbb{E} [\| \sum_{r=1}^{t} \lambda^{t-r}\Delta^r \|^2].
\end{align*}
Let $\Gamma_{t}=\sum_{r=0}^{t-1} \lambda^r=\frac{1-\lambda^{t}}{1-\lambda}$. Since $0 \leq \lambda < 1$, $\Gamma_{t} \leq \frac{1}{1-\lambda}$, we have
\begin{align*}
\mathbb{E}[\| \sum_{r=1}^{t} \lambda^{t-r}\Delta^r \|^2] & =  \Gamma_t^2 \mathbb{E}[ \| \frac{1}{\Gamma_t} \sum_{r=1}^{t} \lambda^{t-r}\Delta^r \|^2] \\
& \leq  \Gamma_t  \sum_{r=1}^{t} \lambda^{t-r} \mathbb{E} [\| \Delta^r\|^2] 
\\ &\leq \frac{1}{1-\lambda} \sum_{r=1}^{t} \lambda^{t-r} \mathbb{E} [\| \Delta^r\|^2] .
\end{align*}
By summing the above inequality for $t \in  \{0,\dots,T-1\}$, we have
\begin{align*}
\sum\limits_{t=1}^{T} \mathbb{E} \! \left[\left\|m^{t}\right\|^{2}\right]
&\leq \sum_{t=1}^{T} \frac{1}{1-\lambda} \sum_{r=1}^{t} \lambda^{t-r} \mathbb{E} [\| \Delta^r\|^2]  \\
&\leq \frac{1}{(1-\lambda)^2} \sum_{t=1}^{T} \mathbb{E}[ \| \Delta^t\|^2 ],
\end{align*}
which finishes the proof. $\square$

\begin{lemma}\label{lem:IV_nonconvex}
For all $t \geq 1$, Algorithm~\ref{alg:proposed_method}  satisfies
\begin{align*}
\mathbb{E}[\|\Delta^{t}\|^{2}] \leq 2(\eta K)^2\bigg(\frac{2L^2}{K N}\sum\limits_{k,C_i } \mathbb{E} [\|  \theta^{t}_{i,k} - \theta^{t}_{i,0}\|^2] + 4(B^2 + 1) \|\nabla \mathcal{F}(\Phi^{t-1})\|^2 + \frac{4(1 - \frac{|S_t|}{N})}{ |S_t|}\sigma_g^2+ \frac{\sigma^2}{K |S_t|}\bigg),
\end{align*}
where $\theta^{t}_{i,0}$ denotes the initial point for the local model of the $i$-th client, i.e., $\theta^{t}_{i,0} = \Phi^{t-1}$.
\end{lemma}

\noindent \textit{Proof.} By applying Lemma~\ref{lem:independent}, we have
\begin{align*}
\mathbb{E} [\|\Delta^{t}\|^{2}] &= \mathbb{E} [\| \frac{\eta K}{K |S_t|}\sum\limits_{k,C_i \in S_t} \nabla f_i(\theta^{t}_{i,k}) \|^2] 
\\ & \leq 2(\eta K)^2\big(\mathbb{E} [\| \frac{1}{K |S_t|}\sum\limits_{k,C_i \in S_t} \nabla \mathcal{F}_i(\theta^{t}_{i,k}) \|^2]  + \frac{\sigma^2}{K |S_t|}\big),
\end{align*}
\noindent We note that
\begin{align*}
&\mathbb{E} [\| \frac{1}{K |S_t|}\sum\limits_{k,C_i \in S_t} \nabla \mathcal{F}_i(\theta^{t}_{i,k}) \|^2] 
\\ & = \mathbb{E} [\| \frac{1}{K |S_t|}\sum\limits_{k,C_i \in S_t} \left(\nabla \mathcal{F}_i(\theta^{t}_{i,k}) -\nabla \mathcal{F}_i(\theta^{t}_{i,0}) + \nabla \mathcal{F}_i(\theta^{t}_{i,0}) \right)\|^2] 
\\ & \leq 2 \mathbb{E} [\| \frac{1}{K |S_t|}\sum\limits_{k,C_i \in S_t}\left( \nabla \mathcal{F}_i(\theta^{t}_{i,k}) -\nabla \mathcal{F}_i(\theta^{t}_{i,0})\right)\|^2] + 2 \mathbb{E}[\|\frac{1}{ |S_t|}\sum\limits_{C_i \in S_t}\nabla \mathcal{F}_i(\theta^{t}_{i,0}) \|^2] 
\\ & \leq \frac{2}{K N}\sum\limits_{k,C_i} \mathbb{E} [\|  \nabla \mathcal{F}_i(\theta^{t}_{i,k}) -\nabla \mathcal{F}_i(\theta^{t}_{i,0})\|^2] + \mathbb{E}[\|\frac{2}{ |S_t|}\sum\limits_{C_i \in S_t}\left( \nabla \mathcal{F}_i(\theta^{t}_{i,0}) -\nabla \mathcal{F}(\Phi^{t-1}) +\nabla \mathcal{F}(\Phi^{t-1}) \right) \|^2] 
\\ & \leq \frac{2L^2}{K N}\sum\limits_{k,C_i } \mathbb{E} [\|  \theta^{t}_{i,k} - \theta^{t}_{i,0}\|^2] + \mathbb{E}[\|\frac{2}{ |S_t|}\sum\limits_{C_i \in S_t} \left(\nabla \mathcal{F}_i(\theta^{t}_{i,0}) -\nabla \mathcal{F}(\Phi^{t-1}) +\nabla \mathcal{F}(\Phi^{t-1})\right) \|^2] 
\\  & \leq \frac{2L^2}{K N}\sum\limits_{k,C_i } \mathbb{E} [\|  \theta^{t}_{i,k} - \theta^{t}_{i,0}\|^2] + 4 \|\nabla \mathcal{F}(\Phi^{t-1})\|^2 + \frac{4(1 - \frac{|S_t|}{N})}{ |S_t|N}\sum\limits_{C_i}\|\nabla \mathcal{F}_i(\theta^{t}_{i,0}) \|^2 
\\ & \leq \frac{2L^2}{K N}\sum\limits_{k,C_i } \mathbb{E} [\|  \theta^{t}_{i,k} - \theta^{t}_{i,0}\|^2] + 4(B^2 + 1) \|\nabla \mathcal{F}(\Phi^{t-1})\|^2 + \frac{4(1 - \frac{|S_t|}{N})}{ |S_t|}\sigma_g^2,
\end{align*}
where, in the fourth inequality, the improvement of $(1 - \frac{|S_t|}{N})$ follows from sampling the active client set $S_t$ without replacement at the $t$-th communication round.
The last inequality holds because the average norm of local gradients is bounded as $\frac{1}{N} \sum_{i=1}^N\left\|\nabla \mathcal{F}_i(x)\right\|^2 \leq \sigma _{g}^2+B^2\|\nabla \mathcal{F}(x)\|^2$, which concludes the proof. $\square$

\begin{lemma}\label{lem:I^t}
For all $t \geq 1$, we have
\begin{align*}
\frac{1}{KN} \sum_{k,C_i}
\mathbb{E}[\| \theta^t_{i,k} -\theta_{i, 0}^{t}\|^2] \leq 6 \eta^{2} K^{2}\left(\delta_{g}^{2}+B^{2}\mathbb{E}[\|\nabla \mathcal{F}\left(\Phi^{t-1}\right)\|^{2}]\right)+3 \eta^{2} K^{} \sigma^2.
\end{align*}
\end{lemma}

\noindent \textit{Proof.} We first define the following terms as
\begin{align*}
I_{i, k}^{t} =\mathbb{E}[\|\theta_{i, k}^{t}-\theta_{i, 0}^{t}\|^{2}], \
I_{i}^{t}=\frac{1}{K}\sum_{k=1}^{K} I_{i, k}^{t}, \
I^{t} =\frac{1}{N} \sum_{C_i} I_{i}^{t}. \numberthis \label{i_def}
\end{align*}
Initially, we commence by deriving an upper bound for the variable $I_{i,k}^t$ as
\begin{align}
I_{i,k}^t
& =\mathbb{E}[\|\theta_{i, k}^{t}-\theta_{i, 0}^{t}\|^{2}] \nonumber \\
& = \mathbb{E}[\|\theta_{i, k-1}^{t}-\theta_{i, 0}^{t}-\eta \nabla {f}_{i}\left(\theta_{i, k-1}^{t}\right)\|^{2}] \nonumber \\
& = \mathbb{E}[\|\theta_{i, k-1}^{t}-\theta_{i, 0}^{t} -\eta \nabla {\mathcal{F}}_{i}\left(\theta_{i, k-1}^{t}\right) +\eta \nabla {\mathcal{F}}_{i}\left(\theta_{i, k-1}^{t}\right) -\eta \nabla {f}_{i}\left(\theta_{i, k-1}^{t}\right)\|^{2}] \nonumber \\
& \leq \mathbb{E}[\|\theta_{i, k-1}^{t}-\theta_{i, 0}^{t} -\eta \nabla {\mathcal{F}}_{i}\left(\theta_{i, k-1}^{t}\right) \|^{2}] + \eta^{2}\sigma^2 \nonumber \\
& \leq\left(1+\frac{1}{K-1}\right) \mathbb{E}[\|\theta_{i, k-1}^{t}-\theta_{i, 0}^{t}\|^{2}]+K \eta^{2}\mathbb{E}[\|\nabla \mathcal{F}_{i}\left(\theta_{i, k-1}^{t}\right)\|^{2}]+\eta^{2} \sigma^2 \label{eq:I_definition},
\end{align}
where the first inequality follows because the stochastic gradient possesses a bounded variance, while the second inequality follows from the Lemma~\ref{lem:norm-sum}. \\

\noindent We note that
\begin{align*}
\mathbb{E}[\|\nabla \mathcal{F}_{i}\left(\theta_{i, k-1}^{t}\right)\|^{2}] 
& = \mathbb{E}[\|\nabla \mathcal{F}_{i}\left(\theta_{i, k-1}^{t}\right) - \nabla \mathcal{F}_{i}\left(\theta_{i, 0}^{t}\right) + \nabla \mathcal{F}_{i}\left(\theta_{i, 0}^{t}\right)\|^{2}]\\
&\leq 2\mathbb{E}[\|\nabla \mathcal{F}_{i}\left(\theta_{i, k-1}^{t}\right) - \nabla \mathcal{F}_{i}\left(\theta_{i, 0}^{t}\right)\|^{2}] + 2\mathbb{E}[\|\nabla \mathcal{F}_{i}\left(\theta_{i, 0}^{t}\right)\|^{2}] \\
&\leq 2L^2\mathbb{E}[\|\theta_{i, k-1}^{t} - \theta_{i, 0}^{t}\|^{2}] + 2\mathbb{E}[\|\nabla \mathcal{F}_{i}\left(\theta_{i, 0}^{t}\right)\|^{2}]. \numberthis \label{eq:local_norm}
\end{align*}
%
By substituting Eq.~\eqref{eq:local_norm} into Eq. (\ref{eq:I_definition}), we have
\begin{align*}
I_{i,k}^t
& \leq\left(1+\frac{1}{K-1}+2 K \eta^{2} L^{2}\right) \mathbb{E}[\|\theta_{i, k-1}^{t}-\theta_{i, 0}^{t}\|^{2}]+2 K \eta^2\mathbb{E}[\|\nabla \mathcal{F}_{i}\left(\theta_{i,0}^{t}\right)\|^{2}]+\eta^{2} \sigma^2 \\
& \leq\left(1+\frac{1}{K-1}+2 K \eta^{2} L^{2}\right) I_{i, k-1}^t+2 K \eta^2\mathbb{E}[\|\nabla \mathcal{F}_{i}\left(\theta_{i,0}^{t}\right)\|^{2}]+\eta^{2} \sigma^2.
\end{align*}
By unrolling the recursion, we have
\begin{align*}
I_{i, k}^{t} 
\leq \sum_{r=0}^{k-1}\left(2 K \eta^{2} \mathbb{E}[\|\nabla \mathcal{F}_{i}\left(\theta_{i,0}^{t}\right)\|^{2}]+\eta^{2} \sigma^2\right)\left(1+\frac{2}{K-1}\right)^{r} 
\leq 3K\left(2 K \eta^{2} \mathbb{E}[\|\nabla \mathcal{F}_{i}\left(\theta_{i, 0}^{t}\right)\|^{2}]+\eta^{2} \sigma^2\right).
\end{align*}
By the definitions in Eq.~\eqref{i_def}, we have
\begin{align*}
I_{i}^{t} & =\frac{1}{K}\sum_{k=1}^{K} I_{i, k}^{t} \leq 3 K^{}\left(2 K \eta^{2} \mathbb{E}[\|\nabla \mathcal{F}_{i}\left(\theta_{i, 0}^{t}\right)\|^{2}]+\eta^{2} \sigma^2\right) \\
& =6\eta^{2} K^{2}\mathbb{E}[\|\nabla \mathcal{F}_{i}\left(\theta_{i, 0}^{t}\right)\|^{2}]+3 \eta^{2} K^{} \sigma^2.
\end{align*}

\begin{align*}
I^{t} &= 6 \eta^{2} K^{2} \frac{1}{N} \sum_{C_i} \mathbb{E}[\|\nabla \mathcal{F}_{i}\left(\theta_{i, 0}^{t}\right)\|^{2}]+ 3 \eta^{2} K^{} \sigma^2 \\
& \leq 6 \eta^{2} K^{2}\left(\delta_{g}^{2}+B^{2}\mathbb{E}[\|\nabla \mathcal{F}\left(\Phi^{t-1}\right)\|^{2}]\right)+3 \eta^{2} K^{} \sigma^2,
\end{align*}
where the inequality follows due to the assumption that the average norm of the local gradients is bounded,~\ie, $\frac{1}{N} \sum_{i=1}^N\left\|\nabla \mathcal{F}_i(x)\right\|^2 \leq \sigma _{g}^2+B^2\|\nabla \mathcal{F}(x)\|^2$, which completes the proof. $\square$

%% file: main.bbl
\begin{thebibliography}{50}
\providecommand{\natexlab}[1]{#1}
\providecommand{\url}[1]{\texttt{#1}}
\expandafter\ifx\csname urlstyle\endcsname\relax
  \providecommand{\doi}[1]{doi: #1}\else
  \providecommand{\doi}{doi: \begingroup \urlstyle{rm}\Url}\fi

\bibitem[Acar et~al.(2021)Acar, Zhao, Matas, Mattina, Whatmough, and
  Saligrama]{acar2021federated}
Durmus Alp~Emre Acar, Yue Zhao, Ramon Matas, Matthew Mattina, Paul Whatmough,
  and Venkatesh Saligrama.
\newblock Federated learning based on dynamic regularization.
\newblock In \emph{ICLR}, 2021.

\bibitem[Al-Shedivat et~al.(2021)Al-Shedivat, Gillenwater, Xing, and
  Rostamizadeh]{al2020federated}
Maruan Al-Shedivat, Jennifer Gillenwater, Eric Xing, and Afshin Rostamizadeh.
\newblock Federated learning via posterior averaging: A new perspective and
  practical algorithms.
\newblock In \emph{ICLR}, 2021.

\bibitem[Basu et~al.(2020)Basu, Data, Karakus, and Diggavi]{basu2020qsparse}
Debraj Basu, Deepesh Data, Can Karakus, and Suhas~N Diggavi.
\newblock Qsparse-local-sgd: Distributed sgd with quantization, sparsification,
  and local computations.
\newblock \emph{IEEE Journal on Selected Areas in Information Theory},
  1\penalty0 (1):\penalty0 217--226, 2020.

\bibitem[Caldarola et~al.(2022)Caldarola, Caputo, and
  Ciccone]{caldarola2022improving}
Debora Caldarola, Barbara Caputo, and Marco Ciccone.
\newblock Improving generalization in federated learning by seeking flat
  minima.
\newblock In \emph{ECCV}, 2022.

\bibitem[Caldas et~al.(2019)Caldas, Duddu, Wu, Li, Kone{\v{c}}n{\`y}, McMahan,
  Smith, and Talwalkar]{caldas2018leaf}
Sebastian Caldas, Sai Meher~Karthik Duddu, Peter Wu, Tian Li, Jakub
  Kone{\v{c}}n{\`y}, H~Brendan McMahan, Virginia Smith, and Ameet Talwalkar.
\newblock Leaf: A benchmark for federated settings.
\newblock In \emph{NeurIPSW}, 2019.

\bibitem[Cutkosky and Orabona(2019)]{cutkosky2019momentum}
Ashok Cutkosky and Francesco Orabona.
\newblock Momentum-based variance reduction in non-convex sgd.
\newblock In \emph{NeurIPS}, 2019.

\bibitem[Das et~al.(2020)Das, Acharya, Hashemi, Sanghavi, Dhillon, and
  Topcu]{das2020faster}
Rudrajit Das, Anish Acharya, Abolfazl Hashemi, Sujay Sanghavi, Inderjit~S
  Dhillon, and Ufuk Topcu.
\newblock Faster non-convex federated learning via global and local momentum.
\newblock \emph{arXiv preprint arXiv:2012.04061}, 2020.

\bibitem[Finn et~al.(2017)Finn, Abbeel, and Levine]{finn2017model}
Chelsea Finn, Pieter Abbeel, and Sergey Levine.
\newblock Model-agnostic meta-learning for fast adaptation of deep networks.
\newblock In \emph{ICML}, 2017.

\bibitem[Foret et~al.(2021)Foret, Kleiner, Mobahi, and
  Neyshabur]{foret2020sharpness}
Pierre Foret, Ariel Kleiner, Hossein Mobahi, and Behnam Neyshabur.
\newblock Sharpness-aware minimization for efficiently improving
  generalization.
\newblock In \emph{ICLR}, 2021.

\bibitem[Gao et~al.(2022)Gao, Fu, Li, Chen, Xu, and Xu]{gao2022feddc}
Liang Gao, Huazhu Fu, Li Li, Yingwen Chen, Ming Xu, and Cheng-Zhong Xu.
\newblock Feddc: Federated learning with non-iid data via local drift
  decoupling and correction.
\newblock In \emph{CVPR}, 2022.

\bibitem[Haddadpour et~al.(2021)Haddadpour, Kamani, Mokhtari, and
  Mahdavi]{haddadpour2021federated}
Farzin Haddadpour, Mohammad~Mahdi Kamani, Aryan Mokhtari, and Mehrdad Mahdavi.
\newblock Federated learning with compression: Unified analysis and sharp
  guarantees.
\newblock In \emph{AISTATS}, 2021.

\bibitem[He et~al.(2016)He, Zhang, Ren, and Sun]{he2016deep}
Kaiming He, Xiangyu Zhang, Shaoqing Ren, and Jian Sun.
\newblock Deep residual learning for image recognition.
\newblock In \emph{CVPR}, 2016.

\bibitem[Hsieh et~al.(2020)Hsieh, Phanishayee, Mutlu, and
  Gibbons]{hsieh2020non}
Kevin Hsieh, Amar Phanishayee, Onur Mutlu, and Phillip Gibbons.
\newblock The non-iid data quagmire of decentralized machine learning.
\newblock In \emph{ICML}, 2020.

\bibitem[Hsu et~al.(2019)Hsu, Qi, and Brown]{hsu2019measuring}
Tzu-Ming~Harry Hsu, Hang Qi, and Matthew Brown.
\newblock Measuring the effects of non-identical data distribution for
  federated visual classification.
\newblock \emph{arXiv preprint arXiv:1909.06335}, 2019.

\bibitem[Karimireddy et~al.(2020)Karimireddy, Kale, Mohri, Reddi, Stich, and
  Suresh]{karimireddy2019scaffold}
Sai~Praneeth Karimireddy, Satyen Kale, Mehryar Mohri, Sashank~J Reddi,
  Sebastian~U Stich, and Ananda~Theertha Suresh.
\newblock Scaffold: Stochastic controlled averaging for on-device federated
  learning.
\newblock In \emph{ICML}, 2020.

\bibitem[Karimireddy et~al.(2021)Karimireddy, Jaggi, Kale, Mohri, Reddi, Stich,
  and Suresh]{karimireddy2021breaking}
Sai~Praneeth Karimireddy, Martin Jaggi, Satyen Kale, Mehryar Mohri, Sashank
  Reddi, Sebastian~U Stich, and Ananda~Theertha Suresh.
\newblock Breaking the centralized barrier for cross-device federated learning.
\newblock In \emph{NeurIPS}, 2021.

\bibitem[Khaled et~al.(2019)Khaled, Mishchenko, and
  Richt{\'a}rik]{khaled2019first}
Ahmed Khaled, Konstantin Mishchenko, and Peter Richt{\'a}rik.
\newblock First analysis of local gd on heterogeneous data.
\newblock \emph{arXiv preprint arXiv:1909.04715}, 2019.

\bibitem[Khanduri et~al.(2021)Khanduri, Sharma, Yang, Hong, Liu, Rajawat, and
  Varshney]{khanduri2021stem}
Prashant Khanduri, Pranay Sharma, Haibo Yang, Mingyi Hong, Jia Liu, Ketan
  Rajawat, and Pramod~K Varshney.
\newblock Stem: A stochastic two-sided momentum algorithm achieving
  near-optimal sample and communication complexities for federated learning.
\newblock In \emph{NeurIPS}, 2021.

\bibitem[Kim et~al.(2022)Kim, Kim, and Han]{kim2022multi}
Jinkyu Kim, Geeho Kim, and Bohyung Han.
\newblock Multi-level branched regularization for federated learning.
\newblock In \emph{ICML}, 2022.

\bibitem[Krizhevsky et~al.(2009)Krizhevsky, Hinton,
  et~al.]{krizhevsky2009learning}
Alex Krizhevsky, Geoffrey Hinton, et~al.
\newblock Learning multiple layers of features from tiny images.
\newblock 2009.

\bibitem[Le and Yang(2015)]{le2015tiny}
Ya Le and Xuan Yang.
\newblock Tiny imagenet visual recognition challenge.
\newblock \emph{CS 231N}, 7\penalty0 (7):\penalty0 3, 2015.

\bibitem[Lee et~al.(2022)Lee, Jeong, Shin, Bae, and Yun]{lee2022preservation}
Gihun Lee, Minchan Jeong, Yongjin Shin, Sangmin Bae, and Se-Young Yun.
\newblock Preservation of the global knowledge by not-true distillation in
  federated learning.
\newblock In \emph{NeurIPS}, 2022.

\bibitem[Li et~al.(2021)Li, He, and Song]{li2021model}
Qinbin Li, Bingsheng He, and Dawn Song.
\newblock Model-contrastive federated learning.
\newblock In \emph{CVPR}, 2021.

\bibitem[Li et~al.(2019)Li, Sahu, Zaheer, Sanjabi, Talwalkar, and
  Smithy]{li2019feddane}
Tian Li, Anit~Kumar Sahu, Manzil Zaheer, Maziar Sanjabi, Ameet Talwalkar, and
  Virginia Smithy.
\newblock Feddane: A federated newton-type method.
\newblock In \emph{ACSCC}, 2019.

\bibitem[Li et~al.(2020{\natexlab{a}})Li, Sahu, Zaheer, Sanjabi, Talwalkar, and
  Smith]{li2020federated}
Tian Li, Anit~Kumar Sahu, Manzil Zaheer, Maziar Sanjabi, Ameet Talwalkar, and
  Virginia Smith.
\newblock Federated optimization in heterogeneous networks.
\newblock In \emph{MLSys}, 2020{\natexlab{a}}.

\bibitem[Li et~al.(2020{\natexlab{b}})Li, Huang, Yang, Wang, and
  Zhang]{li2019convergence}
Xiang Li, Kaixuan Huang, Wenhao Yang, Shusen Wang, and Zhihua Zhang.
\newblock On the convergence of fedavg on non-iid data.
\newblock In \emph{ICLR}, 2020{\natexlab{b}}.

\bibitem[McMahan et~al.(2017)McMahan, Moore, Ramage, Hampson, and
  y~Arcas]{mcmahan2017communication}
Brendan McMahan, Eider Moore, Daniel Ramage, Seth Hampson, and Blaise~Aguera y
  Arcas.
\newblock Communication-efficient learning of deep networks from decentralized
  data.
\newblock In \emph{AISTATS}, 2017.

\bibitem[Mu et~al.(2021)Mu, Shen, Cheng, Geng, Fu, Zhang, and
  Zhang]{mu2021fedproc}
Xutong Mu, Yulong Shen, Ke Cheng, Xueli Geng, Jiaxuan Fu, Tao Zhang, and Zhiwei
  Zhang.
\newblock Fedproc: Prototypical contrastive federated learning on non-iid data.
\newblock \emph{arXiv preprint arXiv:2109.12273}, 2021.

\bibitem[Nam et~al.(2022)Nam, Ye-Bin, and Oh]{nam2022fedpara}
Hyeon-Woo Nam, Moon Ye-Bin, and Tae-Hyun Oh.
\newblock Fedpara: Low-rank hadamard product for communication-efficient
  federated learning.
\newblock In \emph{ICLR}, 2022.

\bibitem[Ozfatura et~al.(2021)Ozfatura, Ozfatura, and
  G{\"u}nd{\"u}z]{ozfatura2021fedadc}
Emre Ozfatura, Kerem Ozfatura, and Deniz G{\"u}nd{\"u}z.
\newblock Fedadc: Accelerated federated learning with drift control.
\newblock In \emph{ISIT}, 2021.

\bibitem[Paszke et~al.(2019)Paszke, Gross, Massa, Lerer, Bradbury, Chanan,
  Killeen, Lin, Gimelshein, Antiga, et~al.]{paszke2019pytorch}
Adam Paszke, Sam Gross, Francisco Massa, Adam Lerer, James Bradbury, Gregory
  Chanan, Trevor Killeen, Zeming Lin, Natalia Gimelshein, Luca Antiga, et~al.
\newblock Pytorch: An imperative style, high-performance deep learning library.
\newblock In \emph{NeurIPS}, 2019.

\bibitem[Qu et~al.(2022)Qu, Li, Duan, Liu, Tang, and Lu]{qu2022generalized}
Zhe Qu, Xingyu Li, Rui Duan, Yao Liu, Bo Tang, and Zhuo Lu.
\newblock Generalized federated learning via sharpness aware minimization.
\newblock In \emph{ICML}, 2022.

\bibitem[Reddi et~al.(2021)Reddi, Charles, Zaheer, Garrett, Rush,
  Kone{\v{c}}n{\`y}, Kumar, and McMahan]{reddi2021adaptive}
Sashank~J Reddi, Zachary Charles, Manzil Zaheer, Zachary Garrett, Keith Rush,
  Jakub Kone{\v{c}}n{\`y}, Sanjiv Kumar, and Hugh~Brendan McMahan.
\newblock Adaptive federated optimization.
\newblock In \emph{ICLR}, 2021.

\bibitem[Reisizadeh et~al.(2020)Reisizadeh, Mokhtari, Hassani, Jadbabaie, and
  Pedarsani]{reisizadeh2020fedpaq}
Amirhossein Reisizadeh, Aryan Mokhtari, Hamed Hassani, Ali Jadbabaie, and
  Ramtin Pedarsani.
\newblock {FedPAQ}: A communication-efficient federated learning method with
  periodic averaging and quantization.
\newblock In \emph{AISTATS}, 2020.

\bibitem[Seo et~al.(2024)Seo, Kim, Kim, and Han]{seo2024relaxed}
Seonguk Seo, Jinkyu Kim, Geeho Kim, and Bohyung Han.
\newblock Relaxed contrastive learning for federated learning.
\newblock In \emph{CVPR}, 2024.

\bibitem[Shi et~al.(2023)Shi, Liang, Zhang, Tan, and Bai]{shi2022towards}
Yujun Shi, Jian Liang, Wenqing Zhang, Vincent~YF Tan, and Song Bai.
\newblock Towards understanding and mitigating dimensional collapse in
  heterogeneous federated learning.
\newblock In \emph{ICLR}, 2023.

\bibitem[Stich(2019)]{stich2018local}
Sebastian~U Stich.
\newblock Local sgd converges fast and communicates little.
\newblock In \emph{ICLR}, 2019.

\bibitem[Stich and Karimireddy(2019)]{stich2019error}
Sebastian~U Stich and Sai~Praneeth Karimireddy.
\newblock The error-feedback framework: Better rates for sgd with delayed
  gradients and compressed communication.
\newblock \emph{arXiv preprint arXiv:1909.05350}, 2019.

\bibitem[Wang and Joshi(2021)]{wang2021cooperative}
Jianyu Wang and Gauri Joshi.
\newblock Cooperative sgd: A unified framework for the design and analysis of
  local-update sgd algorithms.
\newblock \emph{Journal of Machine Learning Research}, 22\penalty0
  (213):\penalty0 1--50, 2021.

\bibitem[Wang et~al.(2019{\natexlab{a}})Wang, Tantia, Ballas, and
  Rabbat]{wang2019slowmo}
Jianyu Wang, Vinayak Tantia, Nicolas Ballas, and Michael Rabbat.
\newblock Slowmo: Improving communication-efficient distributed sgd with slow
  momentum.
\newblock In \emph{ICLR}, 2019{\natexlab{a}}.

\bibitem[Wang et~al.(2020)Wang, Liu, Liang, Joshi, and Poor]{wang2020tackling}
Jianyu Wang, Qinghua Liu, Hao Liang, Gauri Joshi, and H~Vincent Poor.
\newblock Tackling the objective inconsistency problem in heterogeneous
  federated optimization.
\newblock In \emph{NeurIPS}, 2020.

\bibitem[Wang et~al.(2019{\natexlab{b}})Wang, Tuor, Salonidis, Leung, Makaya,
  He, and Chan]{wang2019adaptive}
Shiqiang Wang, Tiffany Tuor, Theodoros Salonidis, Kin~K Leung, Christian
  Makaya, Ting He, and Kevin Chan.
\newblock Adaptive federated learning in resource constrained edge computing
  systems.
\newblock \emph{IEEE Journal on Selected Areas in Communications}, 37\penalty0
  (6):\penalty0 1205--1221, 2019{\natexlab{b}}.

\bibitem[Wang et~al.(2022)Wang, Lin, and Chen]{wang2022communication}
Yujia Wang, Lu Lin, and Jinghui Chen.
\newblock Communication-efficient adaptive federated learning.
\newblock In \emph{ICML}, 2022.

\bibitem[Wu et~al.(2023)Wu, Huang, Hu, and Huang]{wu2023faster}
Xidong Wu, Feihu Huang, Zhengmian Hu, and Heng Huang.
\newblock Faster adaptive federated learning.
\newblock In \emph{AAAI}, 2023.

\bibitem[Xu et~al.(2021)Xu, Wang, Wang, and Yao]{xu2021fedcm}
Jing Xu, Sen Wang, Liwei Wang, and Andrew Chi-Chih Yao.
\newblock Fedcm: Federated learning with client-level momentum.
\newblock \emph{arXiv preprint arXiv:2106.10874}, 2021.

\bibitem[Yu et~al.(2019)Yu, Yang, and Zhu]{yu2019parallel}
Hao Yu, Sen Yang, and Shenghuo Zhu.
\newblock Parallel restarted sgd with faster convergence and less
  communication: Demystifying why model averaging works for deep learning.
\newblock In \emph{AAAI}, 2019.

\bibitem[Zhang et~al.(2022)Zhang, Li, Li, Xu, Wu, Ding, and
  Wu]{zhang2022Federated}
Jie Zhang, Zhiqi Li, Bo Li, Jianghe Xu, Shuang Wu, Shouhong Ding, and Chao Wu.
\newblock Federated learning with label distribution skew via logits
  calibration.
\newblock In \emph{ICML}, 2022.

\bibitem[Zhang et~al.(2020)Zhang, Hong, Dhople, Yin, and Liu]{zhang2020fedpd}
Xinwei Zhang, Mingyi Hong, Sairaj Dhople, Wotao Yin, and Yang Liu.
\newblock Fedpd: A federated learning framework with optimal rates and
  adaptivity to non-iid data.
\newblock In \emph{arXiv preprint arXiv:2005.11418}, 2020.

\bibitem[Zhao et~al.(2018)Zhao, Li, Lai, Suda, Civin, and
  Chandra]{zhao2018federated}
Yue Zhao, Meng Li, Liangzhen Lai, Naveen Suda, Damon Civin, and Vikas Chandra.
\newblock Federated learning with non-iid data.
\newblock \emph{arXiv preprint arXiv:1806.00582}, 2018.

\bibitem[Zhu et~al.(2021)Zhu, Hong, and Zhou]{zhu2021data}
Zhuangdi Zhu, Junyuan Hong, and Jiayu Zhou.
\newblock Data-free knowledge distillation for heterogeneous federated
  learning.
\newblock In \emph{ICML}, 2021.

\end{thebibliography}
